\documentclass{article}
\usepackage[numbers,sort&compress]{natbib}
\usepackage[preprint]{neurips_2025}

\usepackage[utf8]{inputenc}
\usepackage[T1]{fontenc}

\usepackage[normalem]{ulem}
\usepackage[shortlabels,inline]{enumitem}
\usepackage{algorithm}
\usepackage{algpseudocode}
\usepackage{amsfonts}
\usepackage{amsmath}
\usepackage{amssymb}
\usepackage{amsthm}
\usepackage{bbm}
\usepackage{bm}
\usepackage{booktabs}
\usepackage{caption}
\usepackage{colortbl}
\usepackage{comment}
\usepackage{etoolbox}
\usepackage{gensymb}
\usepackage{graphicx}
\usepackage{makecell}
\usepackage{microtype}
\usepackage{multirow}
\usepackage{nicefrac}
\usepackage{overpic}
\usepackage{pifont}
\usepackage{soul}
\usepackage{subcaption}
\usepackage{textcomp}
\usepackage{url}
\usepackage{xcolor}

\usepackage{tikz}
\usetikzlibrary{spy,backgrounds}
\definecolor{cvprblue}{rgb}{0.21,0.49,0.74}
\definecolor{tablered}{rgb}{1, 0.7, 0.7}

\usepackage[pagebackref,breaklinks,colorlinks,citecolor=cvprblue]{hyperref}

\usepackage[capitalize]{cleveref}
\crefname{section}{section}{sections}
\Crefname{section}{Section}{Sections}
\Crefname{table}{Table}{Tables}
\crefname{table}{table}{tables}
\crefname{figure}{figure}{figures}
\Crefname{figure}{Figure}{Figures}
\crefname{equation}{}{}
\Crefname{equation}{Eq.}{Eqs.}

\newcommand{\supp}{\textit{Supplementary Material}\xspace}

\newcommand{\REMOVAL}[1]{}

\definecolor{Highlight}{HTML}{39b54a}  

\newcommand{\triangleThreeD}{T_{\text{\tiny{3D}}}\xspace}
\newcommand{\triangleTwoD}{T_{\text{\tiny{2D}}}\xspace}

\makeatletter
\AfterEndEnvironment{algorithm}{\let\@algcomment\relax}
\AtEndEnvironment{algorithm}{\kern2pt\hrule\relax\vskip3pt\@algcomment}
\let\@algcomment\relax
\newcommand\algcomment[1]{\def\@algcomment{\footnotesize#1}}
\renewcommand\fs@ruled{\def\@fs@cfont{\bfseries}\let\@fs@capt\floatc@ruled
\def\@fs@pre{\hrule height.8pt depth0pt \kern2pt}%
\def\@fs@post{}%
\def\@fs@mid{\kern2pt\hrule\kern2pt}%
\let\@fs@iftopcapt\iftrue}
\makeatother

\newcommand{\cmmnt}[1]{}

\makeatletter
\renewcommand{\paragraph}{%
\@startsection{paragraph}{4}%
{\z@}{-0pt}{-0.5em}%
{\normalfont\normalsize\bfseries}%
}
\makeatother

\setlist[itemize]{noitemsep,leftmargin=*,topsep=0em}
\setlist[enumerate]{noitemsep,leftmargin=*,topsep=0em}

\DeclareRobustCommand{\zoomin}[9]{ %
\begin{tikzpicture}[spy using outlines={rectangle,#9,magnification=#8,size=#6}]   
\node[anchor=south west,inner sep=0]  {\includegraphics[width=#7]{#1}};
\spy on (#2, #3) in node at (#4,#5);
\end{tikzpicture}
}

\title{
Triangle Splatting \\ for Real-Time Radiance Field Rendering
}

\begin{document}
\maketitle

\begin{figure}[ht]
\centering

\includegraphics[width=\linewidth]{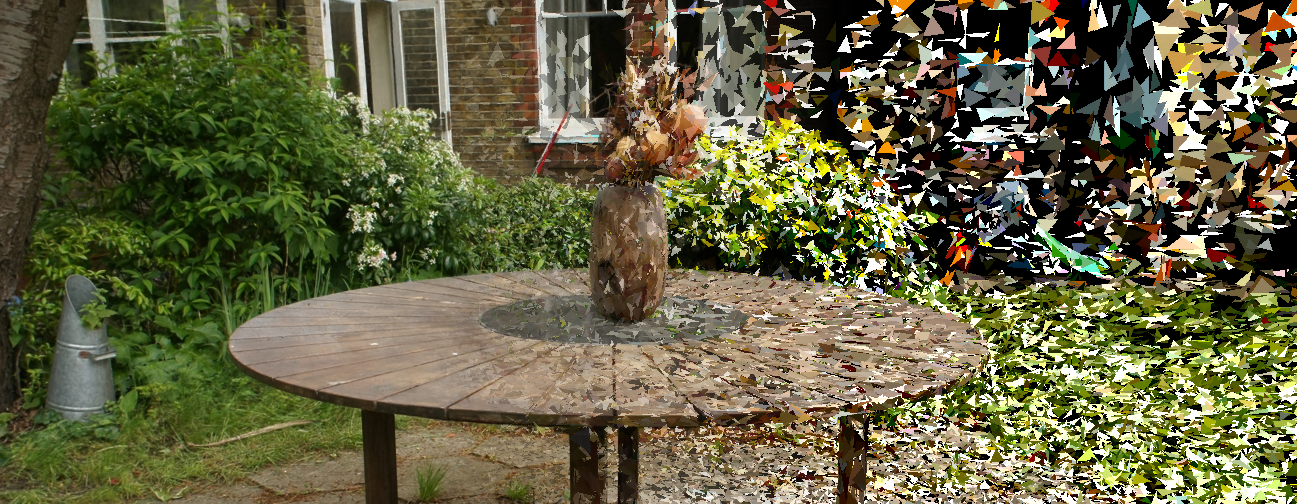}

\caption{\small
We propose a new representation for differentiable rendering based on the most classical of 3D primitives: the \textit{triangle}.
We show how a triangle soup (\ie unstructured, disconnected triangles) can be optimized effectively, generating state-of-the-art novel view synthesis images while being immediately compatible with classical rendering pipelines.
The figure shows the final rendered output (left), a visualization of soft blending (middle), and the rendering of a random subset of triangles to highlight their structure (right).
}%
\label{fig:teaser}
\end{figure}
\begin{abstract}
The field of computer graphics was revolutionized by models such as Neural Radiance Fields and 3D Gaussian Splatting, displacing triangles as the dominant representation for photogrammetry.
In this paper, we argue for a triangle comeback.
We develop a differentiable renderer that directly optimizes triangles via end-to-end gradients.
We achieve this by rendering each triangle as differentiable splats, combining the efficiency of triangles with the adaptive density of representations based on independent primitives.
Compared to popular 2D and 3D Gaussian Splatting methods, our approach achieves higher visual fidelity, faster convergence, and increased rendering throughput.
On the Mip-NeRF360 dataset, our method outperforms concurrent non-volumetric primitives in visual fidelity and achieves higher perceptual quality than the state-of-the-art Zip-NeRF on indoor scenes.
Triangles are simple, compatible with standard graphics stacks and GPU hardware, and highly efficient: 
for the \textit{Garden} scene, we achieve over 2,400 FPS at 1280×720 resolution using an off-the-shelf mesh renderer.
These results highlight the efficiency and effectiveness of triangle-based representations for high-quality novel view synthesis.
Triangles bring us closer to mesh-based optimization by combining classical computer graphics with modern differentiable rendering frameworks.
The project page is \href{https://trianglesplatting.github.io/}{https://trianglesplatting.github.io/}
\end{abstract}

\section{Introduction}%
\label{sec:intro}

\begin{figure}[t]
\setlength\mytmplen{0.48\linewidth}
\centering
\includegraphics[width=0.46\linewidth]{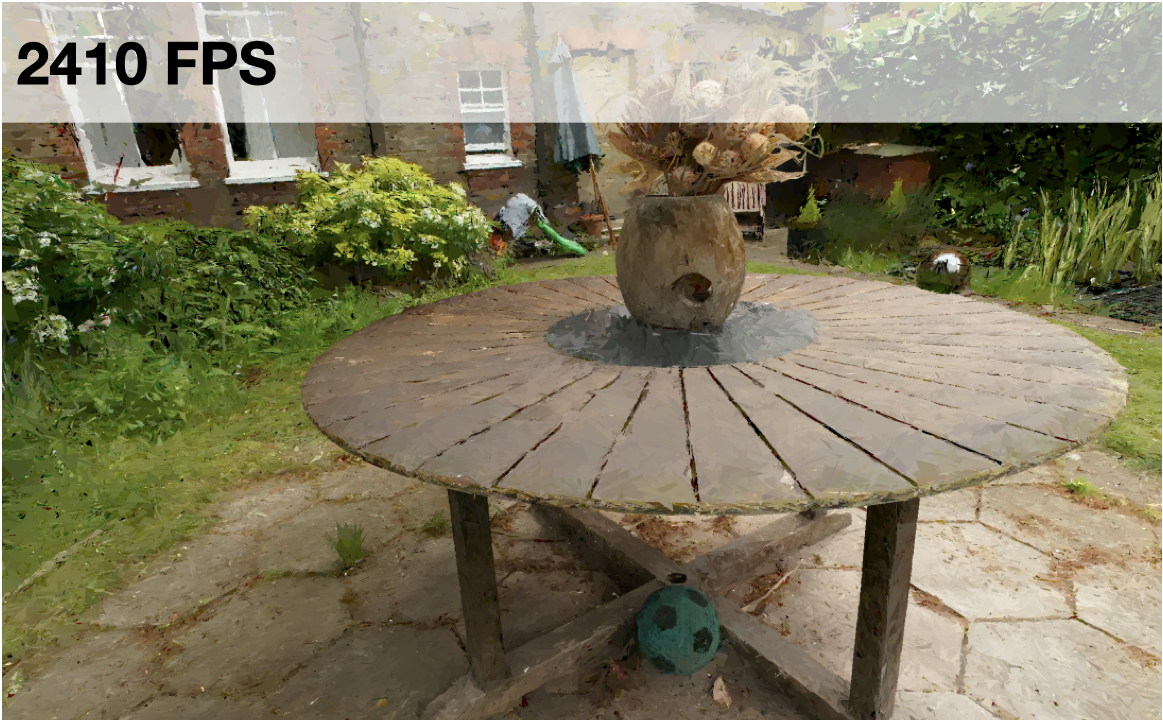}
\includegraphics[width=0.46\linewidth]{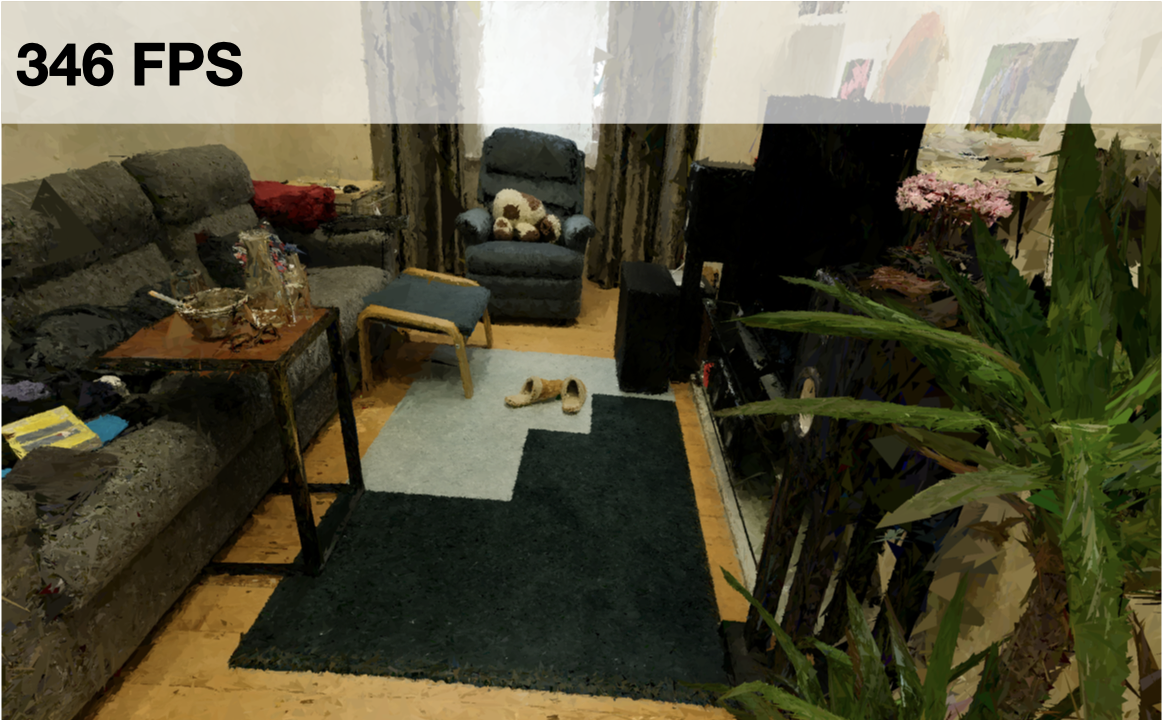}
\captionof{figure}{\small
\myTitle{Byproduct of the triangle-based representation}
\methodname unifies differentiable scene optimization with traditional graphics pipelines.
The optimized triangle soup is compatible with any mesh-based renderer, enabling seamless integration into traditional graphics pipelines. In a game engine, we render at 2,400+ FPS at 1280×720 resolution on an RTX4090 (left) and 300+ FPS on a consumer laptop (right).  
\label{fig:render_integration}
\vspace{-1em}
}
\end{figure}

One of the enduring challenges in 3D vision and graphics is identifying a truly \emph{universal} primitive for representing 3D content in a differentiable form, enabling gradient-based optimization of geometry and appearance.
Despite extensive research, no single data structure has emerged as a silver bullet.
Instead, researchers have explored a variety of approaches, including neural fields~\cite{Mildenhall2021NeRF}, explicit grids~\cite{FridovichKeil2022Plenoxels}, hash tables~\cite{Muller2022Instant}, convex primitives~\cite{Held20253DConvex, Deng2020CvxNet}, and anisotropic Gaussians~\cite{Kerbl20233DGaussian}, among others.
Conversely, in conventional graphics pipelines, the triangle remains the undisputed workhorse.
Game engines and other real-time systems primarily rely on triangles, as GPUs feature dedicated hardware pipelines for ultra-efficient triangle processing and rendering.
Although other primitives exist~(\eg, quads in 2D or tetrahedra in 3D), they can always be subdivided into triangles.
Moreover, surface reconstruction in 3D vision and graphics predominantly relies on triangle meshes to represent continuous, watertight geometry in an efficient, renderable form~\cite{Kazhdan2006Poisson}.

Despite their ubiquity, triangles are difficult to optimize in differentiable frameworks due to their discrete nature.
Early attempts at differentiable optimization softened the non-differentiable occlusion at polygon edges, enabling gradients from image loss to flow into geometry and appearance parameters~\cite{Kato2018Neural, Liu2019SoftRasterizer}.
However, these methods require a predefined mesh template, making them unsuitable when the scene's topology is unknown a priori.
As a result, they struggle to capture fine geometric details and adapt to novel structures.
To address these challenges, researchers adopted volumetric primitives, such as anisotropic 3D Gaussians in 3DGS~\cite{Kerbl20233DGaussian}, which can be optimized for high-quality novel view synthesis.
However, the unbounded support of Gaussians makes it difficult to define the ``surface'' of the representation,
and their inherent smoothness hinders accurate modeling of sharp details.
Surface structures can be partially restored using 2D Gaussian Splatting~\cite{Huang20242DGaussian} or 3D convex polytopes~\cite{Held20253DConvex}.
Yet, a pivotal question remains: \textit{can triangles themselves be optimized directly?}

Learning to optimize a ``triangle soup'' (\ie unstructured, disconnected triangles) via gradient-based methods could represent a major step towards the goal of template-free mesh optimization.
Such an approach leverages decades of GPU-accelerated triangle processing and the mature mesh processing literature, making it easier to integrate these techniques within differentiable rendering pipelines.

In this work, we introduce \textbf{Triangle Splatting}, a real-time differentiable renderer that splats a soup of triangles into screen space while enabling end-to-end gradient-based optimization.
Triangle Splatting merges the adaptability of Gaussians with the efficiency of triangle primitives, surpassing 3D Gaussian Splatting (3DGS), 2D Gaussian Splatting (2DGS), and 3D Convex Splatting (3DCS) in visual fidelity, training speed, and rendering throughput.
The optimized triangle soup is directly compatible with any standard mesh-based renderer.
As shown in \Cref{fig:render_integration}, our representation can be rendered in traditional game engines at over 2,400 FPS at 1280×720 resolution, demonstrating both high efficiency and seamless compatibility.
To our knowledge, Triangle Splatting is the first splatting-based approach to directly optimize triangle primitives for novel-view synthesis and 3D reconstruction, delivering state-of-the-art results while bridging classical rendering pipelines with modern differentiable frameworks.

\paragraph{Contributions.}

\begin{enumerate*}[label=\textbf{(\roman*)}]
\item We propose Triangle Splatting, a novel approach that directly optimizes unstructured triangles, bridging traditional computer graphics and radiance fields.
\item We introduce a differentiable window function for soft triangle boundaries, enabling effective gradient flow.
\item We demonstrate qualitatively and quantitatively that \methodname outperforms concurrent methods in terms of visual quality and rendering speed, and achieves superior perceptual quality compared to the state-of-the-art Zip-NeRF on indoor scenes.
\item The optimized triangles are directly compatible with standard mesh-based renderers, enabling seamless integration into traditional graphics pipelines. 

\end{enumerate*}
\section{Related work}%
\label{sec:related_work}

Neural radiance fields have become the de-facto standard for image-based 3D reconstruction~\cite{Mildenhall2020NeRF-eccv}.  
A large body of work has since addressed NeRF's slow training and rendering by introducing multi-resolution grids or hybrid representations~\cite{Chen2022TensoRF, FridovichKeil2022Plenoxels, Kulhanek2023TetraNeRF, Muller2022Instant, Sun2022Direct}, or baking procedures for real-time playback~\cite{Chen2023MobileNeRF, Hedman2021Baking, Reiser2021KiloNeRF, Reiser2023MERF}.
Improvements in robustness include anti-aliasing~\cite{Barron2021MipNeRF, Barron2022MipNeRF360, Barron2023ZipNeRF}, handling unbounded scenes~\cite{Barron2022MipNeRF360, Zhang2020NeRF++-arxiv}, and few-shot generalization~\cite{Chan2022Efficient, Du2023Learning, Jain2021Putting}.  
Despite their success, implicit fields still require costly volume integration at render time.
Our Triangle Splatting sidesteps this by optimizing \emph{explicit} triangles that are traced once per pixel, leading to comparable fidelity but orders-of-magnitude faster rendering.
For example, our triangles render ten times faster than Instant-NGP~\cite{Muller2022Instant}, while matching its optimization speed and achieving significantly higher visual fidelity.

\paragraph{Primitive-based differentiable rendering.}

Differentiable renderers back-propagate image loss to scene parameters, enabling end-to-end optimization of explicit primitives such as points~\cite{Gross2007Point,Kato2018Neural}, voxels~\cite{FridovichKeil2022Plenoxels}, meshes~\cite{Kato2018Neural, Liu2019SoftRasterizer, Loper2014OpenDR}, and Gaussians~\cite{Kerbl20233DGaussian}.  
3D Gaussian Splatting~\cite{Kerbl20233DGaussian} demonstrated that millions of anisotropic Gaussians can be fitted in minutes and rendered in real time.
Follow-up work improved anti-aliasing~\cite{Yu2024MipSplatting}, offered exact volumetric integration~\cite{Mai2024EVER-arxiv}, or modeled dynamics~\cite{Luiten2024Dynamic,Zhou2024HUGS}.  
Because Gaussians have infinite support and inherently smooth fall-off, they struggle with sharp creases and watertight surfaces; recent extensions therefore experiment with alternative kernels~\cite{Huang2024Deformable-arxiv}, learnable basis functions~\cite{Chen2024Beyond-arxiv}, or linear primitives~\cite{vonLutzow2025LinPrim-arxiv}.  
Convex Splatting~\cite{Held20253DConvex} replaced Gaussians with smooth convexes, capturing hard edges more faithfully, but at the cost of slow optimization time and larger memory footprints.
Compared with Gaussian~\cite{Huang20242DGaussian,Kerbl20233DGaussian} or Convex Splatting~\cite{Held20253DConvex}, which explored either \emph{volumetric} (\eg Gaussian, voxel) or \emph{solid} (\eg convex, tetrahedral) primitives, Triangle Splatting proposes \emph{surface} primitives, aligning with the surface of solid objects most typically found in real-world scenes.  
In extensive experiments, we show that \methodname surpasses 3DGS~\cite{Kerbl20233DGaussian}, 2DGS~\cite{Huang20242DGaussian}, and 3DCS~\cite{Held20253DConvex} in visual quality and speed of rendering and optimization.
\section{Method}%
\label{sec:methodology}

We address the problem of reconstructing a photorealistic 3D scene from multiple images.
To this end, we propose a scene representation that enables efficient, differentiable rendering and can be directly optimized by minimizing a rendering loss.
Similar to prior work, the scene is represented by a large collection of simple geometric primitives.
However, where 3DGS used 3D Gaussians~\cite{Kerbl20233DGaussian}, 3DCS used 3D convex hulls~\cite{Held20253DConvex}, and 2DGS used 2D Gaussians~\cite{Huang20242DGaussian}, we propose the simplest and most efficient primitive, \textit{triangles}.
First, \cref{sec:triangle_rasterization} explains how these triangles are rendered on an image.
Then, \cref{sec:pruning} describes how we adaptively prune and densify the triangle representation.
Finally, \cref{sec:optimization} describes how to optimize the triangles' parameters to fit the input images.

\subsection{Differentiable rasterization}%
\label{sec:triangle_rasterization}
Our primitives are 3D triangles $\triangleThreeD$, each defined by three vertices $\mathbf{v}_i \in \mathbb{R}^3$, a color $\mathbf{c}$, a smoothness parameter $\sigma$ and an opacity~$o$. 
The three vertices can move freely during optimization.
To render a triangle, we first project each vertex $\mathbf{v}_i$ to the image plane using a standard pinhole camera model.
The projection involves the intrinsic camera matrix $\mathbf{K}$ and the camera pose (parameterized by rotation $\mathbf{R}$ and translation $\mathbf{t}$):
$
\mathbf{q}_i = \mathbf{K} \left( \mathbf{R} \mathbf{v}_i + \mathbf{t} \right)
$,
with $\mathbf{q}_i \in \mathbb{R}^2$ forming the projected triangle $\triangleTwoD$ in the 2D image space.
Instead of rendering the triangle as fully opaque, we weigh its influence smoothly, based on a window function $I$ mapping pixels $\mathbf{p}$ to values in the $[0,1]$ range.
As we discuss below, the choice of this function is of critical importance.
Once the triangles are projected, the color of each image pixel~$\mathbf{p}$ is computed by accumulating contributions from all overlapping triangles, in depth order, treating the value $I(\mathbf{p})$ as opacity.
The rendering equation is the same as the one used in prior works~\cite{Kerbl20233DGaussian, Held20253DConvex}, and refer the reader to~\cite{Tagliasacchi2022Volume-arxiv} for its derivation.

\begin{figure}[t]
\hfill
\newcommand{\yoffone}{-1}
\newcommand{\yofftwo}{30}
\begin{overpic}[width=.92\linewidth]{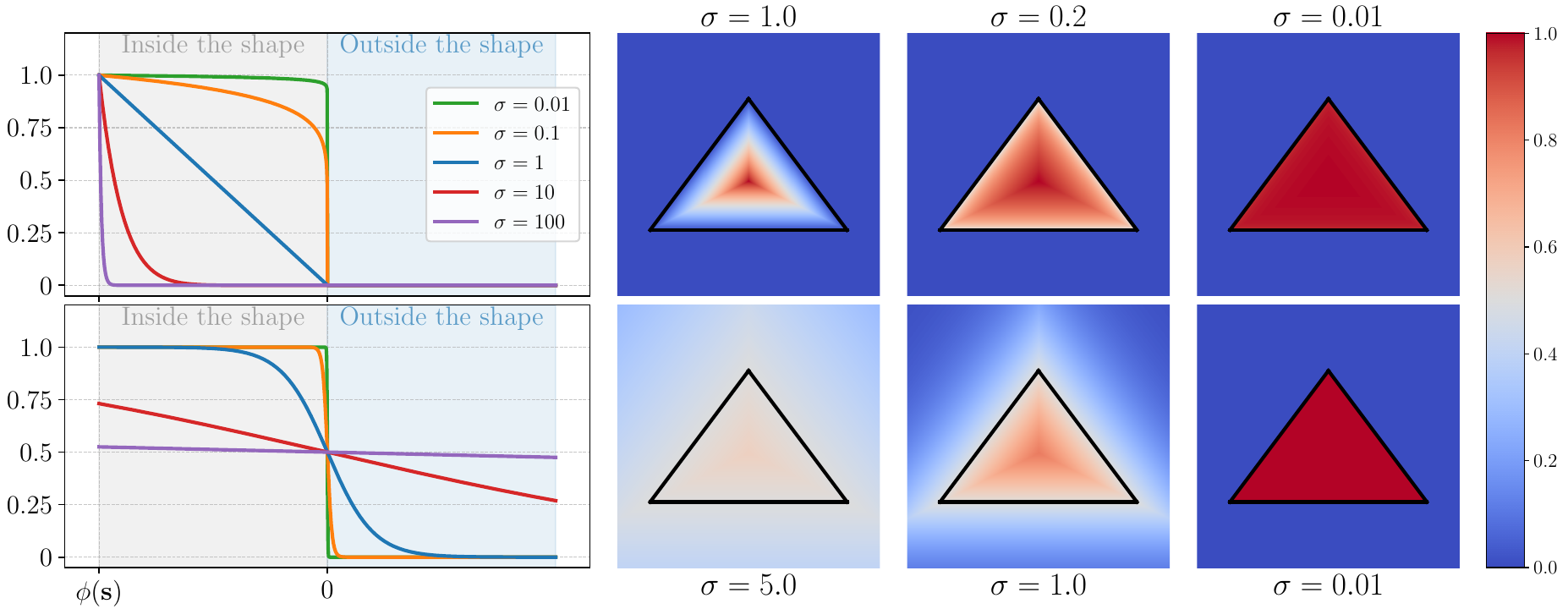}
\put(-8, 26.6){Eq.~(\ref{eq:indicator1})}
\put(-8,  9.2){Eq.~(\ref{eq:indicator2})}
\end{overpic}
\caption{\small
\myTitle{Triangle window function (1D and 2D)}
We visualize the window functions of prior works \cite{Held20253DConvex,Deng2020CvxNet} (bottom) vs.~the one introduced in our paper (top) in both 1D (left) and 2D (right).
We show how the window function changes as we vary the smoothness control parameter $\sigma$.
As $\sigma$ decreases, note that both can approximate the window function of a triangle.
However, as $\sigma$ increases, the support of \cref{eq:indicator2} exceeds the footprint of the triangle, making it unsuitable for \textit{rasterization} workloads.
In the limit, \cref{eq:indicator2} becomes globally supported, with a window value of 0.5 everywhere, causing every triangle to contribute to the color of every pixel in the image.
}%
\label{fig:indicator}
\end{figure}

\paragraph{A new window function.}

We first describe how the window function $I$ is defined, which is one of our core contributions.
We start by defining the \emph{signed distance field} (SDF) $\phi$ of the 2D triangle in image space.
This is given by:
$$
\phi(\mathbf{p}) = \max_{i\in \{1,2,3\}} L_i(\mathbf{p}),
\qquad
L_i(\mathbf{p}) = \mathbf{n}_i \cdot \mathbf{p} + d_i,
$$
where $\mathbf{n}_i$ are the unit normals of the triangle edges pointing outside the triangle, and $d_i$ are offsets such that the triangle is given by the zero-level set of the function $\phi$.
The signed distance field $\phi$ thus takes positive values outside the triangle, negative values inside, and equals zero on its boundary.
Let $\mathbf{s}\in\mathbb{R}^2$ be the \emph{incenter} of the projected triangle $\triangleTwoD$ (\ie, the point within the triangle where the signed distance field is minimum).
With this, we define our new window function $I$ as:
\begin{equation}\label{eq:indicator1}
I(\mathbf{p}) = \text{ReLU}\left( \frac{\phi(\mathbf{p})}{\phi(\mathbf{s})} \right)^{\sigma}
\quad
\text{such that}
\quad
I(\mathbf{p})
~
\begin{cases}
= 1 ~~\text{at the triangle incenter},\\
= 0 ~~\text{at the triangle boundary},\\
= 0 ~~\text{outside the triangle}.
\end{cases}%
\end{equation}
Here, the parameter $\sigma > 0$ controls the \emph{smoothness} of the window function.
$\phi(\mathbf{p})$ is negative inside the triangle, and $\phi(\mathbf{s})$ is its smallest (most negative) value, so the ratio
$
\phi(\mathbf{p})
/
\phi(\mathbf{s})
$
is positive inside the triangle, equal to one at the incenter, and equal to zero at the boundary.
This formulation has three important properties:
\begin{enumerate*}[label=\textbf{(\roman*)}]
\item\label{prop:max} there is a point (the incenter) inside the triangle where the window function obtains the maximum value of one; 
\item\label{prop:zero} the window function is zero at the boundary and outside the triangle so that its support tightly fits the triangle; and
\item\label{prop:smooth} a single parameter can easily control the smoothness of the window function.
\end{enumerate*}
\Cref{fig:indicator} illustrates that all three properties are satisfied for different values of $\sigma$; in particular, for
$
\sigma \rightarrow 0
$
our window function converges to the window function of the triangle.
For larger values, the window function transitions smoothly from zero at the boundary to one in the middle, and for
$
\sigma \rightarrow \infty
$
the window function becomes a delta function at the incenter.

\paragraph{Discussion: window function alternatives.}

Related works~\cite{Held20253DConvex, Deng2020CvxNet} use the $\operatorname{LogSumExp}$ function to approximate $\max$ in the definition of the signed distance field:
\scalebox{0.9}
{$
\phi(\mathbf{p}) = \log \sum_{i=1}^3 \exp L_i(\mathbf{p})
$}.
However, we observed that, for small triangles, this $\max$ approximation is poor, to the point that \textit{only one} of the three vertices has any influence on the final shape.
We thus opted to use the actual $\max$ function which, while not smooth everywhere, accurately defines the signed distance field.
Further, related work~\cite{Held20253DConvex, Deng2020CvxNet, Liu2019SoftRasterizer} also use a different definition for the window function $I$ based on sigmoid:
\begin{equation}
\label{eq:indicator2}
I(\mathbf{p})
=
\operatorname{sigmoid}(-\sigma^{-1} \, \phi(\mathbf{p}))
\quad
\text{such that}
\quad
I(\mathbf{p})
~
\begin{cases}
>1/2 ~~\text{inside the triangle},\\
=1/2 ~~\text{at the triangle boundary},\\
<1/2 ~~\text{outside the triangle}.
\end{cases}
\end{equation}
This definition fails to meet properties~\ref{prop:max} and~\ref{prop:zero} above, as the maximum can be less than $1$, and the support of the window function can be significantly larger than the triangle.
This is illustrated in~\Cref{fig:indicator}, where $\sigma \rightarrow \infty$ results in a constant value everywhere in $\mathbb{R}^2$.

\paragraph{Discussion: simpler depth-dependent scaling.}

In 3D Gaussian Splatting, each 3D Gaussian is defined in world space by a full covariance matrix, which is mapped to image space by accounting for the projective transformation, resulting in a 2D covariance matrix inversely proportional to depth.
The effect is a 2D Gaussian whose size scales consistently with depth.
In Convex Splatting~\cite{Held20253DConvex}, the 2D convex hull scales automatically with depth, but not the window function smoothness parameter $\sigma$.
Because the latter is defined in pixel units, it must be scaled ``manually'' to achieve a depth-consistent effect.
In our case, this is unnecessary because of the normalization in \cref{eq:indicator1}: the same value of $\sigma$ results in consistently scaled 2D window functions for all depth values.

\subsection{Adaptive pruning and splitting}%
\label{sec:pruning}

Triangles have a compact spatial domain (and, therefore, a compact gradient);
hence, we need a mechanism to control coverage of the spatial domain by the triangles, modulating their density and thus representation power at different locations.
This is achieved by pruning and densification routines~(respectively decreasing and increasing the representation power), analogously to 3DGS~\cite{Kerbl20233DGaussian}.

\begin{figure}[t]
\setlength\mytmplen{0.48\linewidth}
\centering

\centering
\includegraphics[width=0.48\linewidth]{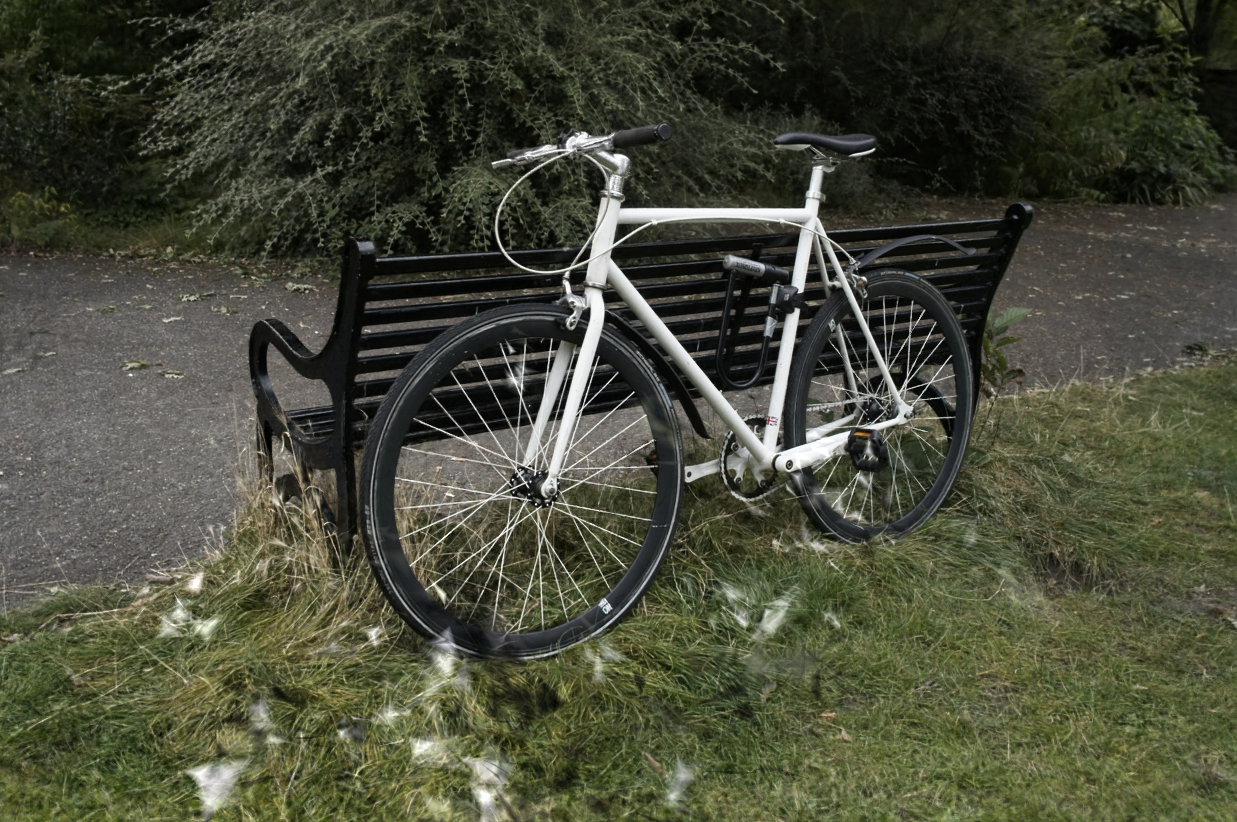}
\includegraphics[width=0.48\linewidth]{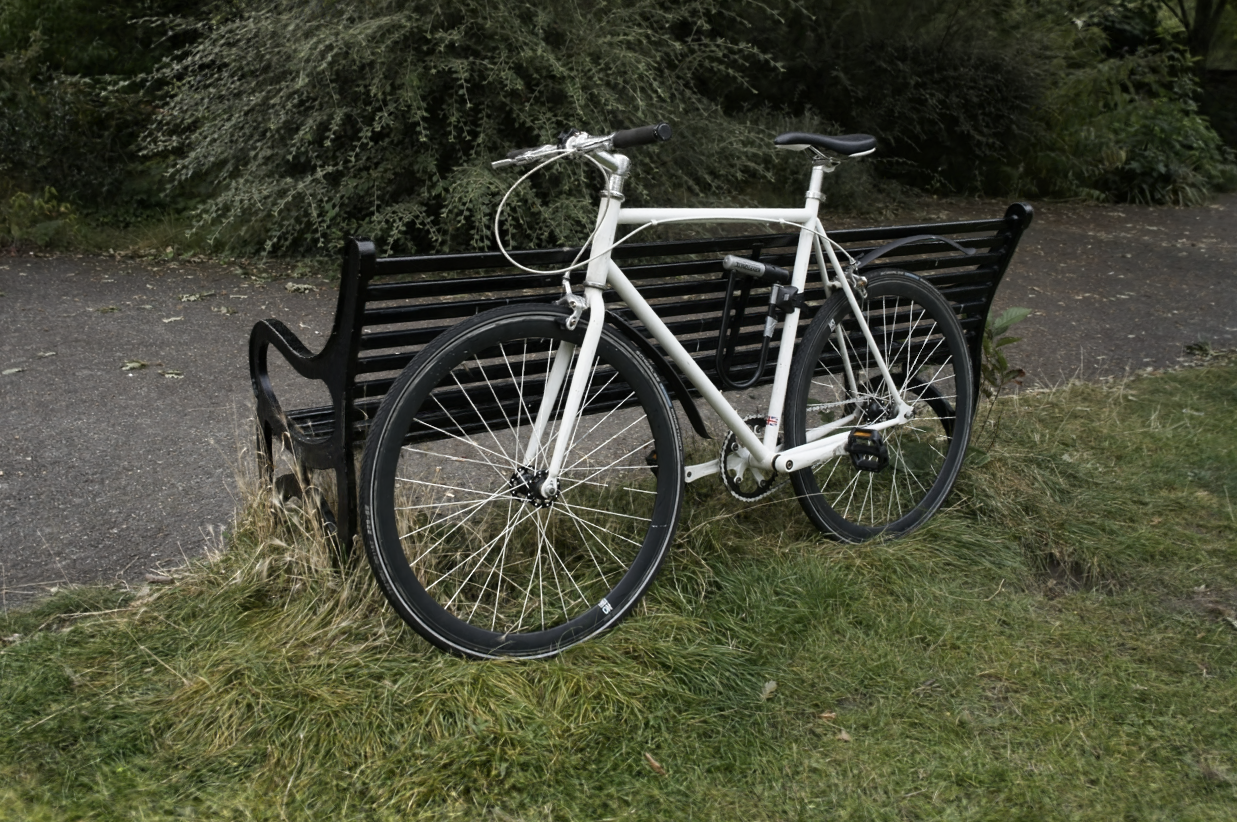}

\captionof{figure}{ \small
\myTitle{Triangle prunning}
To reduce floaters, we prune triangles seen in fewer than two views with more than one pixel of coverage, removing those that are overfitted by a single training view.
\label{fig:pruning}
}
\end{figure}

\paragraph{Pruning.}
During rasterization, we calculate the maximum volume rendering blending weight~$T \cdot o$ (where $T$ is transmittance, and $o$ is opacity) for each triangle, and prune all triangles whose maximum weight is less than a user-defined threshold $\tau_\text{prune}$ across all training views.
Additionally, we prune all triangles that are not rendered at least \textit{twice} with more than one pixel.
In other words, we remove triangles that explain small amounts of data within a single view and are therefore likely to have overfitted to the training data. 
Figure~\ref{fig:pruning} illustrates the impact of this pruning strategy.

\paragraph{Densification.}

Instead of relying on manually tuned heuristics for adding shapes, we adopt the probabilistic framework based on MCMC introduced by~\citet{Kheradmand20243DGaussian}.
At each densification step, we sample from a probability distribution to guide where new shapes should be added.
\citet{Kheradmand20243DGaussian} stochastically allocates new Gaussians proportionally to the \textit{opacity}, and we extend this idea to our representation by incorporating the sharpness parameter~$\sigma$.
Since both opacity and $\sigma$ are learned during training, we build the probability distribution directly from these parameters by alternating between using the inverse of $\sigma$ and the opacity for Bernoulli sampling.
In particular, we preferentially sample triangles with low $\sigma$ values, \ie \textit{solid} triangles.
Because of our window function, the triangle's influence is bounded by its projected geometry, and the diffusion remains confined within the triangle itself.
In high-density regions, many triangles overlap at each pixel, allowing each shape to adopt a higher $\sigma$ and thus a softer contribution.
In contrast, in low-density regions, where fewer triangles influence a pixel, each triangle must contribute more to the reconstruction.
As a result, it adopts a lower $\sigma$ to increase the contribution across its interior, ensuring maximal coverage within the geometric bounds and producing a more solid appearance.

Further, taking inspiration from~\citet{Kheradmand20243DGaussian}, we design updates to avoid disrupting the sampling process. 
In particular, we require that the probability of the state (\ie the current set of parameters of all triangles) remains \textit{unchanged} before and after the transition, allowing it to be interpreted as a move between equally probable samples, preserving the integrity of the Markov chain.
To preserve a consistent representation across sampling steps, we apply \textit{midpoint subdivision} to the selected triangles. 
Each triangle is split into four smaller ones by connecting the midpoints of its edges, ensuring that the combined area and spatial region of the new triangles match that of the original.
As in our parametrization, a triangle is defined by 3D vertices, making this operation straightforward to perform.
Finally, if a triangle is smaller than the threshold $\tau_\text{small}$, we do not split it.
Instead, we clone it and add random noise along the triangle's plane orientation.

\subsection{Optimization}%
\label{sec:optimization}

Our method starts from a set of images and their corresponding camera parameters, calibrated via SfM~\cite{Schonberger2016Structure}, which also produces a sparse point cloud. 
We create a 3D triangle for each 3D point in the sparse point cloud.
We optimize the 3D vertex positions $\{\mathbf{v}_1,\mathbf{v}_2,\mathbf{v}_3\}$, sharpness $\sigma$, opacity $o$, and spherical harmonic color coefficients $\mathbf{c}$ of all such 3D triangles by minimizing the rendering error from the given posed views.
The initialization is done as follows.
Let $\mathbf{q}\in\mathbb{R}^3$ be a SfM 3D point and let $d$ be the average Euclidean distance to its three nearest neighbors.
We initialize the corresponding 3D triangle to be approximately equilateral, randomly oriented, and with a size proportional to $d$.
To do this, we sample uniformly at random three vertices $\{\mathbf{u}_1,\mathbf{u}_2,\mathbf{u}_3\}$ from the unit sphere, we scale them by $d$, and we add $\mathbf{q}$ to center them at the point $\mathbf{q}$:
$
\mathbf{v}_i = \mathbf{q} + k \cdot d \cdot \mathbf{u}_i,
$ where $k \in \mathbb{R}$ is a scaling constant.
Our training loss combines the photometric $\mathcal{L}_1$ and $\mathcal{L}_{\text{D-SSIM}}$ terms~\cite{Kerbl20233DGaussian}, the opacity loss $\mathcal{L}_o$~\cite{Kheradmand20243DGaussian}, and the distortion $\mathcal{L}_d$ and normal $\mathcal{L}_n$ losses~\cite{Huang20242DGaussian}. 
Finally, we add a size regularization term $\mathcal{L}_s = 2 \left\| (\mathbf{v}_1 - \mathbf{v}_0) \times (\mathbf{v}_2 - \mathbf{v}_0) \right\|_2^{-1}$, to encourage larger triangles.
The final loss $\mathcal{L}$ is given by:
\begin{equation}
    \mathcal{L} = (1 - \lambda) \mathcal{L}_1 + \lambda \mathcal{L}_{\text{D-SSIM}} + \beta_1 \mathcal{L}_o + \beta_2 \mathcal{L}_d + \beta_3 \mathcal{L}_n + \beta_4 \mathcal{L}_s\, .
\end{equation}
The full list of thresholds and hyperparameters is detailed in the \supp.

\section{Experiments}%
\label{sec:results}
We compare our method to competitive photorealistic novel view synthesis techniques on the standard benchmarks Mip-NeRF360~\cite{Barron2022MipNeRF360} and Tanks and Temples (T\&T)~\cite{Knapitsch2017Tanks}.
We consider 3DCS~\cite{Held20253DConvex}, which is the most closely related method, as well as to non-volumetric primitives such as BBSplat~\cite{Svitov2024BillBoard-arxiv} and 2DGS~\cite{Huang20242DGaussian}.
We also consider primitive-based volumetric methods, including 3DGS~\cite{Kerbl20233DGaussian}, 3DGS-MCMC~\cite{Kheradmand20243DGaussian}, and DBS~\cite{Liu2025Deformable-arxiv}.
Additionally, we evaluate against implicit neural rendering methods such as Instant-NGP~\cite{Muller2022Instant}, Mip-NeRF360~\cite{Barron2022MipNeRF360}, and the state-of-the-art in novel view synthesis Zip-NeRF~\cite{Barron2023ZipNeRF}.
We evaluate the visual quality of the synthesized images using standard metrics from the novel view synthesis literature: SSIM, PSNR, and LPIPS.
We also report the average training time, rendering speed, and memory usage.
FPS and training time were obtained using an NVIDIA A100.

\paragraph{Implementation details.}
We set the spherical harmonics degree to 3, resulting in 59 parameters per triangle, matching the number of parameters for a single 3D Gaussian primitive in 3DGS~\cite{Kerbl20233DGaussian}\@.
We use different hyperparameter settings for indoor and outdoor scenes; see our \supp.

\subsection{Novel-view synthesis}
\Cref{tab:main_results} presents the quantitative results on the T\&T dataset, as well as on the indoor and outdoor scenes from the Mip-NeRF360 dataset.
In comparison with planar primitive methods, Triangle Splatting achieves a higher visual quality, with a significant improvement in LPIPS~(the metric that best correlates with human visual perception). %
Specifically, \methodname improves over 2DGS and BBSplat by 25\% and 19\% on Mip-NeRF360, respectively.
\methodname achieves consistently better LPIPS scores in outdoor scenes, surpassing both 2DGS and BBSplat.
Similarly, on the T\&T dataset, \methodname yields a substantial improvement over 2DGS and BBSplat.
\begin{table}[t]
\centering
\resizebox{0.98\linewidth}{!}{
\begin{tabular}{@{}l|ccc|ccc|cc||cccc}
& \multicolumn{3}{c|}{Outdoor Mip-NeRF 360} & \multicolumn{3}{c|}{Indoor Mip-NeRF 360} & \multicolumn{2}{c||}{Aver. Mip-NeRF 360} & \multicolumn{4}{c}{Tanks \& Temples} \\ 
& LPIPS~$\downarrow$ & PSNR~$\uparrow$ & SSIM~$\uparrow$ 
& LPIPS~$\downarrow$ & PSNR~$\uparrow$ & SSIM~$\uparrow$ 
& LPIPS~$\downarrow$ & FPS~$\uparrow$
& LPIPS~$\downarrow$ & PSNR~$\uparrow$ & SSIM~$\uparrow$ & FPS~$\uparrow$ \\
\midrule
& \multicolumn{11}{c}{Implicit Methods} \\
\midrule
 Instant-NGP~\cite{Muller2022Instant} & -& -& -& - & - & - & 0.331 &9.43 &0.305 &21.92 & 0.745 &14.4 \\
 Mip-NeRF360 \cite{Barron2022MipNeRF360} & 0.283 & 24.47 & 0.691 & 0.179 & 31.72 & 0.917 & 0.237 & 0.06 & 0.257  & 22.22 &  0.759 & 0.14 \\
Zip-NeRF~\cite{Barron2023ZipNeRF}  & 0.207 & 25.58 & 0.750 & 0.167  & 32.25 & 0.926 & 0.189  & 0.18 & - & - & - & - \\
\midrule
& \multicolumn{11}{c}{Volumetric Primitives} \\
\midrule
3DGS \cite{Kerbl20233DGaussian} & 0.234 & 24.64 & 0.731 & 0.189 & 30.41 & 0.920 & 0.214  & 134 & 0.183 & 23.14 & 0.841 & 154 \\
 3DGS-MCMC \cite{Kheradmand20243DGaussian} $\ddagger$ & 0.210 & 25.51 & 0.76 & 0.208 & 31.08   & 0.917 & 0.210 & 82 & 0.19 & 24.29  & 0.86 & 129 \\
 DBS \cite{Liu2025Deformable-arxiv} $\dagger$ &0.246 &25.10 &	0.751 & 0.22 &32.29 & 0.936 & 0.234 & 123 & 0.140  & 24.85 & 0.870 & 150 \\
3DCS \cite{Held20253DConvex} & 0.238 & 24.07 & 0.700 & 0.166  & 31.33 & 0.927 & 0.207  & 25 & 0.156  & 23.94 & 0.851 & 33 \\
\midrule
& \multicolumn{11}{c}{Non-Volumetric Primitives} \\
\midrule
BBSplat \cite{Svitov2024BillBoard-arxiv} $\dagger$  & 0.281 & 23.55 & 0.669 & 0.178 & 30.62 & 0.921 & 0.236  & 25  & 0.172 & \textbf{25.12} & \textbf{0.868} & 66 \\
2DGS \cite{Huang20242DGaussian}    & 0.246 & \textbf{24.34} & 0.717 & 0.195  & 30.40 & 0.916 & 0.252  & 64 & 0.212  & 23.13 & 0.831 & 122 \\
\textbf{\methodname}    & \textbf{0.217} & 24.27 & \textbf{0.722} & \textbf{0.160} & \textbf{30.80} & \textbf{0.928} & \textbf{0.191}  & \textbf{97}  & \textbf{0.143} & 23.14 & 0.857 &\textbf{ 165}
\vspace{0.5em}
\end{tabular}
}
\caption{ \small
\myTitle{Quantitative results (Mip-NeRF 360~\cite{Barron2022MipNeRF360} and Tank \& Temples~\cite{Knapitsch2017Tanks})}
We evaluate our method on both indoor and outdoor scenes, achieving state-of-the-art performance on the \textit{indoor} benchmark. Across all datasets, our approach consistently outperforms other non-volumetric primitives.
Bold scores indicate the best results among \textit{non-volumetric} methods. 
$\dagger$ denotes reproduced results, while $\ddagger$ marks results reported in \cite{Liu2025Deformable-arxiv}.}
\label{tab:main_results}
\vspace{-1.5em}
\end{table}

\begin{figure}[b]
\centering
\setlength{\mytmplen}{0.3\linewidth}
\begin{tabular}{@{}c@{\hskip 0.15in}c@{\hskip 0.15in}c@{}}
\makebox[\mytmplen][c]{Ground Truth} & 
\makebox[\mytmplen][c]{\textbf{\methodname{} (ours)}} & 
\makebox[\mytmplen][c]{2DGS} \\
\includegraphics[width=\mytmplen]{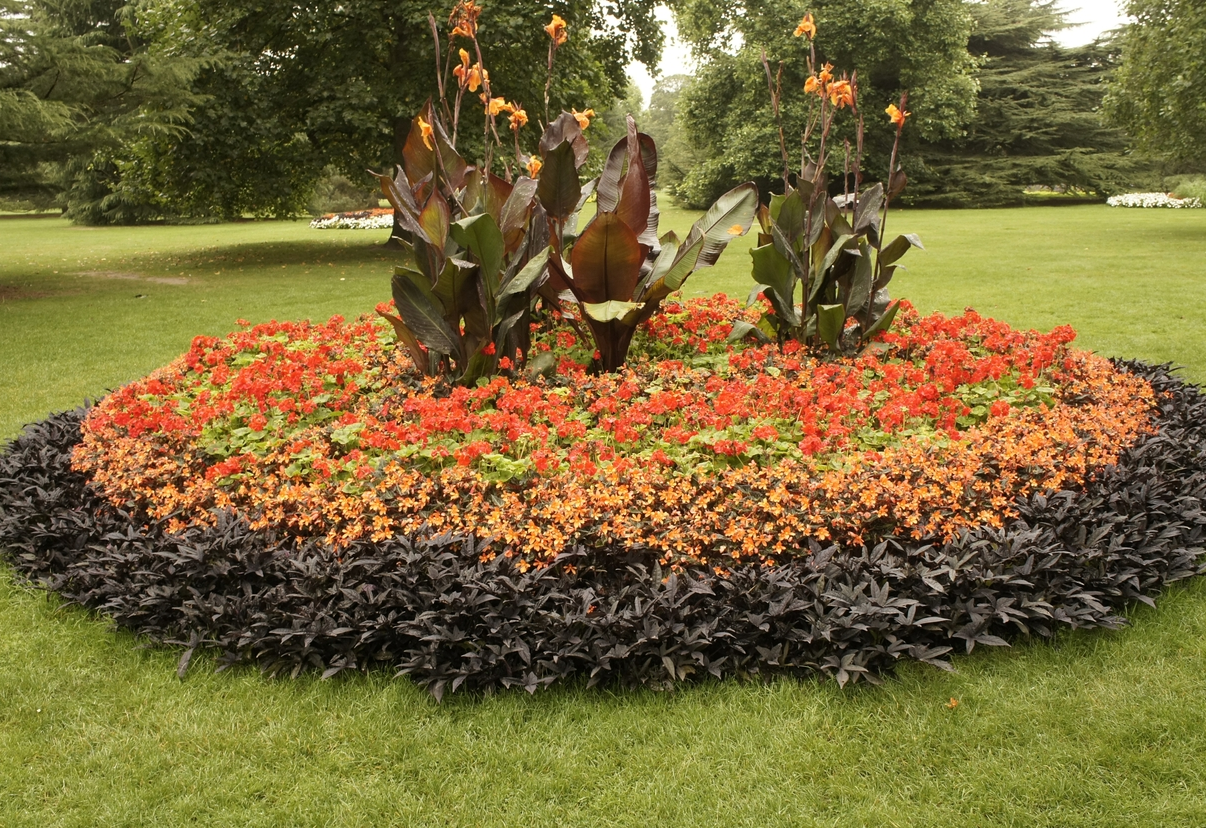} &
\includegraphics[width=\mytmplen]{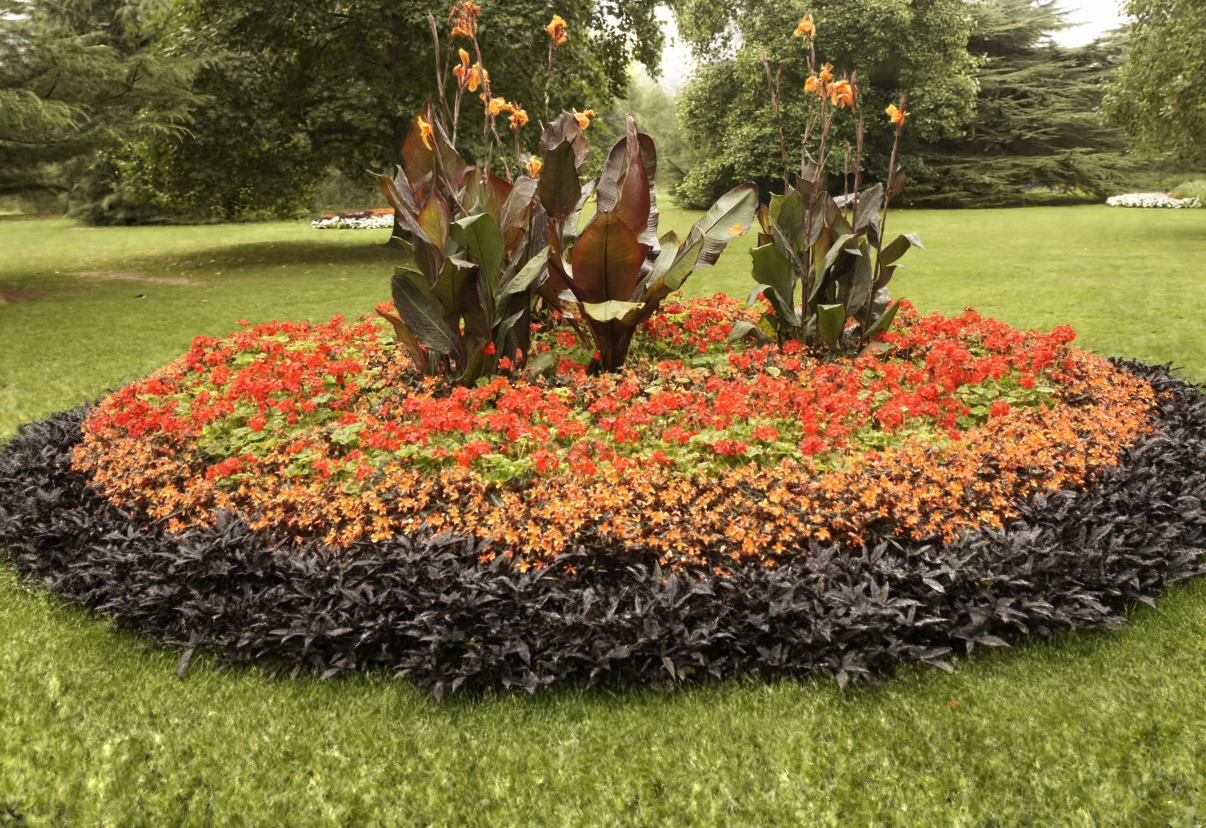} &
\includegraphics[width=\mytmplen]{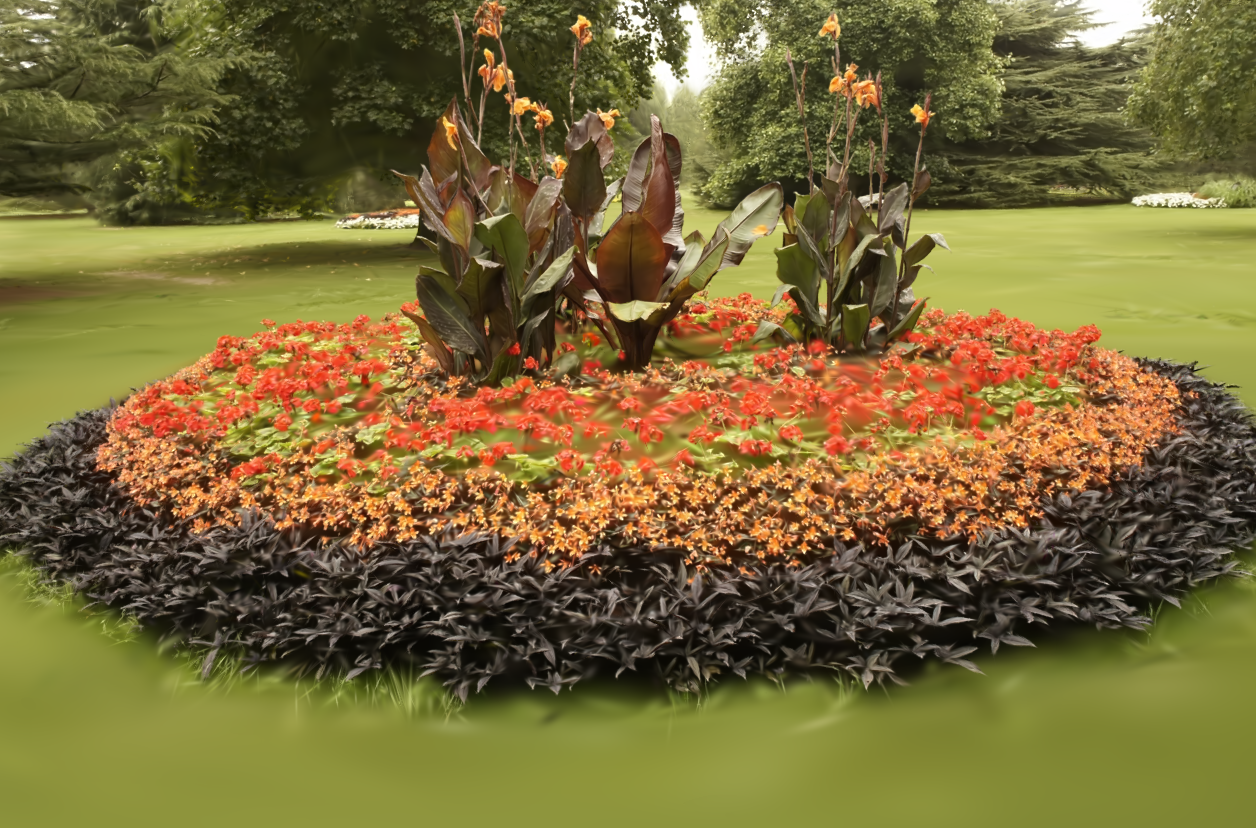} \\
\includegraphics[width=\mytmplen]{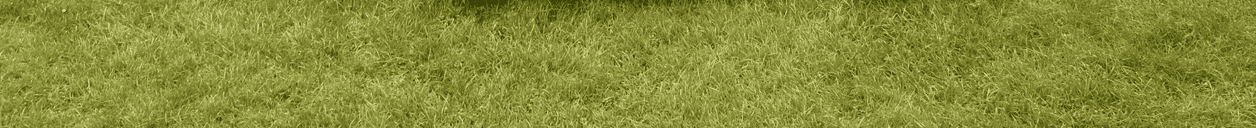} &
\includegraphics[width=\mytmplen]{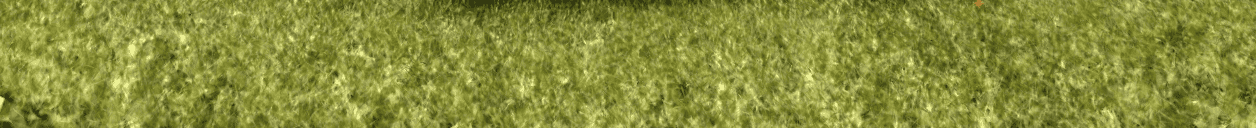} &
\includegraphics[width=\mytmplen]{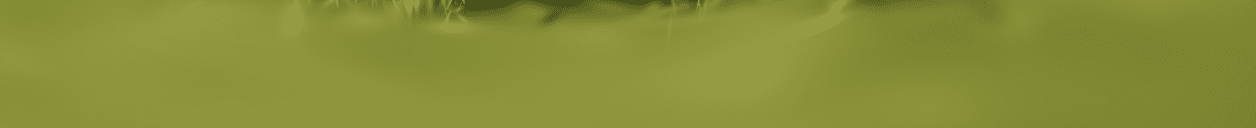} \\
\end{tabular}
\caption{\small
\myTitle{Limitations of PSNR}
Due to its inherent smoothness, the Gaussian primitive tends to perform better on the PSNR metric, which evaluates pixel-wise differences, despite being blurrier.
In the highlighted region, our method (TS) achieves a PSNR of 18.41, compared to 21.27 for 2DGS\@.
}%
\label{fig:psnr_limitation}
\end{figure}

While our method yields slightly lower PSNR values, this metric does not fully capture visual quality due to its inherent limitations~(PSNR generally rewards overly smooth reconstructions that regress to the mean).
As a result, smooth representations, such as Gaussian-based primitives, tend to perform better under PSNR, whereas sharper transitions from solid shapes may be penalized.
\Cref{fig:psnr_limitation} illustrates this limitation of PSNR\@:
although our reconstruction looks visually superior to that of 2DGS, the PSNR \textit{in the highlighted region} is $3$ PSNR higher for~2DGS\@.

Against volumetric primitive methods, Triangle Splatting achieves high visual quality, with a significant improvement in LPIPS\@.
Specifically, \methodname improves over 3DGS and 3DCS by $10\%$ and $7\%$ respectively on Mip-NeRF 360.
Compared to implicit methods, Triangle Splatting matches the visual quality of the state-of-the-art Zip-NeRF, with only a $0.002$ difference in LPIPS, while delivering over $500\times$ faster rendering performance.

Triangles are particularly effective in indoor or structured outdoor scenes, such as those with walls, cars, and other well-defined surfaces, where they can closely approximate geometry. 
This makes \methodname especially well-suited for indoor scenes, where it achieves state-of-the-art performance and outperforms 3DCS and Zip-NeRF\@.
In contrast, unstructured outdoor scenes pose greater challenges for planar primitives due to sparse or ambiguous geometry, making it harder to maintain visual consistency across views.
Despite these challenges, \methodname substantially narrows the performance gap and surpasses 3DGS and 3DCS on the T\&T dataset, achieving a lower LPIPS\@.

\begin{figure*}[t]
\centering
\setlength\mytmplen{0.28\linewidth}
\resizebox{\linewidth}{!}{ 

\begin{tabular}{c@{\hskip 0.2in}c@{\hskip 0.2in}c@{\hskip 0.2in}c}
    
    \makebox[\mytmplen]{Ground Truth} &
    \makebox[\mytmplen]{\textbf{\methodname (ours)}} &
    \makebox[\mytmplen]{2DGS} &
    \makebox[\mytmplen]{3DCS} \\

    \rotatebox{90}{\parbox{2.2cm}{\centering Flowers}}
    \zoomin{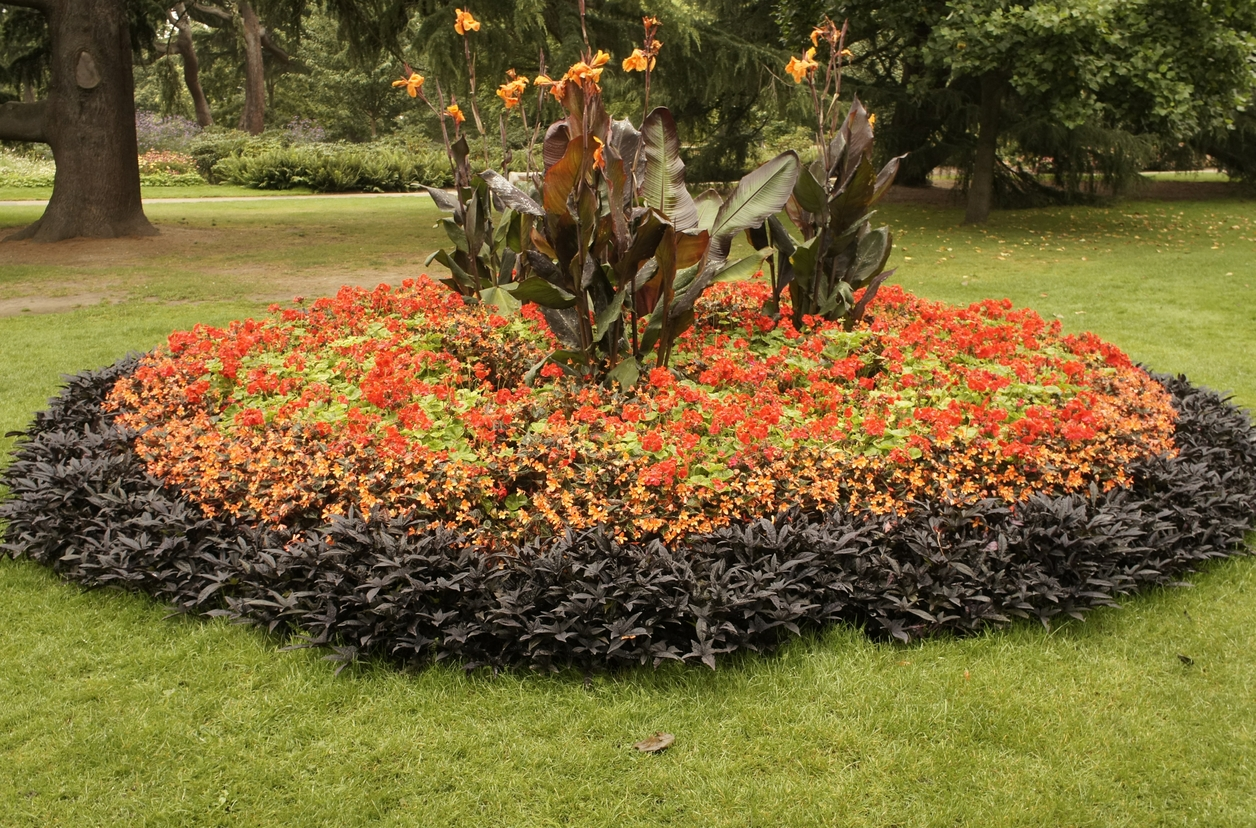}{0.40\mytmplen}{0.32\mytmplen}{0.81\mytmplen}{0.194\mytmplen}{1.2cm}{\mytmplen}{3.5}{red} &
    \zoomin{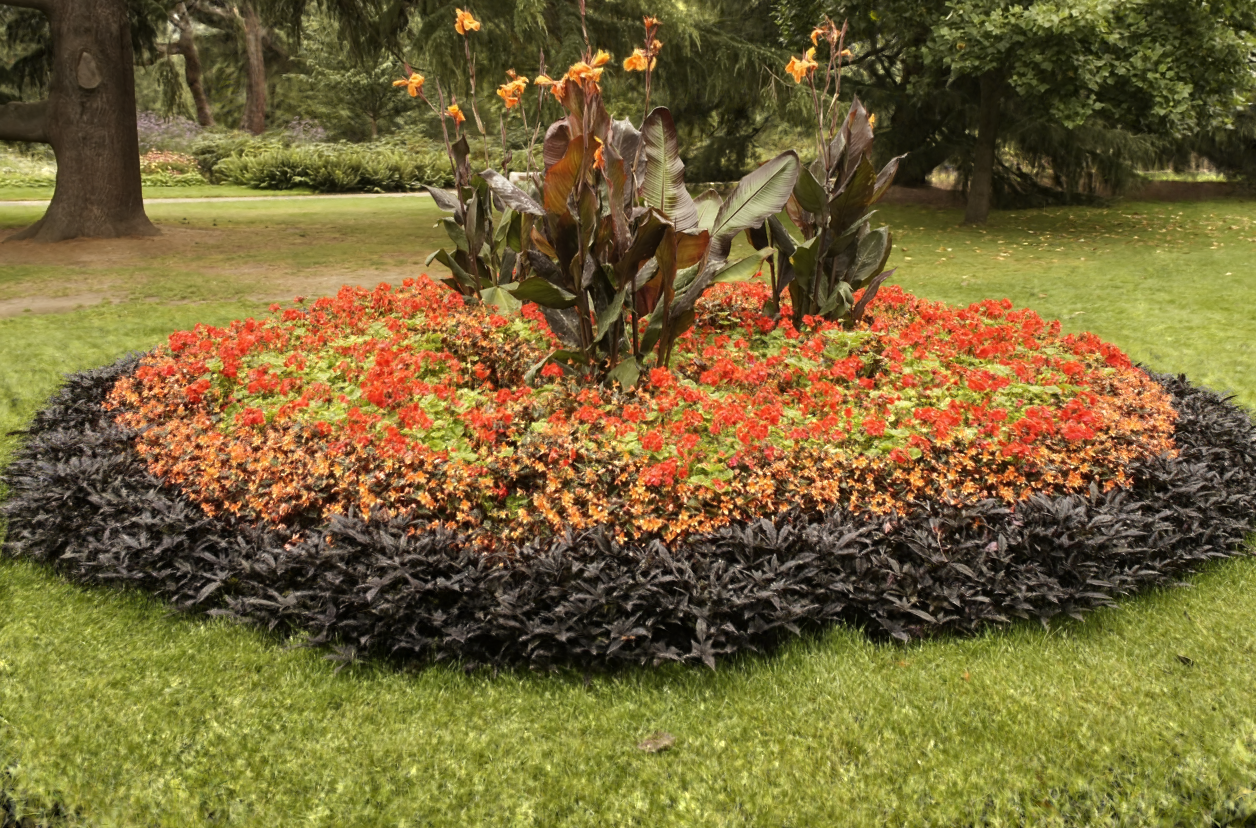}{0.40\mytmplen}{0.32\mytmplen}{0.81\mytmplen}{0.194\mytmplen}{1.2cm}{\mytmplen}{3.5}{red} &
    \zoomin{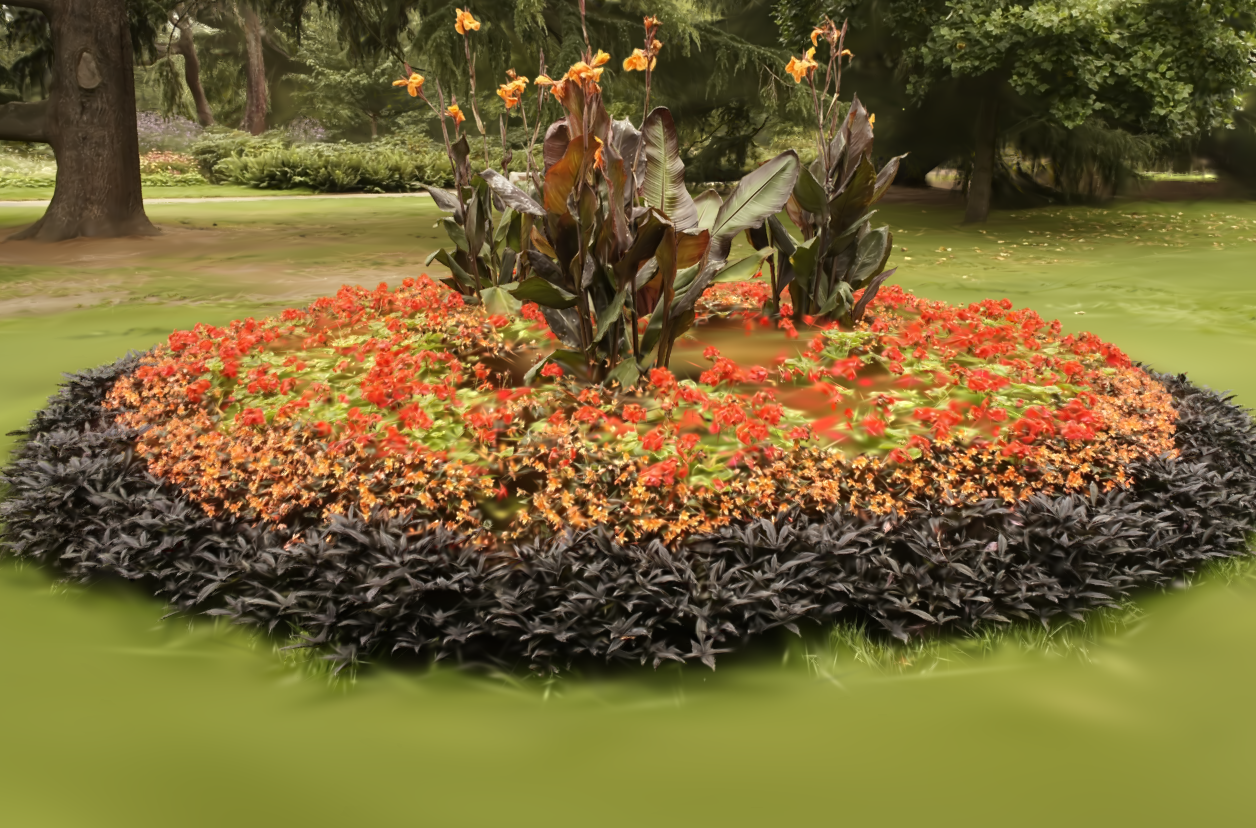}{0.40\mytmplen}{0.32\mytmplen}{0.81\mytmplen}{0.194\mytmplen}{1.2cm}{\mytmplen}{3.5}{red} &
    \zoomin{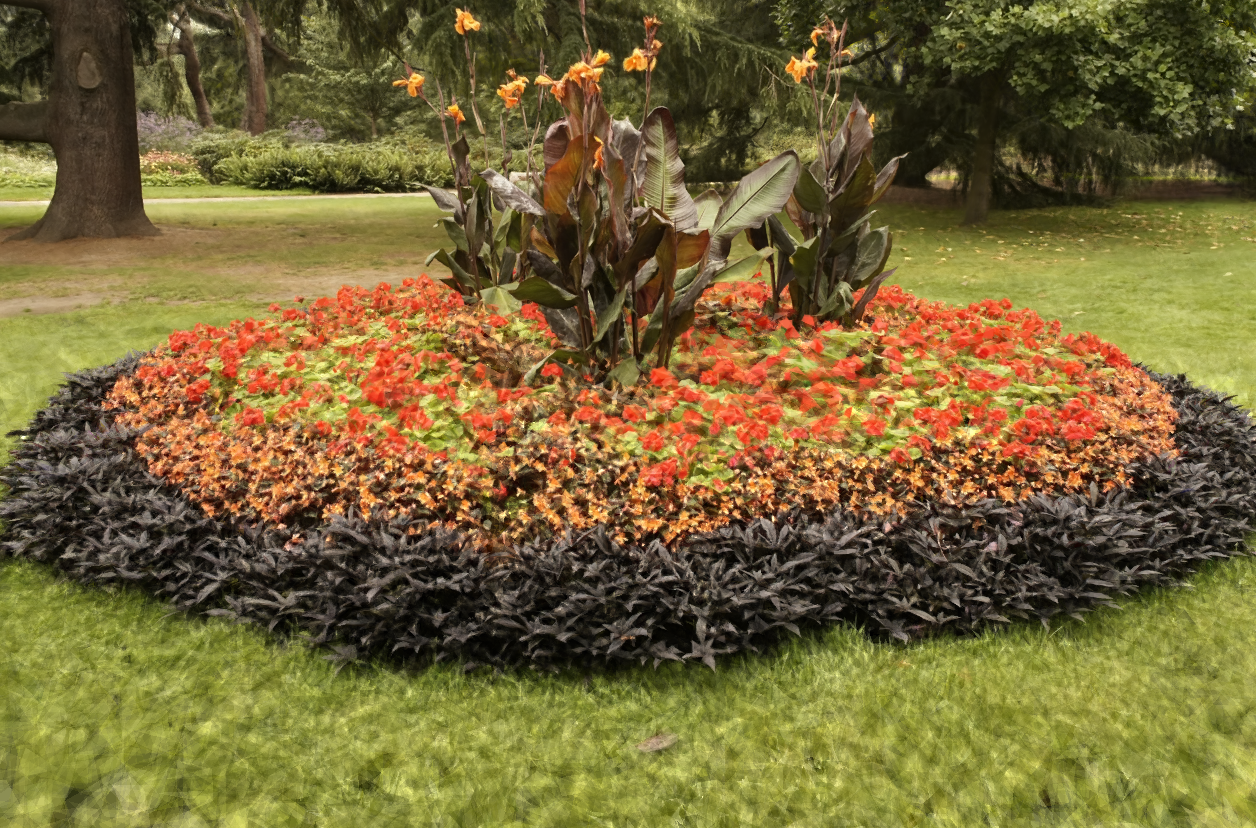}{0.40\mytmplen}{0.32\mytmplen}{0.81\mytmplen}{0.194\mytmplen}{1.2cm}{\mytmplen}{3.5}{red} \\

    \rotatebox{90}{\parbox{2.2cm}{\centering Bicycle}}
    \zoomin{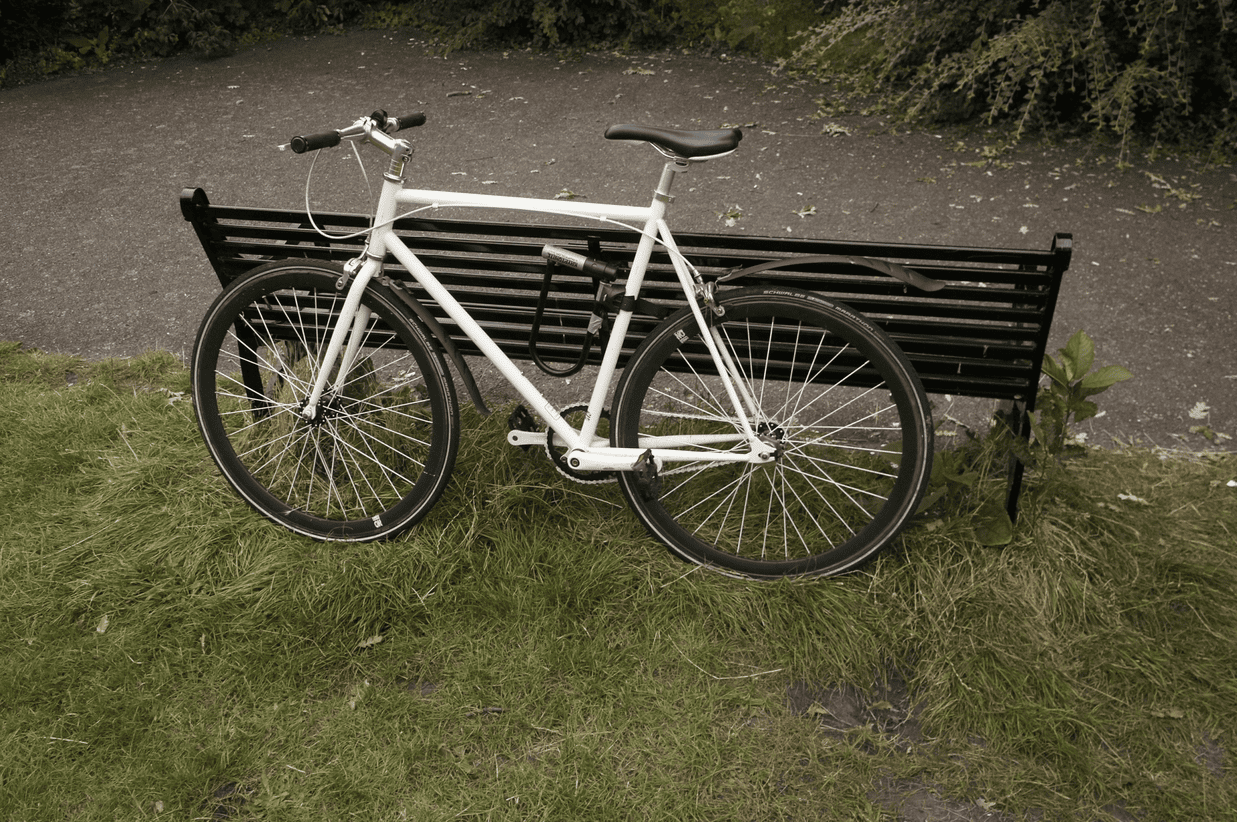}{0.40\mytmplen}{0.32\mytmplen}{0.81\mytmplen}{0.194\mytmplen}{1.2cm}{\mytmplen}{3.5}{red} &
    \zoomin{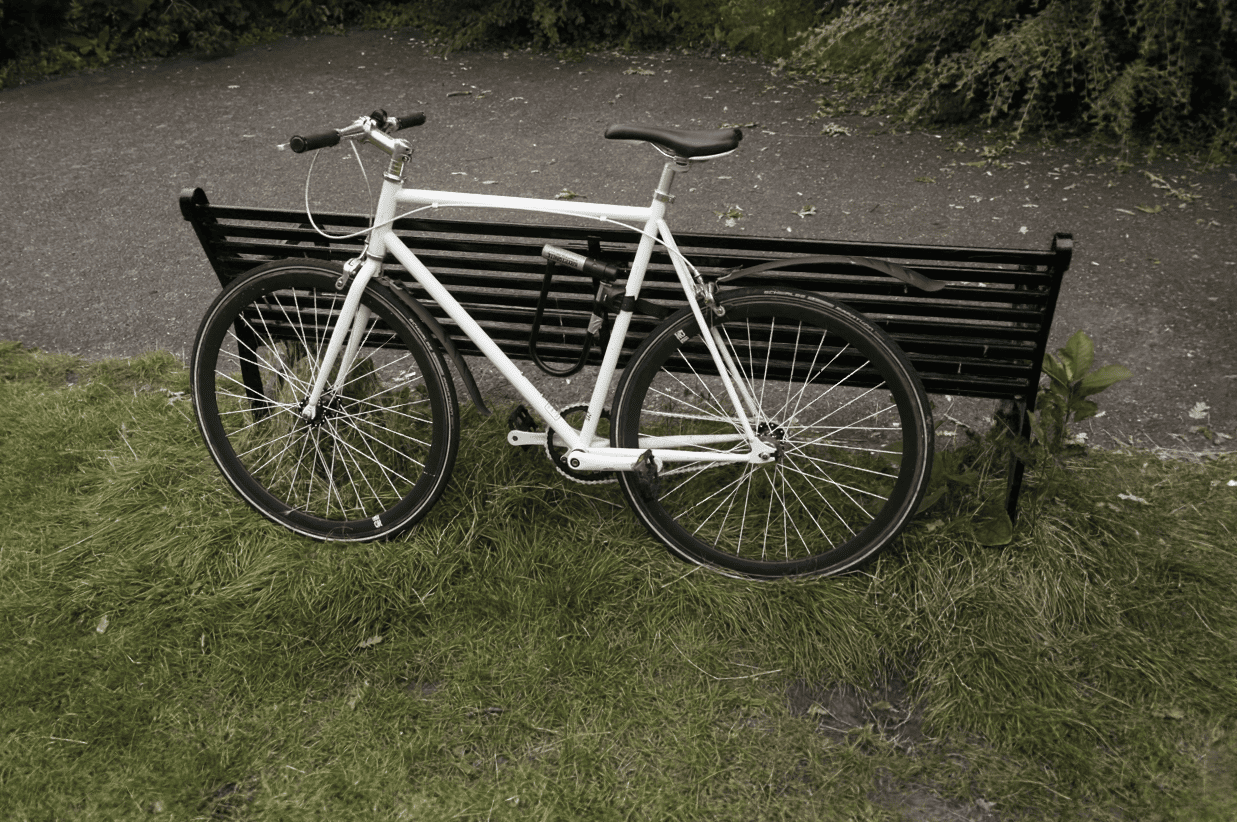}{0.40\mytmplen}{0.32\mytmplen}{0.81\mytmplen}{0.194\mytmplen}{1.2cm}{\mytmplen}{3.5}{red} &
    \zoomin{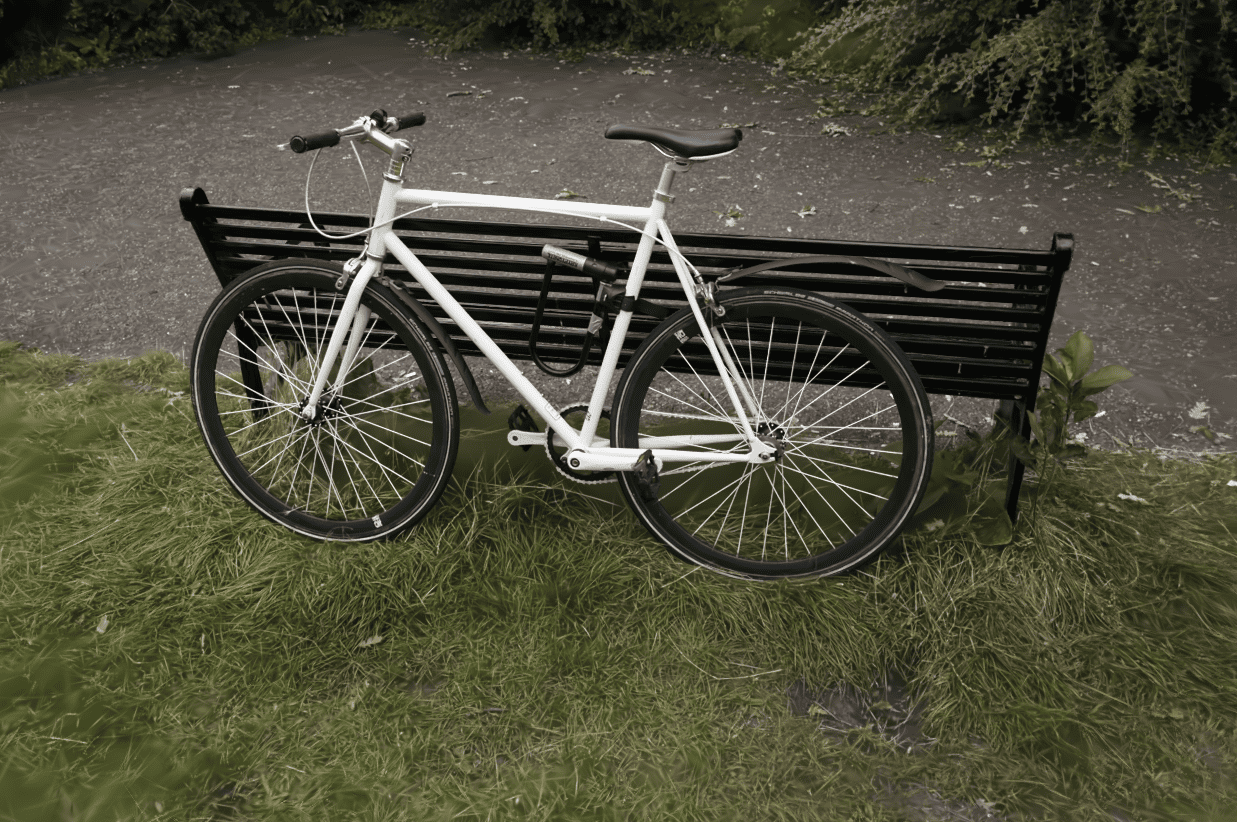}{0.40\mytmplen}{0.32\mytmplen}{0.81\mytmplen}{0.194\mytmplen}{1.2cm}{\mytmplen}{3.5}{red} &
    \zoomin{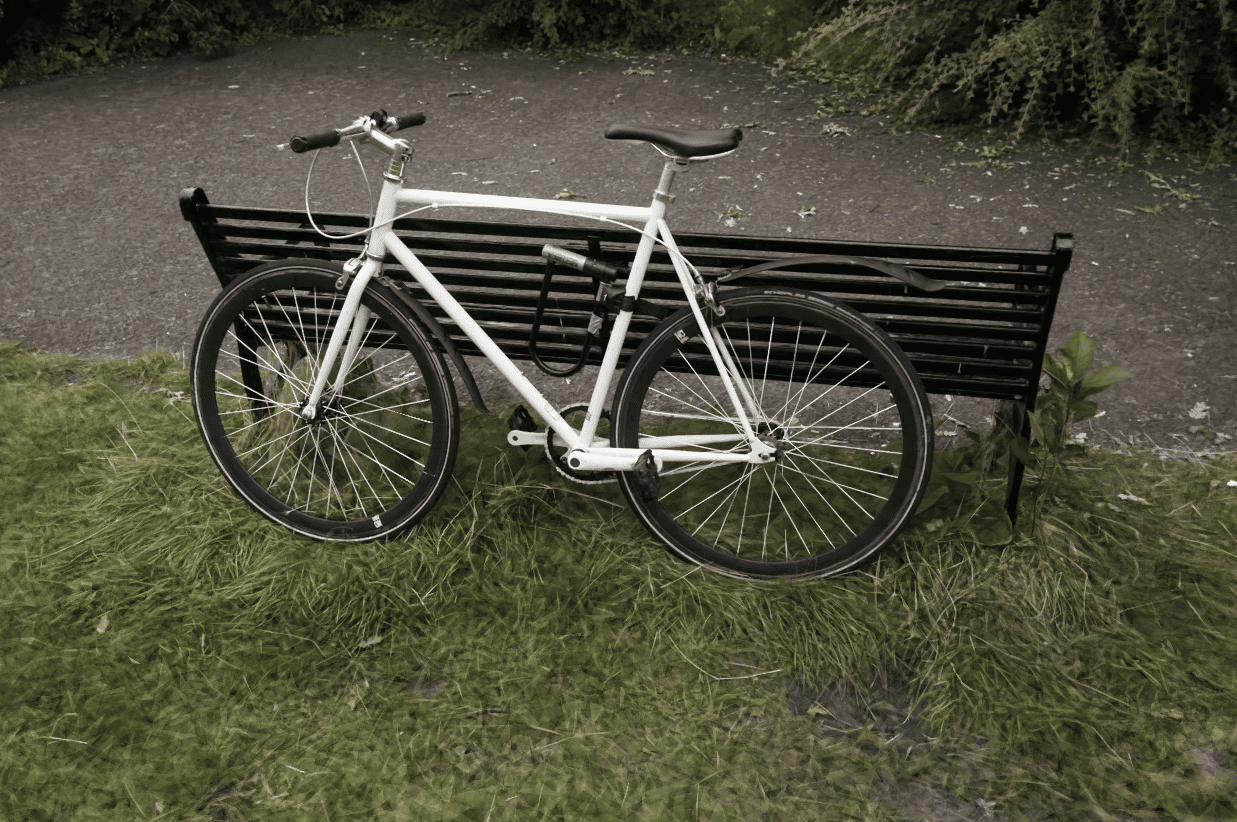}{0.40\mytmplen}{0.32\mytmplen}{0.81\mytmplen}{0.194\mytmplen}{1.2cm}{\mytmplen}{3.5}{red} \\

\end{tabular}
}

\caption{\small \myTitle{Qualitative results}
We visually compare our method to 3DCS~\cite{Held20253DConvex} and 2DGS~\cite{Huang20242DGaussian}.
Triangle Splatting captures finer details and produces more accurate renderings of real-world scenes, with less blurry results than 2DGS, and a higher visual quality than 3DCS~\cite{Held20253DConvex}.
}
\label{fig:qualityresults}
\end{figure*}

\Cref{fig:qualityresults} presents a qualitative comparison between \methodname, 3DCS, and 2DGS\@.
We consistently produce sharper reconstructions, particularly in high-frequency regions.
For instance, in the \texttt{Bicycle} scene, it more accurately captures fine details, as highlighted.

\paragraph{Speed \& Memory.}
\begin{wraptable}{r}{0.4\textwidth}
\vspace*{-1em}
\centering
\resizebox{0.4\textwidth}{!}{
\begin{tabular}{@{}l|ccc}
& Train~$\downarrow$ & FPS~$\uparrow$ & Memory~$\downarrow$  \\
\midrule
ZipNerf &  5h & 0.18 & 569MB \\ 
3DGS     & 42m & 134 & 734MB \\ 
3DCS     & 87m & 25 & 666MB \\
\midrule
BBSplat  & 96m & 25 &  \textbf{175MB} \\
2DGS     & \textbf{29m} & 64 & 484MB  \\
\hline
Ours     & 39m & \textbf{97} & 795MB \\
\end{tabular}
}
\caption{
\myTitle{Speed \& Memory}
\methodname scales efficiently, achieving faster training and rendering despite using more primitives.
\vspace{-1.em}
}
\label{tab:memory}
\end{wraptable}

\Cref{tab:memory} compares the memory consumption and rendering speed of concurrent methods.
Although BBSplat uses fewer primitives than \methodname, it suffers from considerably slower training and slower inference.
\methodname demonstrates strong scalability, despite using more primitives, it renders 4$\times$ faster than BBSplat and achieves a 40\% speedup over 2DGS\@.
Triangle Splatting significantly outperforms 3DCS, achieving 2$\times$ faster training and 4$\times$ faster inference.
Unlike 3DCS, \methodname does not require computing a 2D convex hull and rendering is more efficient.
\methodname computes the signed distance for only three lines per pixel, whereas 3DCS requires calculations for six lines, effectively doubling the per-pixel computational cost.

\section{Ablations}

\begin{wraptable}{r}{0.5\textwidth}
\vspace*{-1em}
\centering
\resizebox{0.5\textwidth}{!}{
\begin{tabular}{@{}l|ccc}
& LPIPS~$\downarrow$ & PSNR~$\uparrow$ & SSIM~$\uparrow$  \\
\midrule
\methodname & \textbf{0.191} &  \textbf{27.14} & \textbf{0.814}  \\
\midrule
w/o $\mathcal{L}_s$ & \textbf{0.191} & 26.97 & 0.812 \\
w/o $\sigma^{-1}$ sampling & 0.193  & 27.03 & 0.811  \\
w/o $o$ sampling &0.193 & 27.02 & 0.811 \\
w/o $\mathcal{L}_d \:\&\: \mathcal{L}_n$ & 0.194 & 27.11 & 0.811  \\
w/o $\mathcal{L}_o$ & 0.207 & 26.38 & 0.794  \\
\end{tabular}
}
\caption{\myTitle{Ablation study} We isolate the impact of each component by removing them individually.
}
\vspace{-0.5em}
\label{tab:ablation}
\end{wraptable}

\paragraph{Loss terms.}

\Cref{tab:ablation} shows the impact on performance when removing different components of our pipeline on Mip-NeRF360. 
The opacity regularization term $\mathcal{L}_o$ is the most impactful, encouraging lower opacity values so that triangles in empty regions become transparent and are eventually reallocated.
The regularization term $\mathcal{L}_s$ encourages larger triangles, significantly increasing PSNR, particularly in indoor scenes.
The initial point cloud is often extremely sparse along walls, frequently with few or no initial triangles.
Without $\mathcal{L}_s$, triangles move too slowly and fail to reach and cover the scene boundaries.
By promoting larger shapes, this regularization enables faster growth, allowing triangles to extend into underrepresented regions and better capture the full structure of the scene.
Sampling based on either $\sigma^{-1}$ or opacity alone yields similar performance, while combining both leads to improved results, especially in outdoor scenes.

\begin{figure}[t]
\setlength\mytmplen{0.48\linewidth}
\centering

\begin{minipage}{\textwidth}
\centering
\begin{minipage}{0.48\linewidth}
\includegraphics[width=\linewidth]{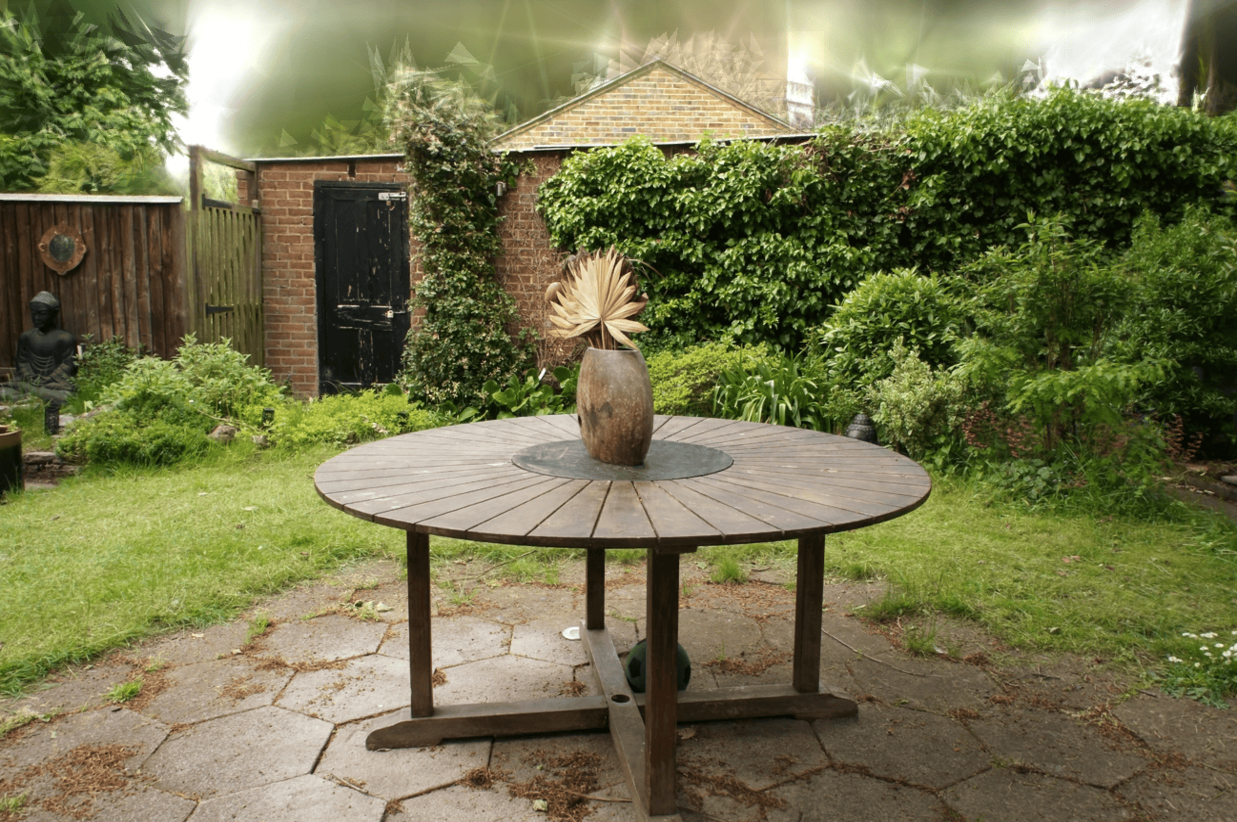}
\end{minipage}\hspace{.01\linewidth}
\begin{minipage}{0.48\linewidth}
\includegraphics[width=\linewidth]{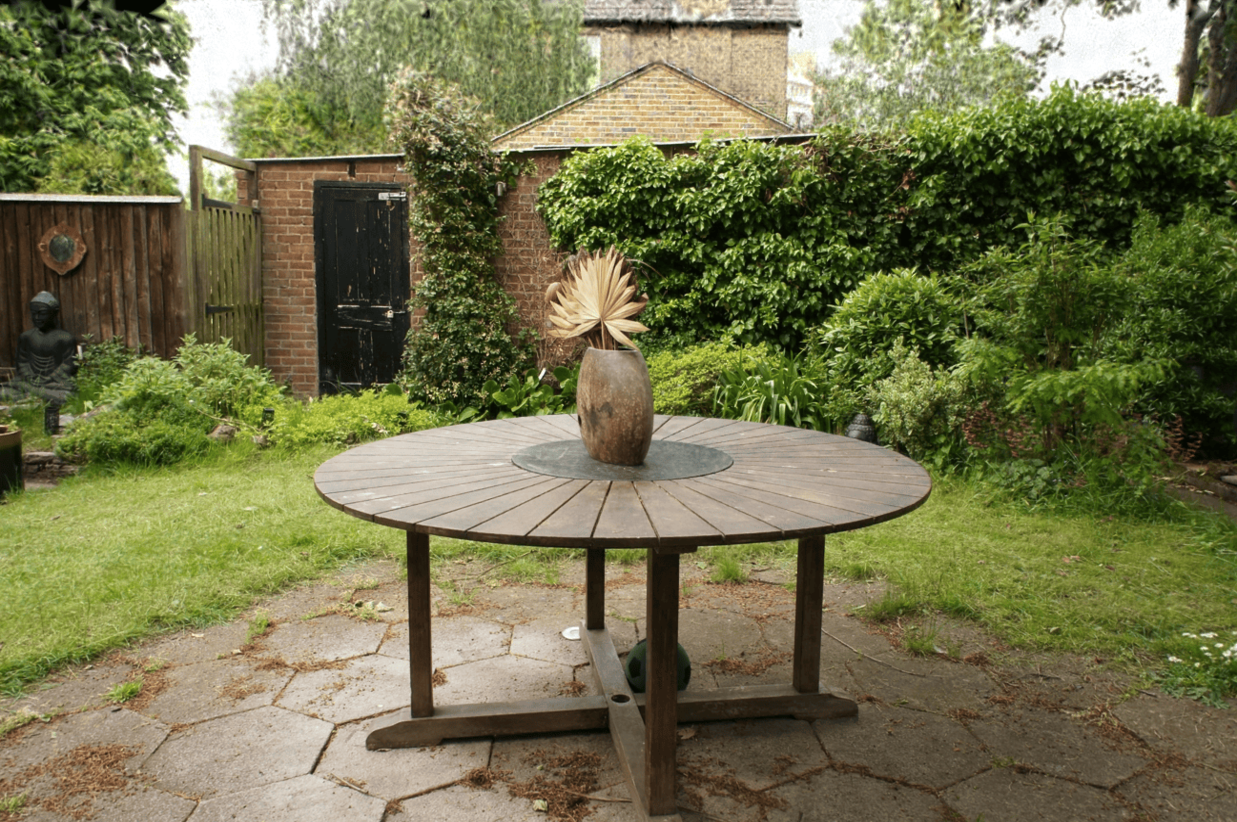}
\end{minipage}
\end{minipage}

\captionof{figure}{\small
\myTitle{Ablation study (window function)}
We compare against the Sigmoid function (left) which fails to recover background regions accurately, while ours doesn't (right).
\label{fig:ablation1}
}

\vspace{.5em}

\begin{minipage}{\textwidth}
\centering
\begin{minipage}{0.48\linewidth}
\zoomin{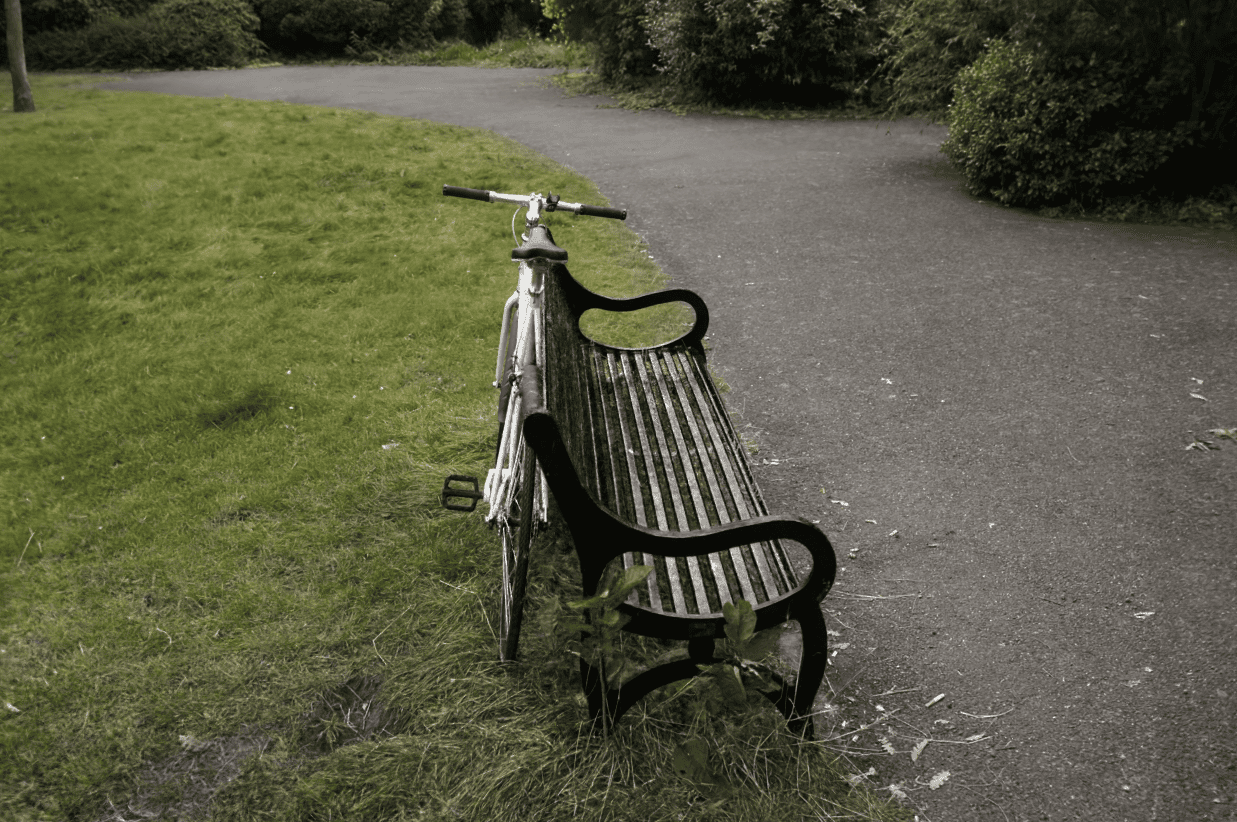}{0.44\mytmplen}{0.53\mytmplen}{0.84\mytmplen}{0.16\mytmplen}{2.0cm}{\mytmplen}{3.5}{red}
\end{minipage}\hspace{.01\linewidth}
\begin{minipage}{0.48\linewidth}
\zoomin{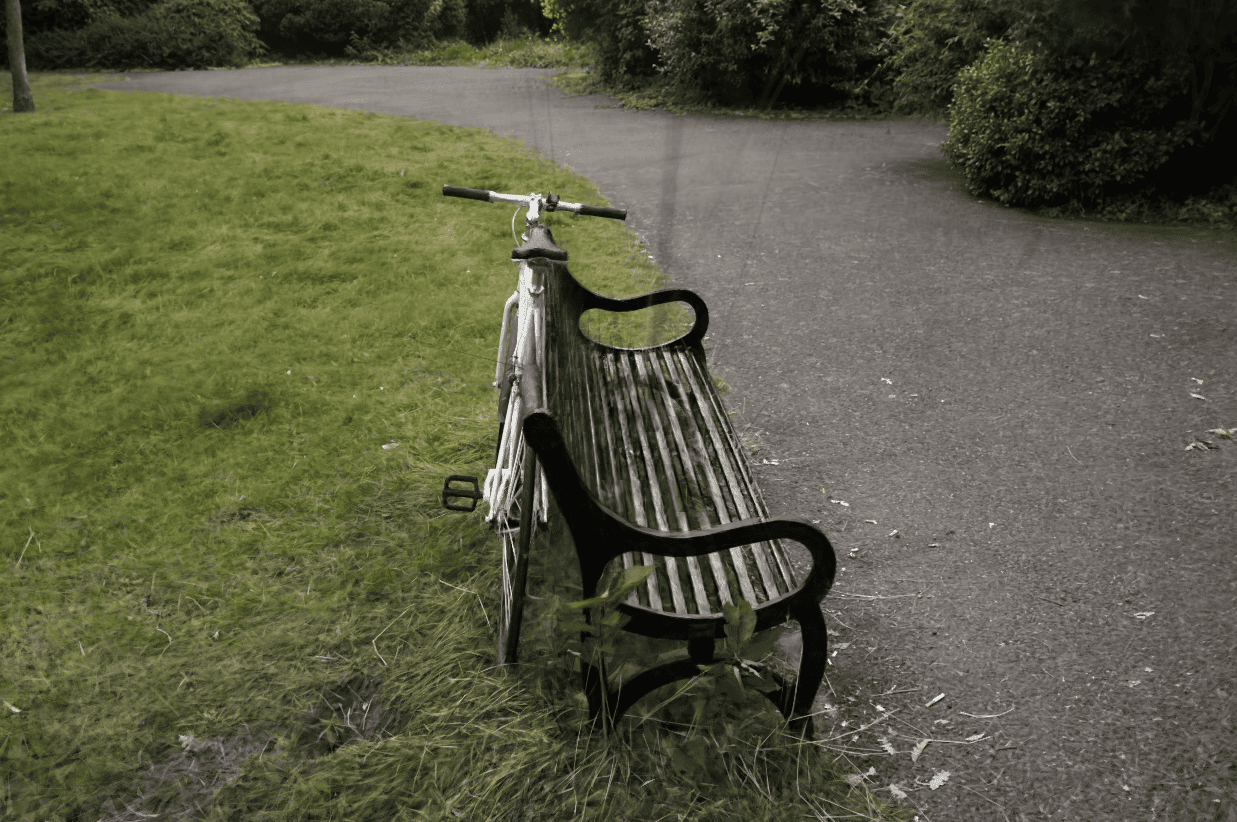}{0.44\mytmplen}{0.53\mytmplen}{0.84\mytmplen}{0.16\mytmplen}{2.0cm}{\mytmplen}{3.5}{red}
\end{minipage}
\end{minipage}
\vspace{-1.2em}
\captionof{figure}{\small
\myTitle{Ablation study (triangles as convexes)}
We compare our method (left) against 3DCS with convexes made of three vertices (right), which results in degenerate geometry, as emphasized in the zoom-ins.
\label{fig:ts_3dcs}
}
\end{figure}

\paragraph{Window functions.}
\Cref{fig:ablation1} highlights the difference between the sigmoid-based window function and the proposed window function.
In regions with sparse initial point cloud, particularly in the background, the sigmoid function fails to recover the scene structure.
Since the sigmoid is not bounded by its geometry's vertices and can grow arbitrarily large, the optimizer tends to increase the sigma values instead of moving vertices to cover empty areas.
This results in small yet very smooth shapes, making them difficult to optimize.
In contrast, our normalized window function enforces spatial bounds, which encourages vertices to move and fill underrepresented regions.
As the size of each shape is explicitly defined, the optimization process becomes more stable and effective.

\paragraph{Triangle \textit{vs.} convex splatting.}
In 3DCS, each convex shape is defined by six 3D vertices.
When the number of vertices is reduced to three, the shape degenerates into a triangle.
In \Cref{fig:ts_3dcs}, we present a visual comparison between \methodname and 3DCS using triangular shapes.
Unlike 3DCS, \methodname does not produce line artifacts, which often appear in 3DCS when handling degenerate triangles.
Furthermore, \methodname obtains a higher visual quality with an improvement of 0.05 LPIPS, 0.61 PSNR and 0.045 SSIM on Mip-NeRF360.

\paragraph{Rendering speed with traditional mesh-based renderer.}
\begin{wraptable}{r}{0.6\textwidth}
\vspace*{-1em}
\centering
\resizebox{0.58\textwidth}{!}{
\begin{tabular}{l|ccccc}
\textbf{Hardware} & \textbf{OS} & \textbf{TFLOPS} & \textbf{HD} & \textbf{Full HD} & \textbf{4k} \\
\midrule
MacBook M4 & MacOS & 8 & 500 & 370  &  160 \\
RTX5000 & Windows & 11 & 570 & 380 & 290  \\
RTX4090 & Linux & 48  & 2,400 & 1900 & 1050   \\
\end{tabular}
}
\caption{\small \myTitle{FPS for different hardwares and resolutions} Evaluated on \textit{Garden} ($\approx$ 2M triangles). OS stands for operating system. 
}
\label{tab:rendering_fps}
\vspace{-1.3em}
\end{wraptable}

By annealing both opacity and $\sigma$ during training and setting SH to 0, the representation gradually converges to solid triangles by the end of optimization.
The final triangle soup can be seamlessly integrated into \textit{any} mesh-based renderer.
This represents a significant advancement over 3DGS: while preserving the benefits of differentiable training, our triangle-based representation is natively compatible with game engines. 
As shown in \Cref{tab:rendering_fps}, our method achieves 500 FPS at HD resolution on a consumer laptop and 2,400 FPS on an RTX 4090 within a game engine, demonstrating both efficiency and practical usability.
Future work could explore training strategies specifically optimized for game engine deployment, as our current setup focuses on novel-view synthesis without targeting game engine visual quality. 
This opens new possibilities for integrating radiance field directly into AR/VR and gaming pipelines.

\section{Conclusions}%
\label{sec:conclusion}

We have introduced \textbf{Triangle Splatting}, a novel differentiable rendering technique that directly optimizes unstructured triangle primitives for novel-view synthesis.
By leveraging the same primitive used in classical mesh representations, our method bridges the gap between neural rendering and traditional graphics pipelines. 
Triangle Splatting offers a compelling alternative to volumetric and implicit methods, achieving high visual fidelity with faster rendering performance. 
These results establish Triangle Splatting as a promising step toward mesh-aware neural rendering, unifying decades of GPU-accelerated graphics with modern differentiable frameworks.

\paragraph{Limitations.}
Our triangle soups can already be rendered directly in any standard mesh-based renderer. However, generating a connected mesh still requires additional steps.
A promising direction for future work is to develop meshing strategies that fully leverage the triangle-based nature of our representation.
For completeness, we applied the 2D Gaussian Splatting meshing approach and include the results in the supplementary material. While this demonstrates compatibility with prior work, we believe future work should focus on more direct and principled meshing techniques that capitalize on the explicit triangle structure of our representation.
While the results on outdoor scenes are promising, our method occasionally suffers from floaters. In large-scale outdoor scenes, volumetric shapes receive stronger training supervision during optimization, as they are visible from a greater number of viewpoints. 
In contrast, non-volumetric shapes are observed from fewer angles and can become overfitted to specific training views, leading to the emergence of floaters.

\mysection{Acknowledgments.}
J. Held, A. Deliege and A. Cioppa are funded by the F.R.S.-FNRS. The research reported in this publication was supported by funding from KAUST Center of Excellence on GenAI, under award number 5940. This work was also supported by KAUST Ibn Rushd Postdoc Fellowship program. The present research benefited from computational resources made available on Lucia, the Tier-1 supercomputer of the Walloon Region, infrastructure funded by the Walloon Region under the grant agreement n°1910247.

\clearpage
\appendix
\clearpage

\begin{thebibliography}{43}
\providecommand{\natexlab}[1]{#1}
\providecommand{\url}[1]{\texttt{#1}}
\expandafter\ifx\csname urlstyle\endcsname\relax
  \providecommand{\doi}[1]{doi: #1}\else
  \providecommand{\doi}{doi: \begingroup \urlstyle{rm}\Url}\fi

\bibitem[Barron et~al.(2021)Barron, Mildenhall, Tancik, Hedman, Martin-Brualla,
  and Srinivasan]{Barron2021MipNeRF}
Jonathan~T. Barron, Ben Mildenhall, Matthew Tancik, Peter Hedman, Ricardo
  Martin-Brualla, and Pratul~P. Srinivasan.
\newblock Mip-{NeRF}: A multiscale representation for anti-aliasing neural
  radiance fields.
\newblock In \emph{IEEE/CVF Int. Conf. Comput. Vis. (ICCV)}, pages 5835--5844,
  Montr{\'e}al, Can., Oct. 2021. Inst. Electr. Electron. Eng. (IEEE).
\newblock URL \url{https://doi.org/10.1109/ICCV48922.2021.00580}.

\bibitem[Barron et~al.(2022)Barron, Mildenhall, Verbin, Srinivasan, and
  Hedman]{Barron2022MipNeRF360}
Jonathan~T. Barron, Ben Mildenhall, Dor Verbin, Pratul~P. Srinivasan, and Peter
  Hedman.
\newblock Mip-{NeRF} 360: Unbounded anti-aliased neural radiance fields.
\newblock In \emph{IEEE/CVF Conf. Comput. Vis. Pattern Recognit. (CVPR)}, pages
  5460--5469, New Orleans, LA, USA, Jun. 2022. Inst. Electr. Electron. Eng.
  (IEEE).
\newblock URL \url{https://doi.org/10.1109/CVPR52688.2022.00539}.

\bibitem[Barron et~al.(2023)Barron, Mildenhall, Verbin, Srinivasan, and
  Hedman]{Barron2023ZipNeRF}
Jonathan~T. Barron, Ben Mildenhall, Dor Verbin, Pratul~P. Srinivasan, and Peter
  Hedman.
\newblock Zip-{NeRF}: Anti-aliased grid-based neural radiance fields.
\newblock In \emph{IEEE/CVF Int. Conf. Comput. Vis. (ICCV)}, pages
  19640--19648, Paris, Fr., Oct. 2023. Inst. Electr. Electron. Eng. (IEEE).
\newblock URL \url{https://doi.org/10.1109/ICCV51070.2023.01804}.

\bibitem[Chan et~al.(2022)Chan, Lin, Chan, Nagano, Pan, de~Mello, Gallo,
  Guibas, Tremblay, Khamis, Karras, and Wetzstein]{Chan2022Efficient}
Eric~R. Chan, Connor~Z. Lin, Matthew~A. Chan, Koki Nagano, Boxiao Pan, Shalini
  de~Mello, Orazio Gallo, Leonidas Guibas, Jonathan Tremblay, Sameh Khamis,
  Tero Karras, and Gordon Wetzstein.
\newblock Efficient geometry-aware {3D} generative adversarial networks.
\newblock In \emph{IEEE/CVF Conf. Comput. Vis. Pattern Recognit. (CVPR)}, pages
  16102--16112, New Orleans, LA, USA, Jun. 2022. Inst. Electr. Electron. Eng.
  (IEEE).
\newblock URL \url{https://doi.org/10.1109/CVPR52688.2022.01565}.

\bibitem[Chen et~al.(2022)Chen, Xu, Geiger, Yu, and Su]{Chen2022TensoRF}
Anpei Chen, Zexiang Xu, Andreas Geiger, Jingyi Yu, and Hao Su.
\newblock {TensoRF}: Tensorial radiance fields.
\newblock In \emph{Eur. Conf. Comput. Vis. (ECCV)}, volume 13692 of \emph{Lect.
  Notes Comput. Sci.}, pages 333--350, Tel Aviv, Isra{\"e}l, 2022. Springer
  Nat. Switz.
\newblock URL \url{https://doi.org/10.1007/978-3-031-19824-3_20}.

\bibitem[Chen et~al.(2024)Chen, Chen, Qu, Wang, Liu, Chen, and
  Chung]{Chen2024Beyond-arxiv}
Haodong Chen, Runnan Chen, Qiang Qu, Zhaoqing Wang, Tongliang Liu, Xiaoming
  Chen, and Yuk~Ying Chung.
\newblock Beyond {Gaussians}: Fast and high-fidelity {3D} splatting with linear
  kernels.
\newblock \emph{arXiv}, abs/2411.12440:\penalty0 1--14, 2024.
\newblock URL \url{https://doi.org/10.48550/arXiv.2411.12440}.

\bibitem[Chen et~al.(2023)Chen, Funkhouser, Hedman, and
  Tagliasacchi]{Chen2023MobileNeRF}
Zhiqin Chen, Thomas Funkhouser, Peter Hedman, and Andrea Tagliasacchi.
\newblock {MobileNeRF}: Exploiting the polygon rasterization pipeline for
  efficient neural field rendering on mobile architectures.
\newblock In \emph{IEEE/CVF Conf. Comput. Vis. Pattern Recognit. (CVPR)}, pages
  16569--16578, Vancouver, Can., Jun. 2023. IEEE.
\newblock URL \url{https://doi.org/10.1109/CVPR52729.2023.01590}.

\bibitem[Deng et~al.(2020)Deng, Genova, Yazdani, Bouaziz, Hinton, and
  Tagliasacchi]{Deng2020CvxNet}
Boyang Deng, Kyle Genova, Soroosh Yazdani, Sofien Bouaziz, Geoffrey Hinton, and
  Andrea Tagliasacchi.
\newblock {CvxNet}: Learnable convex decomposition.
\newblock In \emph{IEEE/CVF Conf. Comput. Vis. Pattern Recognit. (CVPR)}, pages
  31--41, Seattle, WA, USA, Jun. 2020. Inst. Electr. Electron. Eng. (IEEE).
\newblock URL \url{https://doi.org/10.1109/CVPR42600.2020.00011}.

\bibitem[Du et~al.(2023)Du, Smith, Tewari, and Sitzmann]{Du2023Learning}
Yilun Du, Cameron Smith, Ayush Tewari, and Vincent Sitzmann.
\newblock Learning to render novel views from wide-baseline stereo pairs.
\newblock In \emph{IEEE/CVF Conf. Comput. Vis. Pattern Recognit. (CVPR)}, pages
  4970--4980, Vancouver, Can., Jun. 2023. Inst. Electr. Electron. Eng. (IEEE).
\newblock URL \url{https://doi.org/10.1109/CVPR52729.2023.00481}.

\bibitem[Fridovich-Keil et~al.(2022)Fridovich-Keil, Yu, Tancik, Chen, Recht,
  and Kanazawa]{FridovichKeil2022Plenoxels}
Sara Fridovich-Keil, Alex Yu, Matthew Tancik, Qinhong Chen, Benjamin Recht, and
  Angjoo Kanazawa.
\newblock Plenoxels: Radiance fields without neural networks.
\newblock In \emph{IEEE/CVF Conf. Comput. Vis. Pattern Recognit. (CVPR)}, pages
  5491--5500, New Orleans, LA, USA, Jun. 2022. Inst. Electr. Electron. Eng.
  (IEEE).
\newblock URL \url{https://doi.org/10.1109/CVPR52688.2022.00542}.

\bibitem[Gross and Pfister(2007)]{Gross2007Point}
Markus Gross and Hanspeter Pfister.
\newblock \emph{Point-Based Graphics}.
\newblock Morgan Kauffmann Publ. Inc., San Francisco, CA, USA, Jun. 2007.
\newblock URL \url{http://dx.doi.org/10.1016/B978-0-12-370604-1.X5000-7}.

\bibitem[Gu{\'e}don and Lepetit(2024)]{Guedon2024SuGaR}
Antoine Gu{\'e}don and Vincent Lepetit.
\newblock {SuGaR}: Surface-aligned {Gaussian} splatting for efficient {3D} mesh
  reconstruction and high-quality mesh rendering.
\newblock In \emph{IEEE/CVF Conf. Comput. Vis. Pattern Recognit. (CVPR)}, pages
  5354--5363, Seattle, WA, USA, Jun. 2024. Inst. Electr. Electron. Eng. (IEEE).
\newblock URL \url{https://doi.org/10.1109/CVPR52733.2024.00512}.

\bibitem[Hedman et~al.(2021)Hedman, Srinivasan, Mildenhall, Barron, and
  Debevec]{Hedman2021Baking}
Peter Hedman, Pratul~P. Srinivasan, Ben Mildenhall, Jonathan~T. Barron, and
  Paul Debevec.
\newblock Baking neural radiance fields for real-time view synthesis.
\newblock In \emph{IEEE/CVF Int. Conf. Comput. Vis. (ICCV)}, pages 5855--5864,
  Montr{\'e}al, Can., Oct. 2021. Inst. Electr. Electron. Eng. (IEEE).
\newblock URL \url{https://doi.org/10.1109/ICCV48922.2021.00582}.

\bibitem[Held et~al.(2025)Held, Vandeghen, Hamdi, Deli{\`e}ge, Cioppa,
  Giancola, Vedaldi, Ghanem, and Van~Droogenbroeck]{Held20253DConvex}
Jan Held, Renaud Vandeghen, Abdullah Hamdi, Adrien Deli{\`e}ge, Anthony Cioppa,
  Silvio Giancola, Andrea Vedaldi, Bernard Ghanem, and Marc Van~Droogenbroeck.
\newblock {3D} convex splatting: Radiance field rendering with {3D} smooth
  convexes.
\newblock In \emph{IEEE/CVF Conf. Comput. Vis. Pattern Recognit. (CVPR)}, pages
  1--10, Nashville, TN, USA, Jun. 2025. Inst. Electr. Electron. Eng. (IEEE).

\bibitem[Huang et~al.(2024{\natexlab{a}})Huang, Yu, Chen, Geiger, and
  Gao]{Huang20242DGaussian}
Binbin Huang, Zehao Yu, Anpei Chen, Andreas Geiger, and Shenghua Gao.
\newblock {2D} {Gaussian} splatting for geometrically accurate radiance fields.
\newblock In \emph{ACM SIGGRAPH Conf. Pap.}, volume~35, pages 1--11, Denver,
  CO, USA, Jul. 2024{\natexlab{a}}. ACM.
\newblock URL \url{https://doi.org/10.1145/3641519.3657428}.

\bibitem[Huang et~al.(2024{\natexlab{b}})Huang, Lin, Sun, Yang, Lyu, Cao, and
  Qi]{Huang2024Deformable-arxiv}
Yi-Hua Huang, Ming-Xian Lin, Yang-Tian Sun, Ziyi Yang, Xiaoyang Lyu, Yan-Pei
  Cao, and Xiaojuan Qi.
\newblock Deformable radial kernel splatting.
\newblock \emph{arXiv}, abs/2412.11752, 2024{\natexlab{b}}.
\newblock URL \url{https://doi.org/10.48550/arXiv.2412.11752}.

\bibitem[Jain et~al.(2021)Jain, Tancik, and Abbeel]{Jain2021Putting}
Ajay Jain, Matthew Tancik, and Pieter Abbeel.
\newblock Putting {NeRF} on a diet: Semantically consistent few-shot view
  synthesis.
\newblock In \emph{IEEE/CVF Int. Conf. Comput. Vis. (ICCV)}, pages 5865--5874,
  Montr{\'e}al, Can., Oct. 2021. Inst. Electr. Electron. Eng. (IEEE).
\newblock URL \url{https://doi.org/10.1109/ICCV48922.2021.00583}.

\bibitem[Jensen et~al.(2014)Jensen, Dahl, Vogiatzis, Tola, and
  Aanaes]{Jensen2014Large}
Rasmus Jensen, Anders Dahl, George Vogiatzis, Engil Tola, and Henrik Aanaes.
\newblock Large scale multi-view stereopsis evaluation.
\newblock In \emph{IEEE Conf. Comput. Vis. Pattern Recognit. (CVPR)}, pages
  406--413, Columbus, OH, USA, Jun. 2014. Inst. Electr. Electron. Eng. (IEEE).
\newblock URL \url{https://doi.org/10.1109/CVPR.2014.59}.

\bibitem[Kato et~al.(2018)Kato, Ushiku, and Harada]{Kato2018Neural}
Hiroharu Kato, Yoshitaka Ushiku, and Tatsuya Harada.
\newblock Neural {3D} mesh renderer.
\newblock In \emph{IEEE/CVF Conf. Comput. Vis. Pattern Recognit. (CVPR)}, pages
  3907--3916, Salt Lake City, UT, USA, Jun. 2018. Inst. Electr. Electron. Eng.
  (IEEE).
\newblock URL \url{https://doi.org/10.1109/CVPR.2018.00411}.

\bibitem[Kazhdan et~al.(2006)Kazhdan, Bolitho, and Hoppe]{Kazhdan2006Poisson}
Michael Kazhdan, Matthew Bolitho, and Hugues Hoppe.
\newblock Poisson surface reconstruction.
\newblock In \emph{Symp. Geom. Process.}, pages 61--70, Cagliari, Italy, Jun.
  2006. Eurographics Assoc.
\newblock URL \url{http://diglib.eg.org/handle/10.2312/SGP.SGP06.061-070}.

\bibitem[Kerbl et~al.(2023)Kerbl, Kopanas, Leimkuehler, and
  Drettakis]{Kerbl20233DGaussian}
Bernhard Kerbl, Georgios Kopanas, Thomas Leimkuehler, and George Drettakis.
\newblock {3D} {Gaussian} splatting for real-time radiance field rendering.
\newblock \emph{ACM Trans. Graph.}, 42\penalty0 (4):\penalty0 1--14, Jul. 2023.
\newblock URL \url{https://doi.org/10.1145/3592433}.

\bibitem[Kheradmand et~al.(2024)Kheradmand, Rebain, Sharma, Sun, Tseng, Isack,
  Kar, Tagliasacchi, and Yi]{Kheradmand20243DGaussian}
Shakiba Kheradmand, Daniel Rebain, Gopal Sharma, Weiwei Sun, Jeff Tseng, Hossam
  Isack, Abhishek Kar, Andrea Tagliasacchi, and Kwang~Moo Yi.
\newblock {3D} {Gaussian} splatting as {Markov} chain {Monte} {Carlo}.
\newblock In \emph{Adv. Neural Inf. Process. Syst. (NeurIPS)}, volume~37, pages
  80965--80986, Vancouver, Can., Dec. 2024. Curran Assoc. Inc.
\newblock URL \url{https://neurips.cc/virtual/2024/poster/94984}.

\bibitem[Knapitsch et~al.(2017)Knapitsch, Park, Zhou, and
  Koltun]{Knapitsch2017Tanks}
Arno Knapitsch, Jaesik Park, Qian-Yi Zhou, and Vladlen Koltun.
\newblock Tanks and temples: benchmarking large-scale scene reconstruction.
\newblock \emph{ACM Trans. Graph.}, 36\penalty0 (4):\penalty0 1--13, Jul. 2017.
\newblock URL \url{https://doi.org/10.1145/3072959.3073599}.

\bibitem[Kulhanek and Sattler(2023)]{Kulhanek2023TetraNeRF}
Jonas Kulhanek and Torsten Sattler.
\newblock Tetra-{NeRF}: Representing neural radiance fields using tetrahedra.
\newblock In \emph{IEEE/CVF Int. Conf. Comput. Vis. (ICCV)}, pages
  18412--18423, Paris, Fr., Oct. 2023. Inst. Electr. Electron. Eng. (IEEE).
\newblock URL \url{https://doi.org/10.1109/ICCV51070.2023.01692}.

\bibitem[Liu et~al.(2025)Liu, Sun, Chen, Wang, and
  Feng]{Liu2025Deformable-arxiv}
Rong Liu, Dylan Sun, Meida Chen, Yue Wang, and Andrew Feng.
\newblock Deformable beta splatting.
\newblock \emph{arXiv}, abs/2501.18630, 2025.
\newblock URL \url{https://doi.org/10.48550/arXiv.2501.18630}.

\bibitem[Liu et~al.(2019)Liu, Chen, Li, and Li]{Liu2019SoftRasterizer}
Shichen Liu, Weikai Chen, Tianye Li, and Hao Li.
\newblock Soft rasterizer: A differentiable renderer for image-based {3D}
  reasoning.
\newblock In \emph{IEEE/CVF Int. Conf. Comput. Vis. (ICCV)}, pages 7707--7716,
  Seoul, South Korea, Oct. 2019. Inst. Electr. Electron. Eng. (IEEE).
\newblock URL \url{https://doi.org/10.1109/ICCV.2019.00780}.

\bibitem[Loper and Black(2014)]{Loper2014OpenDR}
Matthew~M. Loper and Michael~J. Black.
\newblock {OpenDR}: An approximate differentiable renderer.
\newblock In \emph{Eur. Conf. Comput. Vis. (ECCV)}, volume 8695 of \emph{Lect.
  Notes Comput. Sci.}, pages 154--169, Z{\"u}rich, Switzerland, 2014. Springer
  Int. Publ.
\newblock URL \url{https://doi.org/10.1007/978-3-319-10584-0_11}.

\bibitem[Luiten et~al.(2024)Luiten, Kopanas, Leibe, and
  Ramanan]{Luiten2024Dynamic}
Jonathon Luiten, Georgios Kopanas, Bastian Leibe, and Deva Ramanan.
\newblock Dynamic {3D} {Gaussian}s: Tracking by persistent dynamic view
  synthesis.
\newblock In \emph{Int. Conf. 3D Vis. (3DV)}, pages 800--809, Davos,
  Switzerland, Mar. 2024. Inst. Electr. Electron. Eng. (IEEE).
\newblock URL \url{https://doi.org/10.1109/3DV62453.2024.00044}.

\bibitem[Mai et~al.(2024)Mai, Hedman, Kopanas, Verbin, Futschik, Xu, Kuester,
  Barron, and Zhang]{Mai2024EVER-arxiv}
Alexander Mai, Peter Hedman, George Kopanas, Dor Verbin, David Futschik,
  Qiangeng Xu, Falko Kuester, Jonathan~T. Barron, and Yinda Zhang.
\newblock {EVER}: Exact volumetric ellipsoid rendering for real-time view
  synthesis.
\newblock \emph{arXiv}, abs/2410.01804, 2024.
\newblock URL \url{https://doi.org/10.48550/arXiv.2410.01804}.

\bibitem[Mildenhall et~al.(2020)Mildenhall, Srinivasan, Tancik, Barron,
  Ramamoorthi, and Ng]{Mildenhall2020NeRF-eccv}
Ben Mildenhall, Pratul~P. Srinivasan, Matthew Tancik, Jonathan~T. Barron, Ravi
  Ramamoorthi, and Ren Ng.
\newblock {NeRF}: Representing scenes as neural radiance fields for view
  synthesis.
\newblock In \emph{Eur. Conf. Comput. Vis. (ECCV)}, volume 12346 of \emph{Lect.
  Notes Comput. Sci.}, pages 405--421, Virtual conference, 2020. Springer Int.
  Publ.
\newblock URL \url{https://doi.org/10.1007/978-3-030-58452-8_24}.

\bibitem[Mildenhall et~al.(2021)Mildenhall, Srinivasan, Tancik, Barron,
  Ramamoorthi, and Ng]{Mildenhall2021NeRF}
Ben Mildenhall, Pratul~P. Srinivasan, Matthew Tancik, Jonathan~T. Barron, Ravi
  Ramamoorthi, and Ren Ng.
\newblock {NeRF}.
\newblock \emph{Commun. ACM}, 65\penalty0 (1):\penalty0 99--106, Dec. 2021.
\newblock URL \url{https://doi.org/10.1145/3503250}.

\bibitem[M{\"u}ller et~al.(2022)M{\"u}ller, Evans, Schied, and
  Keller]{Muller2022Instant}
Thomas M{\"u}ller, Alex Evans, Christoph Schied, and Alexander Keller.
\newblock Instant neural graphics primitives with a multiresolution hash
  encoding.
\newblock \emph{ACM Trans. Graph.}, 41\penalty0 (4):\penalty0 1--15, Jul. 2022.
\newblock URL \url{https://doi.org/10.1145/3528223.3530127}.

\bibitem[Radl et~al.(2024)Radl, Steiner, Parger, Weinrauch, Kerbl, and
  Steinberger]{Radl2024StopThePop}
Lukas Radl, Michael Steiner, Mathias Parger, Alexander Weinrauch, Bernhard
  Kerbl, and Markus Steinberger.
\newblock {StopThePop}: Sorted {Gaussian} splatting for view-consistent
  real-time rendering.
\newblock \emph{ACM Trans. Graph.}, 43\penalty0 (4):\penalty0 1--17, Jul. 2024.
\newblock URL \url{https://doi.org/10.1145/3658187}.

\bibitem[Reiser et~al.(2021)Reiser, Peng, Liao, and Geiger]{Reiser2021KiloNeRF}
Christian Reiser, Songyou Peng, Yiyi Liao, and Andreas Geiger.
\newblock {KiloNeRF}: Speeding up neural radiance fields with thousands of tiny
  {MLPs}.
\newblock In \emph{IEEE/CVF Int. Conf. Comput. Vis. (ICCV)}, pages
  14315--14325, Montr{\'e}al, Can., Oct. 2021. Inst. Electr. Electron. Eng.
  (IEEE).
\newblock URL \url{https://doi.org/10.1109/ICCV48922.2021.01407}.

\bibitem[Reiser et~al.(2023)Reiser, Szeliski, Verbin, Srinivasan, Mildenhall,
  Geiger, Barron, and Hedman]{Reiser2023MERF}
Christian Reiser, Rick Szeliski, Dor Verbin, Pratul Srinivasan, Ben Mildenhall,
  Andreas Geiger, Jon Barron, and Peter Hedman.
\newblock {MERF}: Memory-efficient radiance fields for real-time view synthesis
  in unbounded scenes.
\newblock \emph{ACM Trans. Graph.}, 42\penalty0 (4):\penalty0 1--12, Jul. 2023.
\newblock URL \url{https://doi.org/10.1145/3592426}.

\bibitem[Schonberger and Frahm(2016)]{Schonberger2016Structure}
Johannes~L. Schonberger and Jan-Michael Frahm.
\newblock Structure-from-motion revisited.
\newblock In \emph{IEEE Conf. Comput. Vis. Pattern Recognit. (CVPR)}, pages
  4104--4113, Las Vegas, NV, USA, Jun. 2016. Inst. Electr. Electron. Eng.
  (IEEE).
\newblock URL \url{https://doi.org/10.1109/CVPR.2016.445}.

\bibitem[Sun et~al.(2022)Sun, Sun, and Chen]{Sun2022Direct}
Cheng Sun, Min Sun, and Hwann-Tzong Chen.
\newblock Direct voxel grid optimization: Super-fast convergence for radiance
  fields reconstruction.
\newblock In \emph{IEEE/CVF Conf. Comput. Vis. Pattern Recognit. (CVPR)}, pages
  5449--5459, New Orleans, LA, USA, Jun. 2022. Inst. Electr. Electron. Eng.
  (IEEE).
\newblock URL \url{https://doi.org/10.1109/CVPR52688.2022.00538}.

\bibitem[Svitov et~al.(2024)Svitov, Morerio, Agapito, and
  Del~Bue]{Svitov2024BillBoard-arxiv}
David Svitov, Pietro Morerio, Lourdes Agapito, and Alessio Del~Bue.
\newblock {BillBoard} splatting ({BBSplat}): Learnable textured primitives for
  novel view synthesis.
\newblock \emph{arXiv}, abs/2411.08508, 2024.
\newblock URL \url{https://doi.org/10.48550/arXiv.2411.08508}.

\bibitem[Tagliasacchi and Mildenhall(2022)]{Tagliasacchi2022Volume-arxiv}
Andrea Tagliasacchi and Ben Mildenhall.
\newblock Volume rendering digest (for {NeRF}).
\newblock \emph{arXiv}, abs/2209.02417, 2022.
\newblock URL \url{https://doi.org/10.48550/arXiv.2209.02417}.

\bibitem[von L{\"u}tzow and Nie{\ss}ner(2025)]{vonLutzow2025LinPrim-arxiv}
Nicolas von L{\"u}tzow and Matthias Nie{\ss}ner.
\newblock {LinPrim}: Linear primitives for differentiable volumetric rendering.
\newblock \emph{arXiv}, abs/2501.16312, 2025.
\newblock URL \url{https://doi.org/10.48550/arXiv.2501.16312}.

\bibitem[Yu et~al.(2024)Yu, Chen, Huang, Sattler, and
  Geiger]{Yu2024MipSplatting}
Zehao Yu, Anpei Chen, Binbin Huang, Torsten Sattler, and Andreas Geiger.
\newblock Mip-splatting: Alias-free {3D} {Gaussian} splatting.
\newblock In \emph{IEEE/CVF Conf. Comput. Vis. Pattern Recognit. (CVPR)}, pages
  19447--19456, Seattle, WA, USA, Jun. 2024. Inst. Electr. Electron. Eng.
  (IEEE).
\newblock URL \url{https://doi.org/10.1109/CVPR52733.2024.01839}.

\bibitem[Zhang et~al.(2020)Zhang, Riegler, Snavely, and
  Koltun]{Zhang2020NeRF++-arxiv}
Kai Zhang, Gernot Riegler, Noah Snavely, and Vladlen Koltun.
\newblock {NeRF}++: Analyzing and improving neural radiance fields.
\newblock \emph{arXiv}, abs/2010.07492, 2020.
\newblock URL \url{https://doi.org/10.48550/arXiv.2010.07492}.

\bibitem[Zhou et~al.(2024)Zhou, Shao, Xu, Bai, Qiu, Liu, Wang, Geiger, and
  Liao]{Zhou2024HUGS}
Hongyu Zhou, Jiahao Shao, Lu~Xu, Dongfeng Bai, Weichao Qiu, Bingbing Liu, Yue
  Wang, Andreas Geiger, and Yiyi Liao.
\newblock {HUGS}: Holistic urban {3D} scene understanding via {Gaussian}
  splatting.
\newblock In \emph{IEEE/CVF Conf. Comput. Vis. Pattern Recognit. (CVPR)}, pages
  21336--21345, Seattle, WA, USA, Jun. 2024. Inst. Electr. Electron. Eng.
  (IEEE).
\newblock URL \url{https://doi.org/10.1109/CVPR52733.2024.02016}.

\end{thebibliography}

\clearpage
\section{Supplementary Material}

\subsection{Methodology}

\paragraph{Depth-dependant scaling.}

In \methodname, the influence $I(p)$ depends only on the normalized ratio $\phi/\min \phi$.
Since a projection is a uniform in‐plane scaling, a single exponent $\sigma$ suffices for all depths.  Indeed, if we scale a triangle by $a>0$, then each projected vertex $v_i$ and pixel $p$ transforms as $v_i\mapsto a\,v_i$ and $p\mapsto a\,p$, so the signed‐distance $d_i(p)$ to each edge satisfies $d_i'(p')=a\,d_i(p)$, implying $\phi'(p')=\max( d_i'(p'))=a\,\phi(p)$ and $\min\phi'=a\,\min\phi$.  Hence 
\begin{equation}
\frac{\phi'(p')}{\min\phi'}=\frac{a\,\phi(p)}{a\,\min\phi}=\frac{\phi(p)}{\min\phi},
\end{equation}
and so $I'(p')=(\phi'(p')/\min\phi')^\sigma=(\phi(p)/\min\phi)^\sigma=I(p)$.

\paragraph{Tile assignment in \methodname}

\methodname is a tile-based renderer that assigns triangles to pixel tiles by computing their screen-space intersections.
Unlike 3D Gaussian Splatting, where the shape has a soft spatial extent, Triangle Splatting is precisely bounded by the projection of its three vertices.
A simple and efficient initial guess for determining tile coverage is to compute the minimum and maximum x and y coordinates of the projected vertices.
While this method is computationally simple, it is conservative, resulting in unnecessary computations for pixels that are not influenced.
As $\sigma$ increases or opacity $o$ decreases, the influence region may not reach the triangle's vertices, causing the rasterizer to process pixels that have no contribution.
To avoid this, we compute a more precise bounding box by determining the maximum distance $d$ a pixel can be from the triangle edge while still contributing at least $\tau_{\text{cutoff}}$ influence.
We define the influence function as:
\begin{equation}
\tau_{\text{cutoff}} = \left( \frac{d}{\mathbf{L(s)}} \right)^{\sigma} \cdot o.
\end{equation}
where $\mathbf{L(s)}$ is the signed distance from the triangle's edges to the incenter. Rearranging gives:
\begin{equation}
d = \mathbf{L(s)} \cdot \left( \frac{\tau_{\text{cutoff}}}{o} \right)^{\frac{1}{\sigma}}.
\end{equation}
We then subtract the distance $d$ from the offset in each edge equation, effectively tightening the triangle’s boundary.  
The minimum and maximum of the updated edge intersections define a more accurate bounding box.  
As $\sigma$ increases or $o$ decreases, this bounding box shrinks accordingly, significantly reducing unnecessary rasterization.

\paragraph{Depth sorting.}
During rasterization, the triangles are currently sorted based on their center, which can lead to popping and blending artifacts during view rotation.
Instead, future work can implement per-pixel sorting of the triangles, as proposed in \cite{Radl2024StopThePop}, where Radl \etal introduce a hierarchical tile-based rasterization approach that performs local per-pixel sorting to ensure consistent visibility and eliminate popping artifacts.

\begin{figure}[t]
\centering
\setlength{\mytmplen}{0.98\linewidth}
\begin{tabular}{c}
\includegraphics[width=\mytmplen]{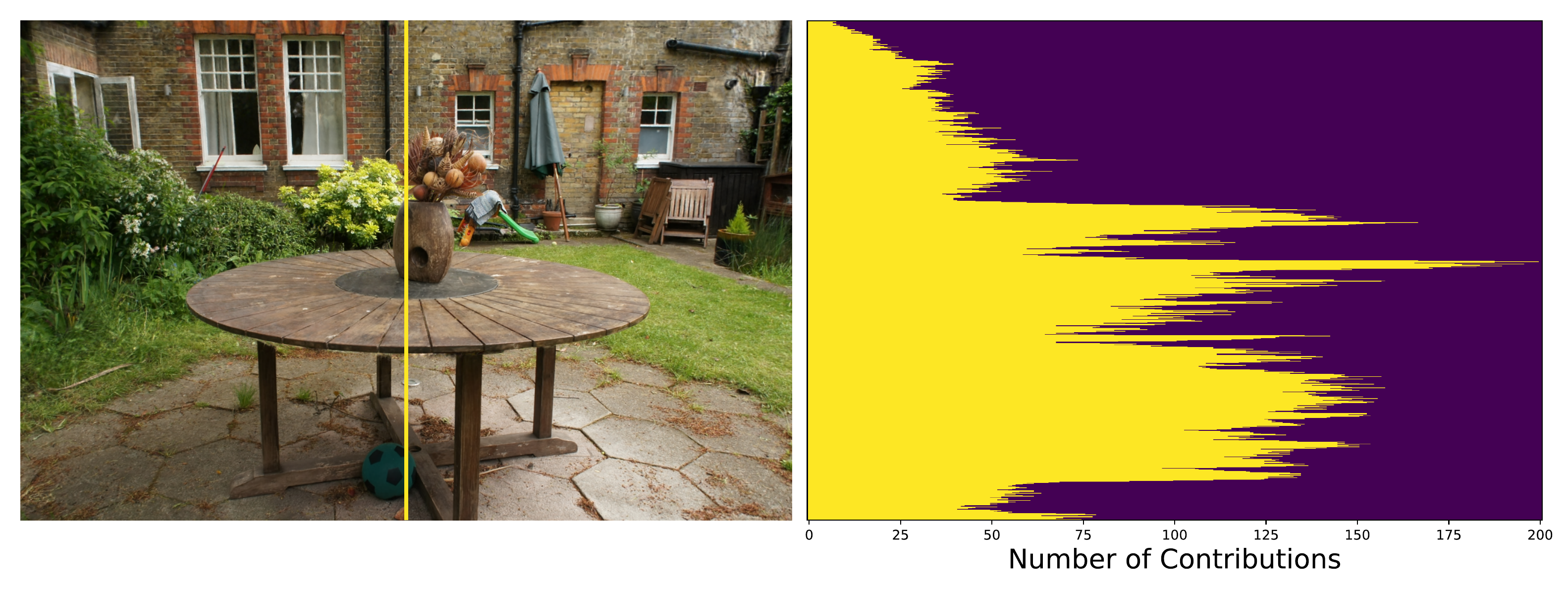}
\\
\end{tabular}
\caption{\small
\myTitle{Number of contributions per pixel}
In background regions where the initial point cloud is sparse, triangles reduce their $\sigma$ to increase coverage across their interior. This leads to more solid shapes and, consequently fewer contributions per pixel.
}
\label{fig:con_per_pixel}
\end{figure}

\paragraph{Densification.}
We prioritize sampling triangles with low $\sigma$ values, corresponding to more \textit{solid} triangles. Our window function ensures that each triangle’s influence is strictly limited to its projected area, preventing any influence beyond its geometric bounds. 
In regions with high triangle density, multiple shapes contribute to each pixel, allowing individual triangles to adopt larger $\sigma$ values and produce softer, more diffuse effects. Conversely, in sparse regions where fewer triangles are present, each triangle must account for more of the pixel-wise reconstruction, leading it to adopt a smaller $\sigma$ and thus contribute more across its surface.
\Cref{fig:con_per_pixel} visualizes the number of contributions per pixel. 
In sparse background regions, triangles adopt lower $\sigma$ values, resulting in solid shapes and fewer overlapping contributions per pixel. 
In contrast, densely sampled regions exhibit higher per-pixel contributions due to many overlapping triangles. As we prioritize adding new triangles to low-density regions, we sample from a probability distribution defined over the inverse of $\sigma$.

\newpage

\subsection{Initialization \& Hyperparameters.}
\begin{wraptable}{r}{0.5\textwidth}
\vspace*{-1em}
\centering
\resizebox{0.48\textwidth}{!}{
\begin{tabular}{l|cc}
Method & Outdoor & Indoor \\
\midrule
feature\_lr & 0.0025 & 0.0025 \\
opacity\_lr & 0.014 & 0.014 \\
lr\_convex\_points\_init & 0.0018 & 0.0015 \\
lr\_sigma & 0.0008 & 0.0008 \\
lambda\_normals & 0.0001 & 0.00004  \\
lambda\_opacity & 0.0055 & 0.0055\\
lambda\_size & 1e-8 & 5e-8 \\
max\_noise\_factor & 1.5 & 1.5 \\
opacity\_dead & 0.014 & 0.014\\
split\_size & 24.0 & 24.0\\
importance\_threshold & 0.022 & 0.0256 \\
\end{tabular}
}
\caption{\small \textbf{Hyperparameters} 
}
\label{tab:hyperparameters}
\vspace{-1.3em}
\end{wraptable}

\paragraph{Initialization.}
At initialization, each triangle is assigned a fixed opacity of $0.28$, and a sigma value of $1.16$. The scaling constant $k$ of the convex hull is set to $2.2$, which defines the spatial extent of the triangle in the 3D scene. These parameter values were determined empirically to ensure stable initialization and consistent rendering behavior across scenes.

\paragraph{Densification.}
We perform densification every $500$ iterations, starting from iteration $500$ until iteration $25{,}000$. At each densification step, we increase the number of shapes by $30\%$.

All other hyperparameters are defined in \Cref{tab:hyperparameters}.

\subsection{More novel-view synthesis results}

\begin{figure*}[b]
\centering
\setlength\mytmplen{0.28\linewidth}
\resizebox{\linewidth}{!}{ 

\begin{tabular}{c@{\hskip 0.2in}c@{\hskip 0.2in}c@{\hskip 0.2in}c}
    
    \makebox[\mytmplen]{Ground Truth} &
    \makebox[\mytmplen]{\textbf{\methodname (ours)}} &
    \makebox[\mytmplen]{2DGS} &
    \makebox[\mytmplen]{3DCS} \\

    \rotatebox{90}{\parbox{2.2cm}{\centering Garden}}
    \zoomin{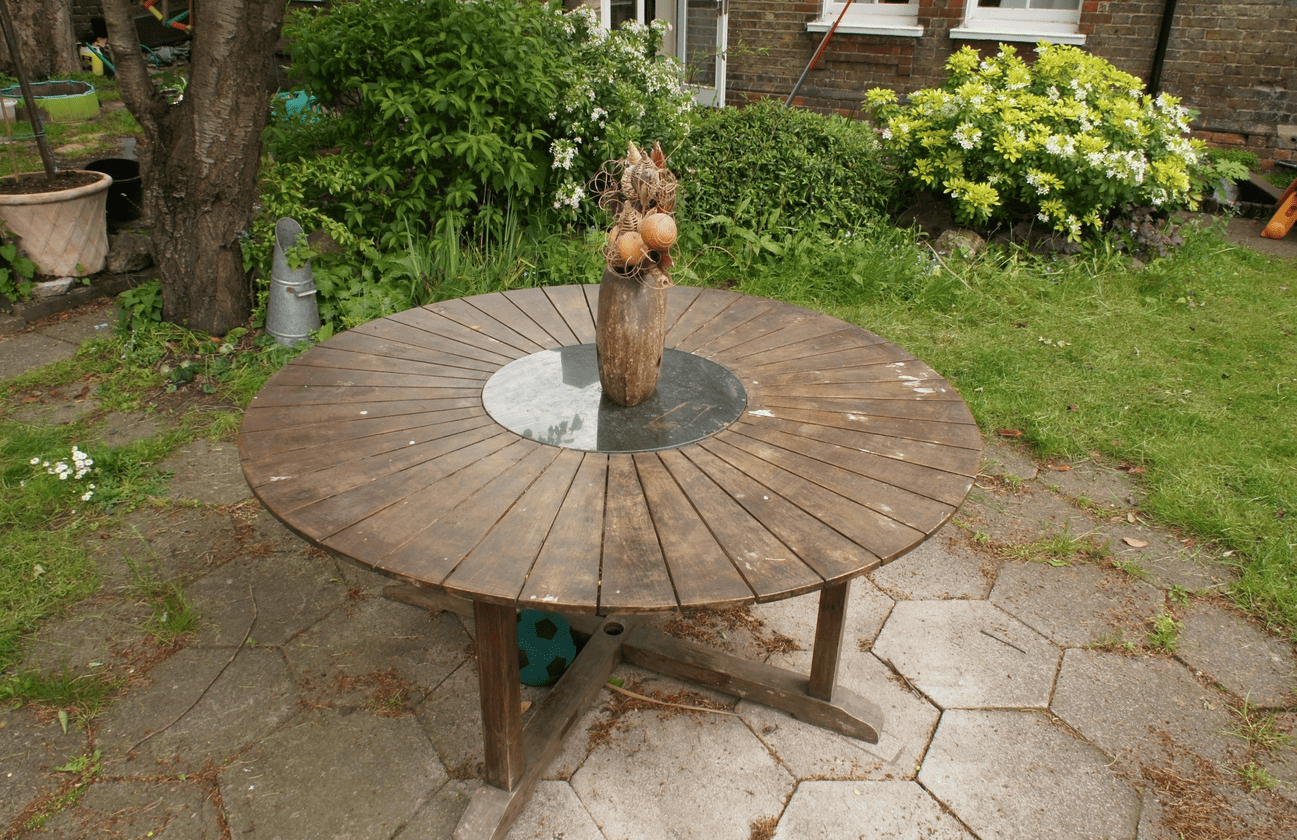}{0.10\mytmplen}{0.49\mytmplen}{0.81\mytmplen}{0.194\mytmplen}{1.2cm}{\mytmplen}{3.5}{red} &
   \zoomin{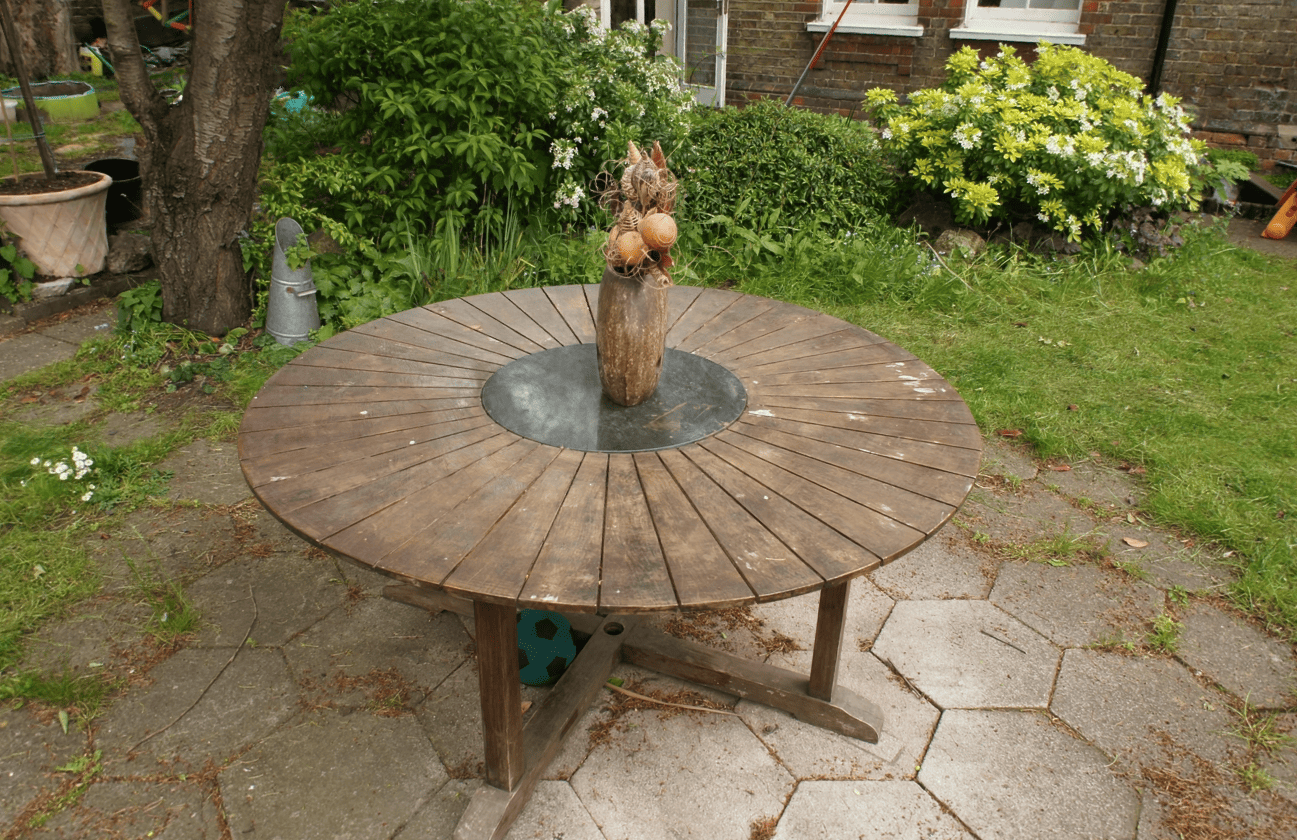}{0.10\mytmplen}{0.49\mytmplen}{0.81\mytmplen}{0.194\mytmplen}{1.2cm}{\mytmplen}{3.5}{red} &
    \zoomin{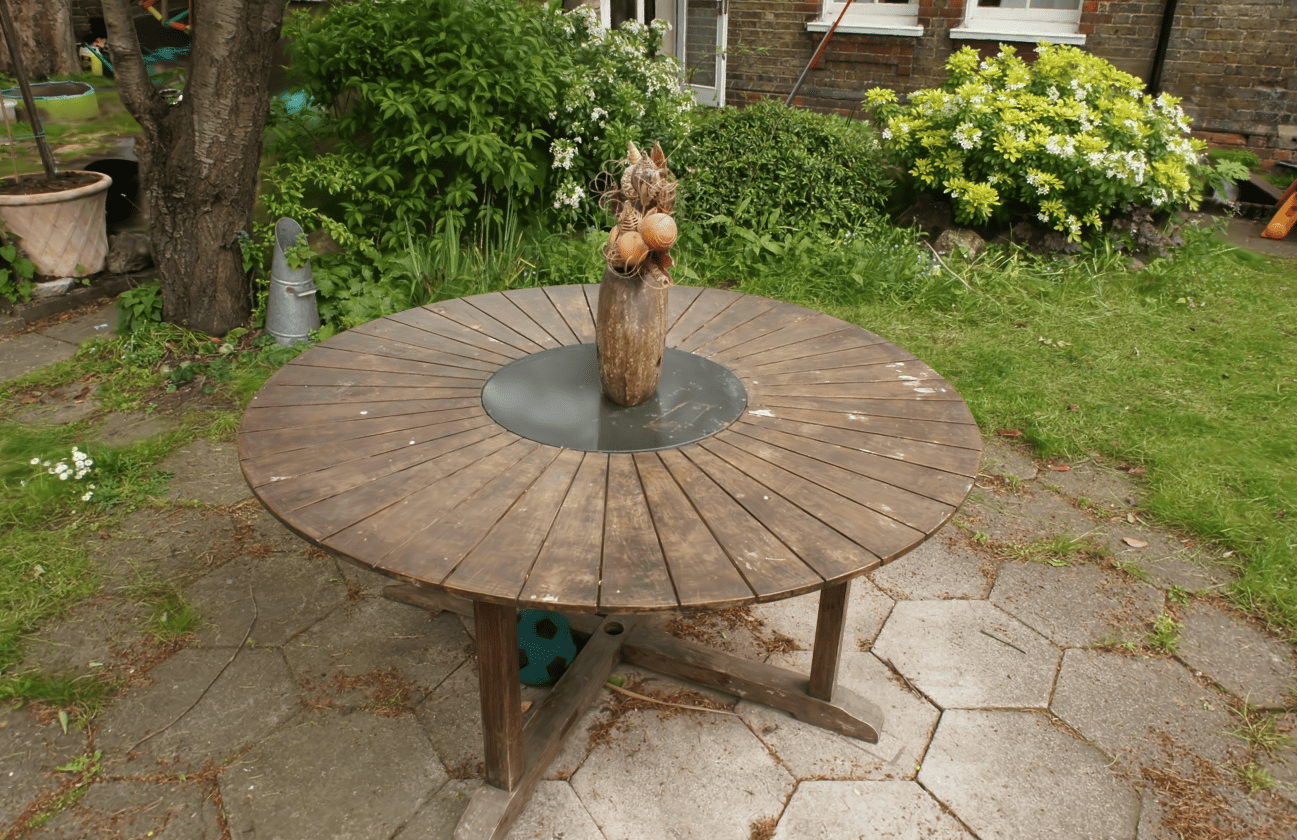}{0.10\mytmplen}{0.49\mytmplen}{0.81\mytmplen}{0.194\mytmplen}{1.2cm}{\mytmplen}{3.5}{red} &
    \zoomin{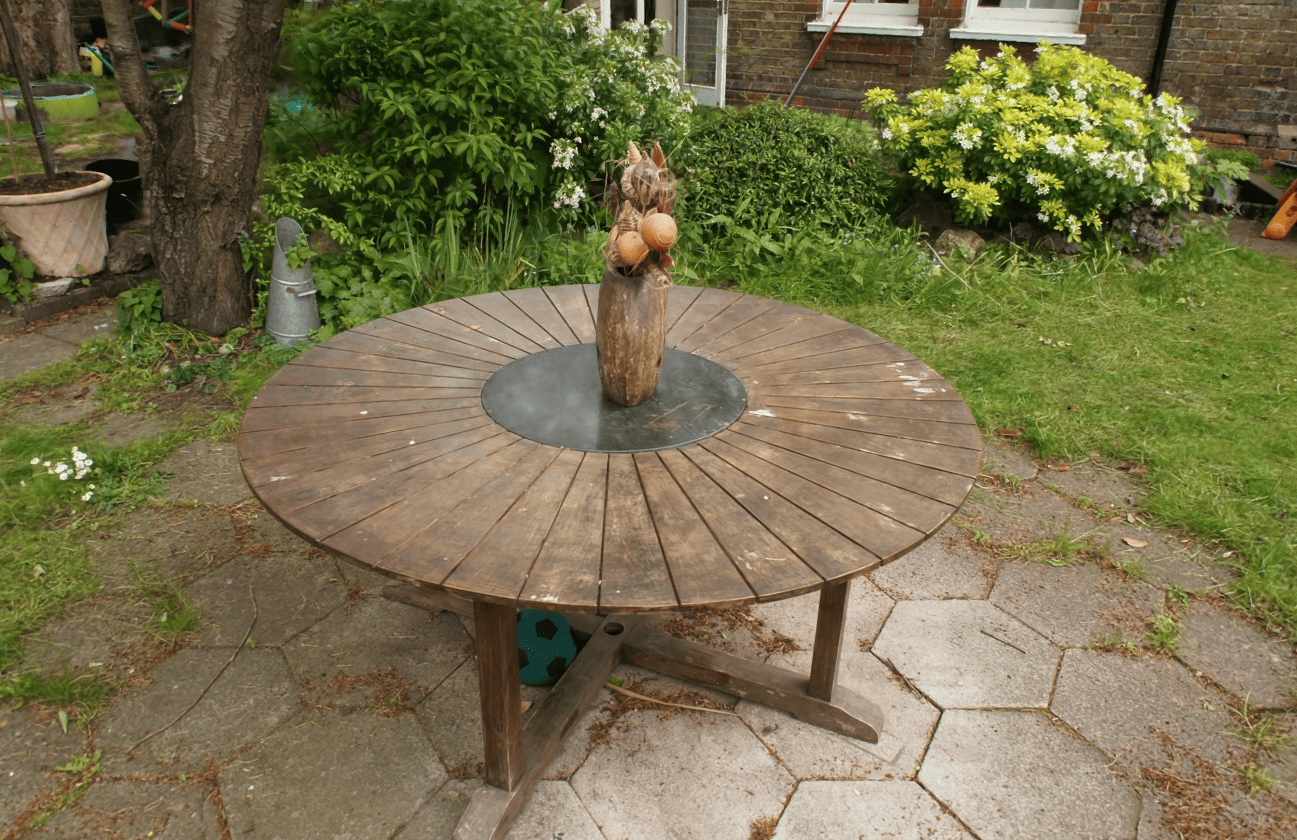}{0.10\mytmplen}{0.49\mytmplen}{0.81\mytmplen}{0.194\mytmplen}{1.2cm}{\mytmplen}{3.5}{red}   \\

   \rotatebox{90}{\parbox{2.2cm}{\centering Kitchen}}
    \zoomin{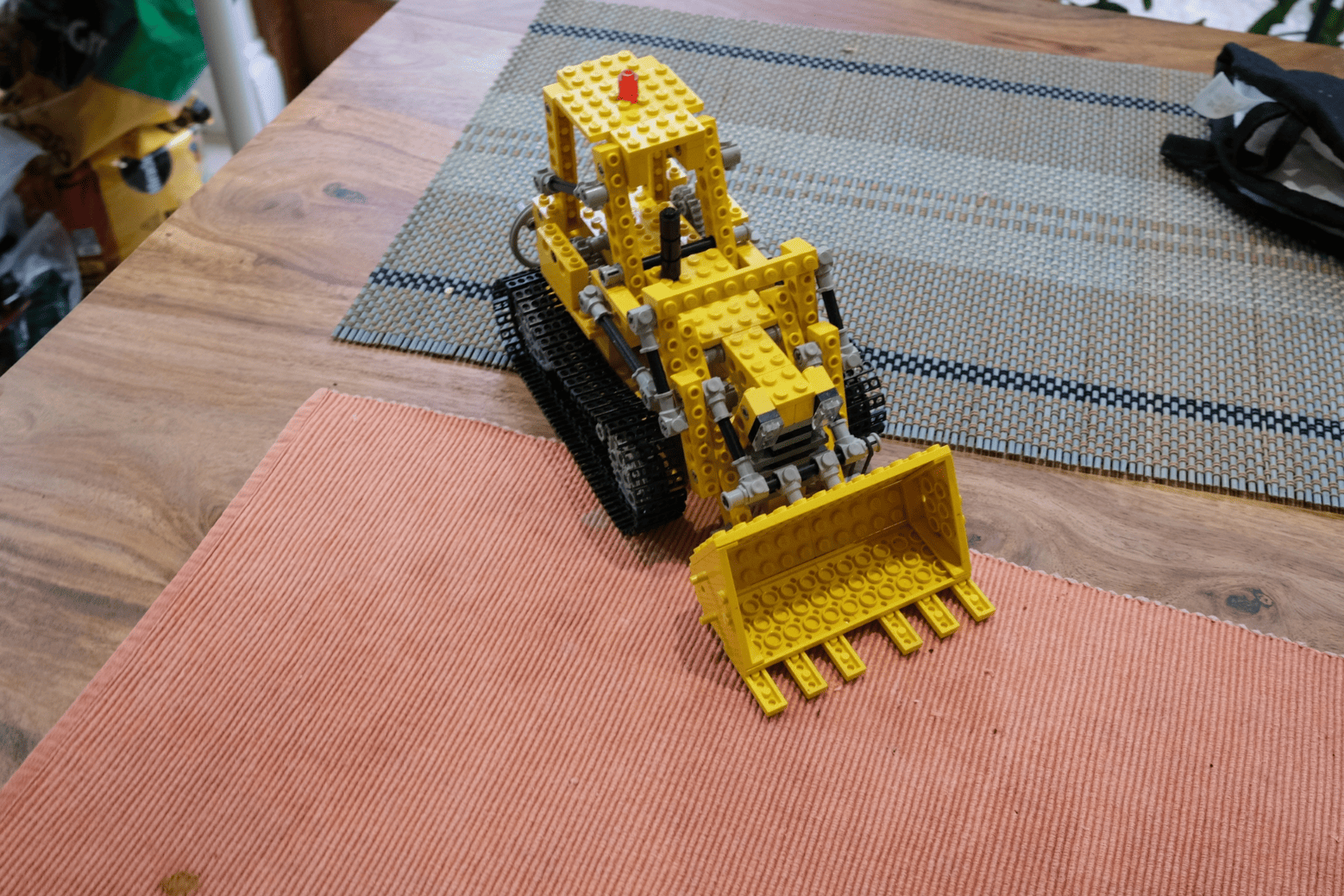}{0.10\mytmplen}{0.57\mytmplen}{0.81\mytmplen}{0.194\mytmplen}{1.2cm}{\mytmplen}{3.5}{red} &
    \zoomin{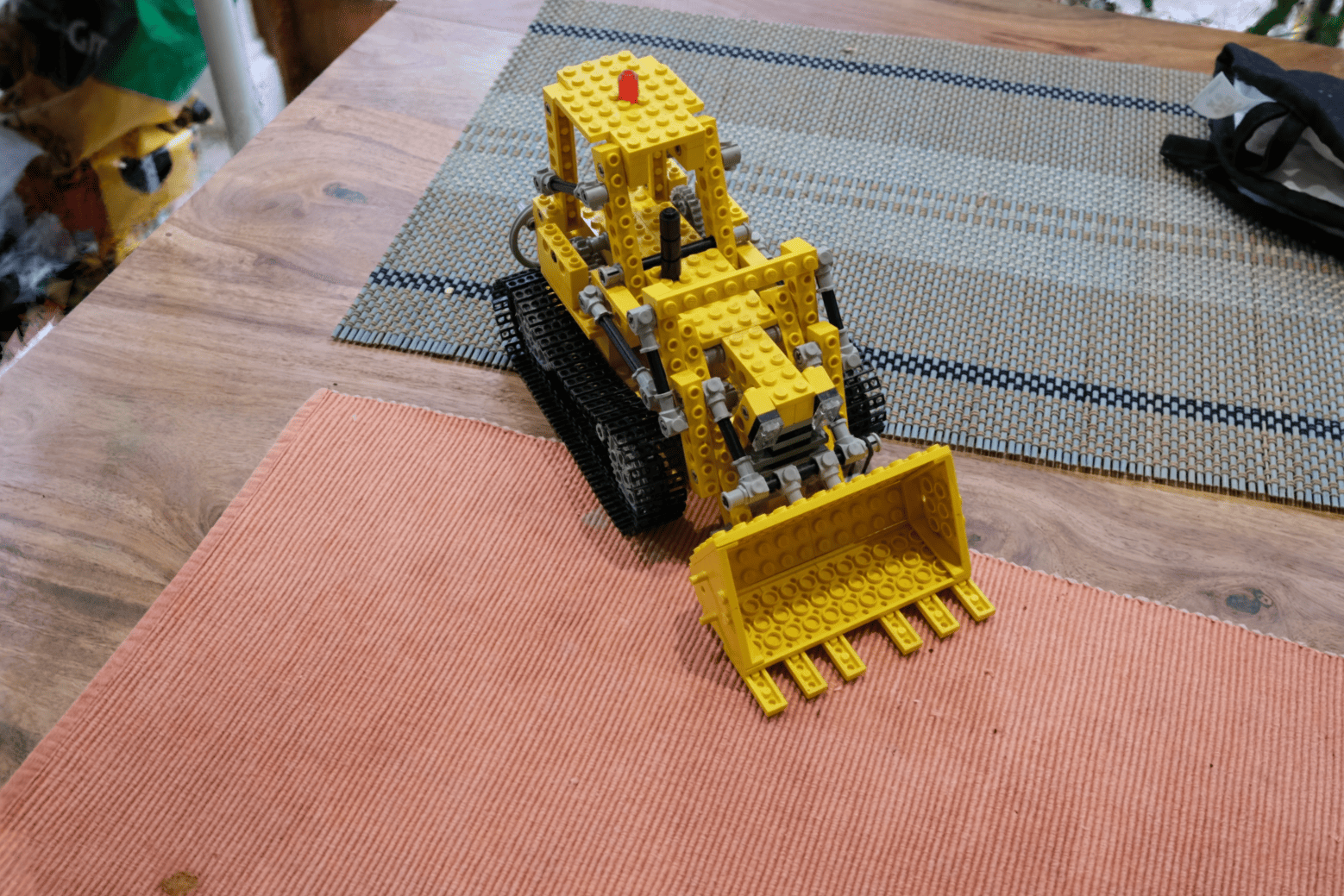}{0.10\mytmplen}{0.57\mytmplen}{0.81\mytmplen}{0.194\mytmplen}{1.2cm}{\mytmplen}{3.5}{red} &
    \zoomin{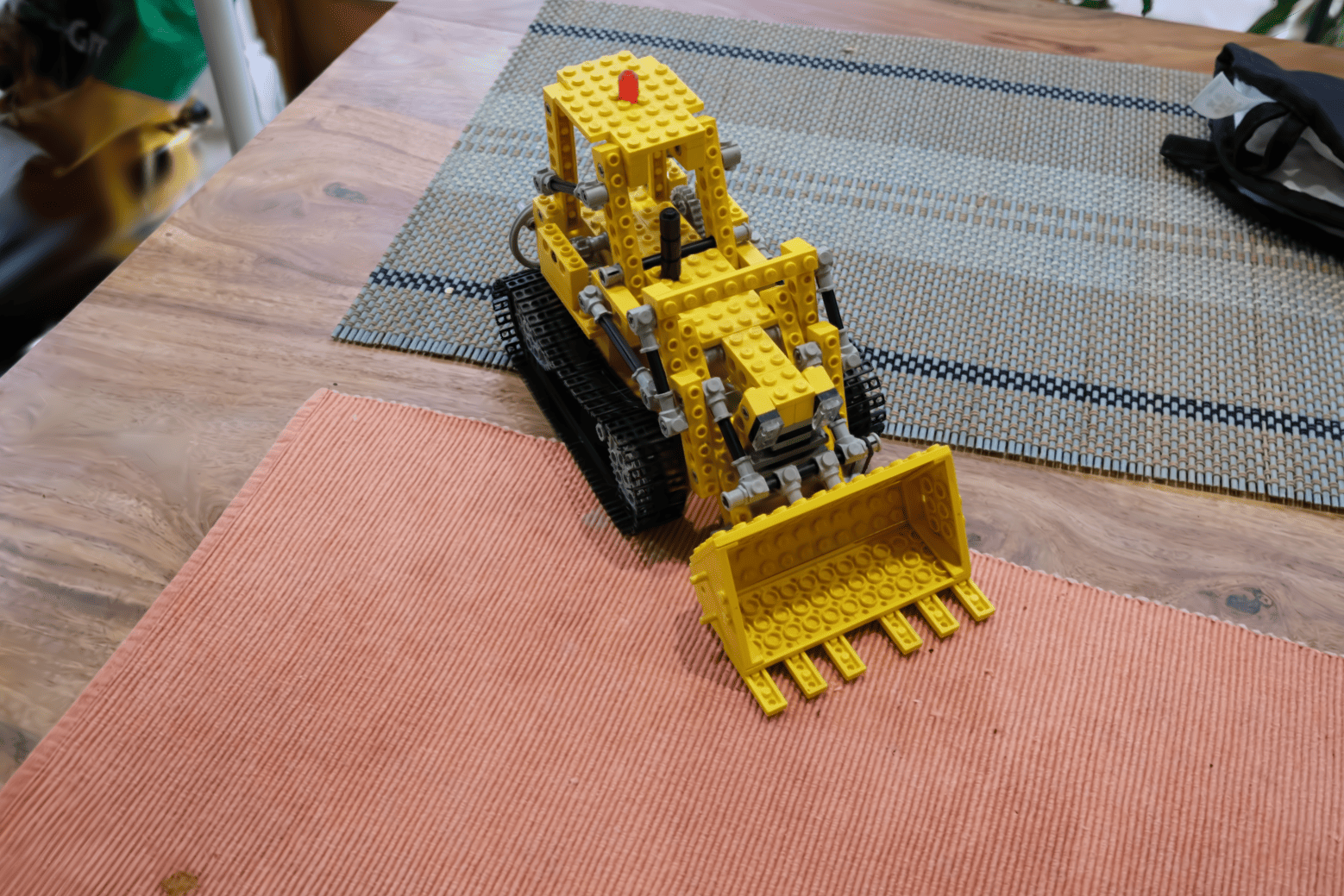}{0.10\mytmplen}{0.57\mytmplen}{0.81\mytmplen}{0.194\mytmplen}{1.2cm}{\mytmplen}{3.5}{red} &
   \zoomin{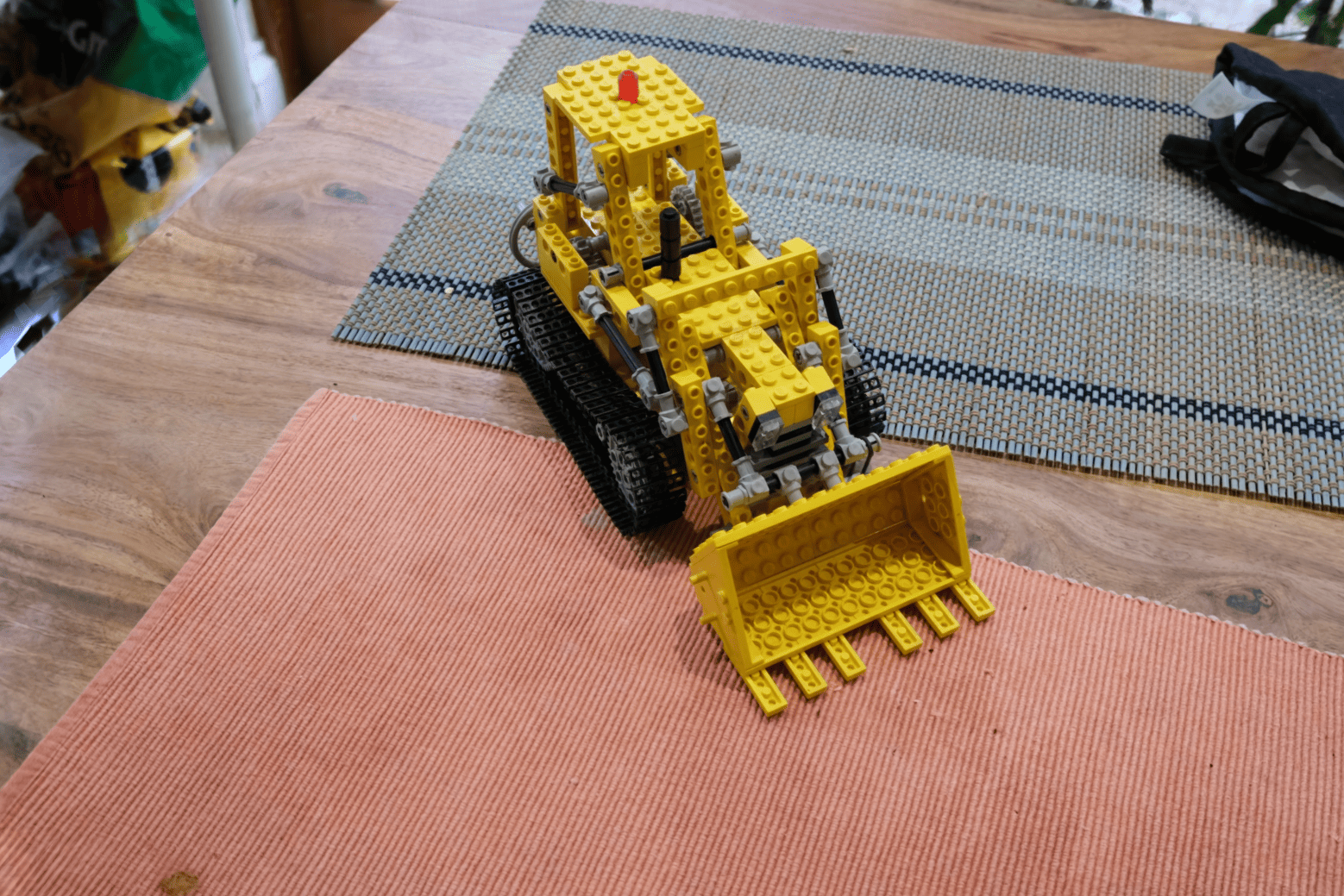}{0.10\mytmplen}{0.57\mytmplen}{0.81\mytmplen}{0.194\mytmplen}{1.2cm}{\mytmplen}{3.5}{red} \\

    \rotatebox{90}{\parbox{2.2cm}{\centering Room}}
    \zoomin{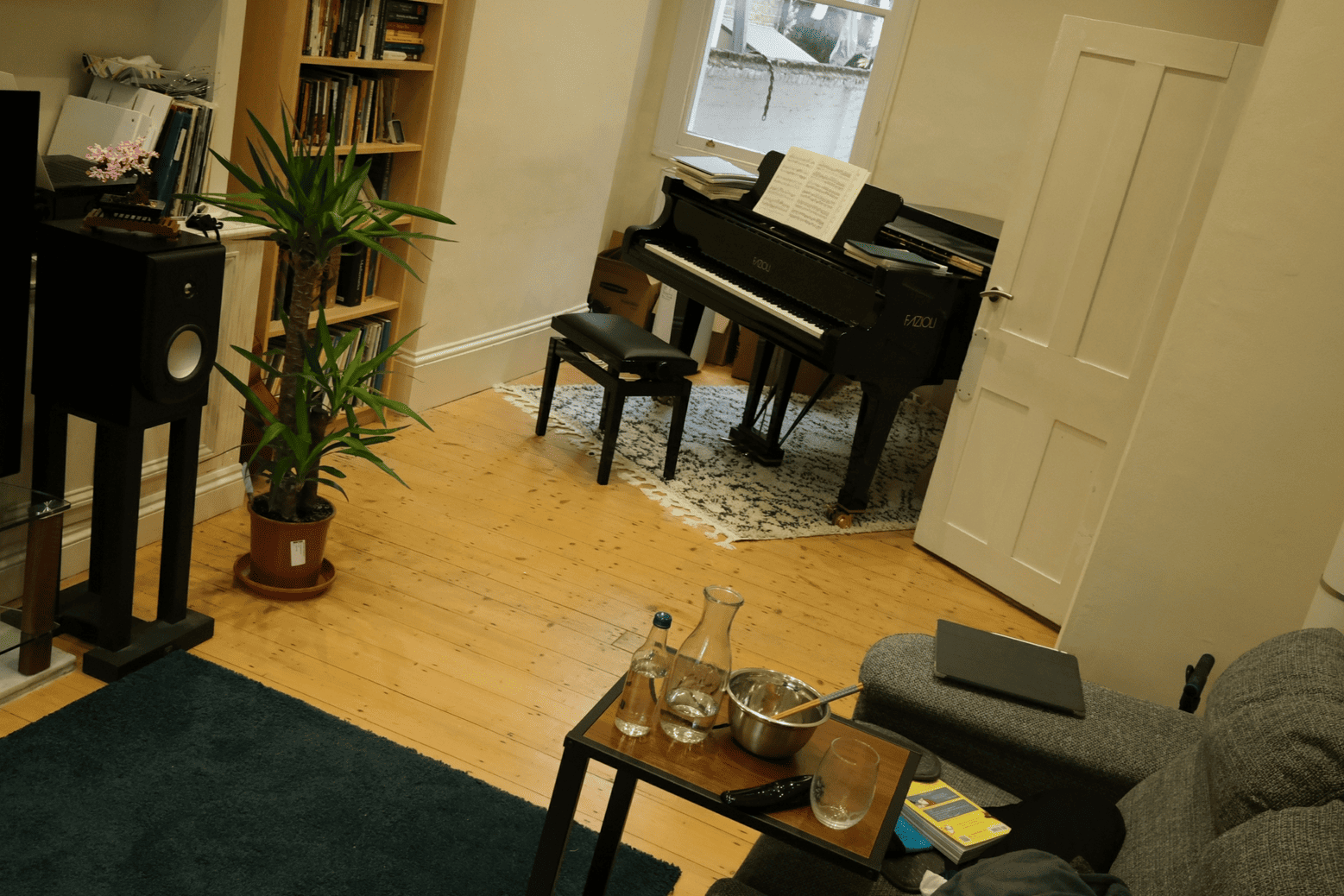}{0.81\mytmplen}{0.594\mytmplen}{0.20\mytmplen}{0.20\mytmplen}{1.2cm}{\mytmplen}{3.5}{red} &
    \zoomin{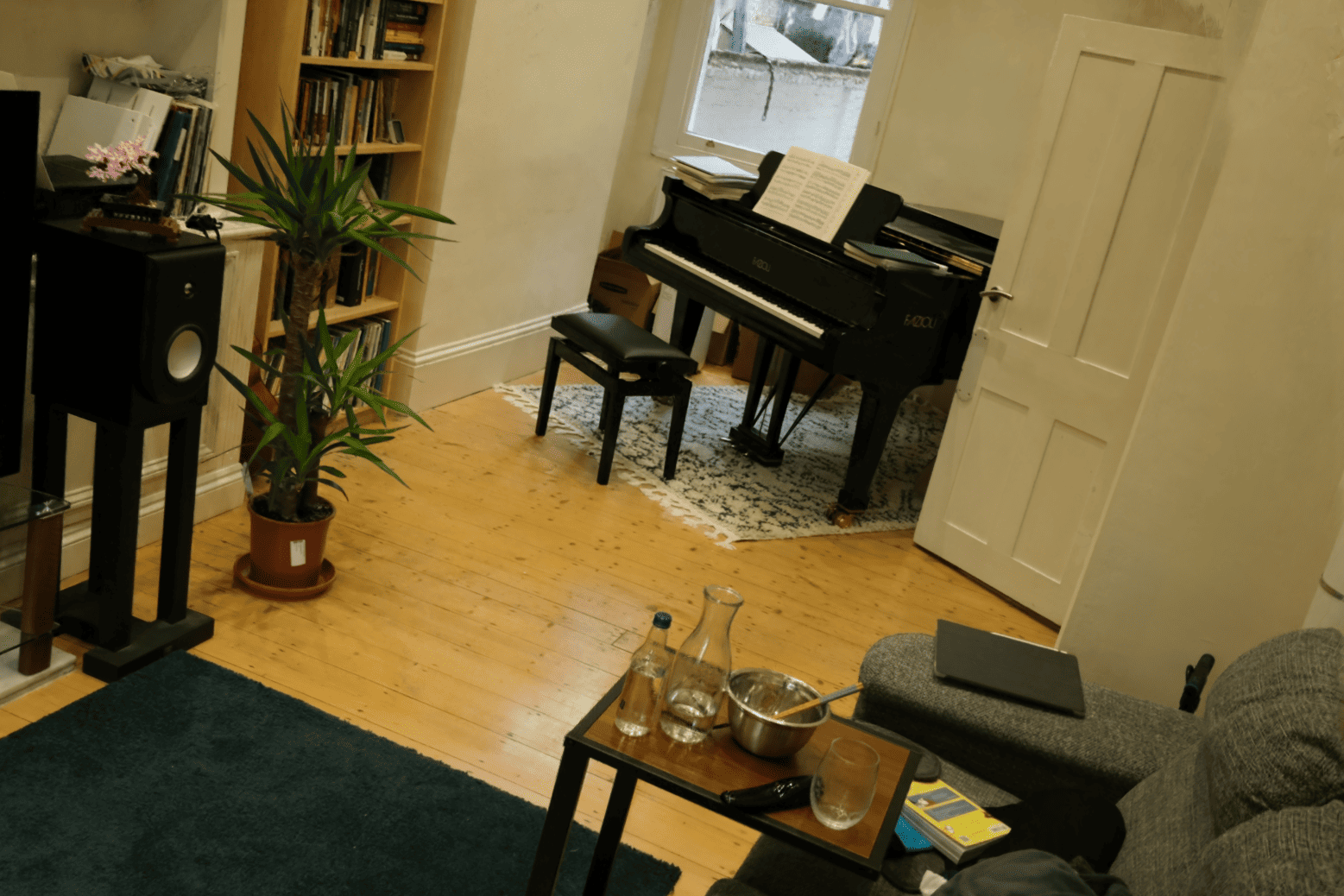}{0.81\mytmplen}{0.594\mytmplen}{0.20\mytmplen}{0.20\mytmplen}{1.2cm}{\mytmplen}{3.5}{red} &
    \zoomin{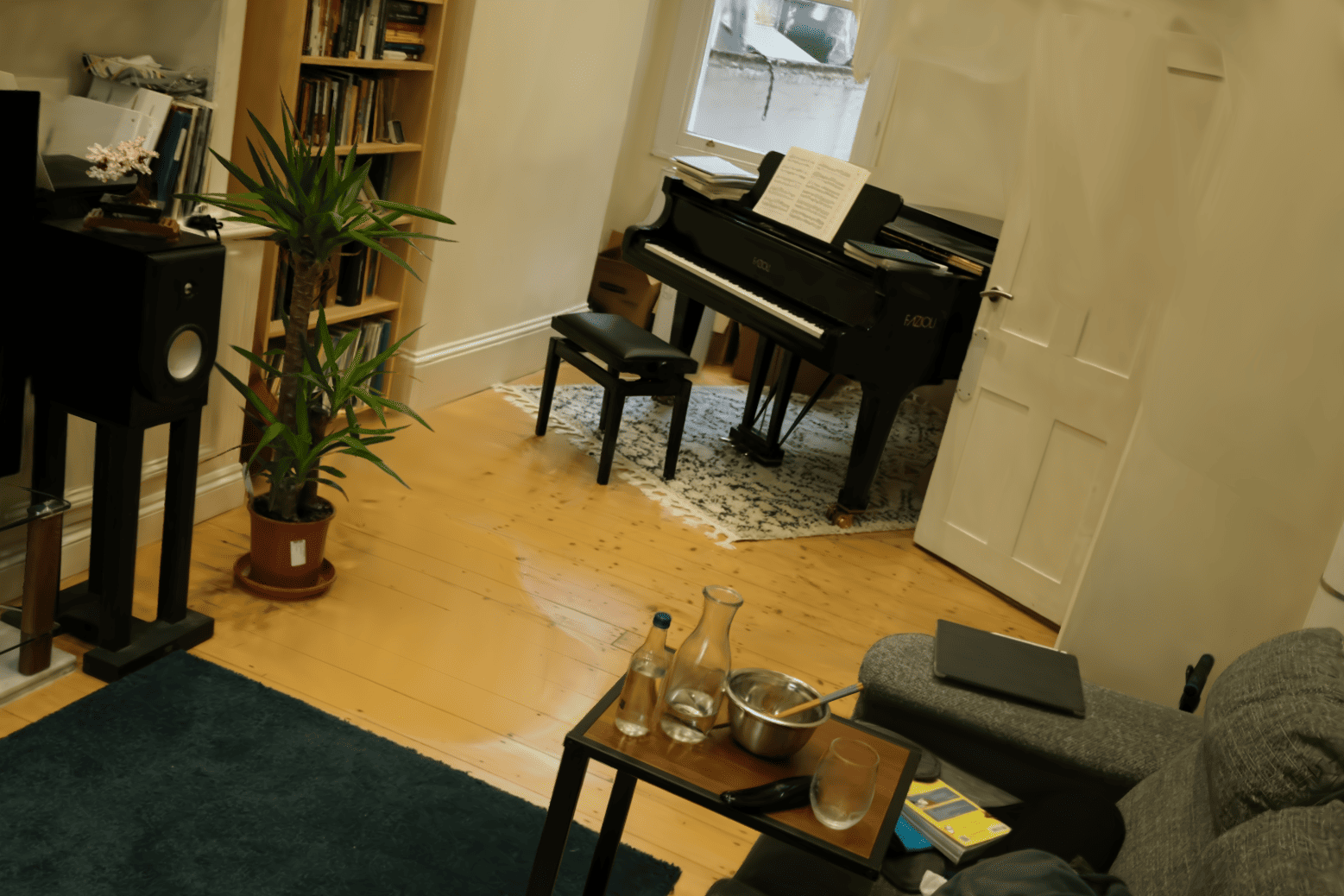}{0.81\mytmplen}{0.594\mytmplen}{0.20\mytmplen}{0.20\mytmplen}{1.2cm}{\mytmplen}{3.5}{red} &
  \zoomin{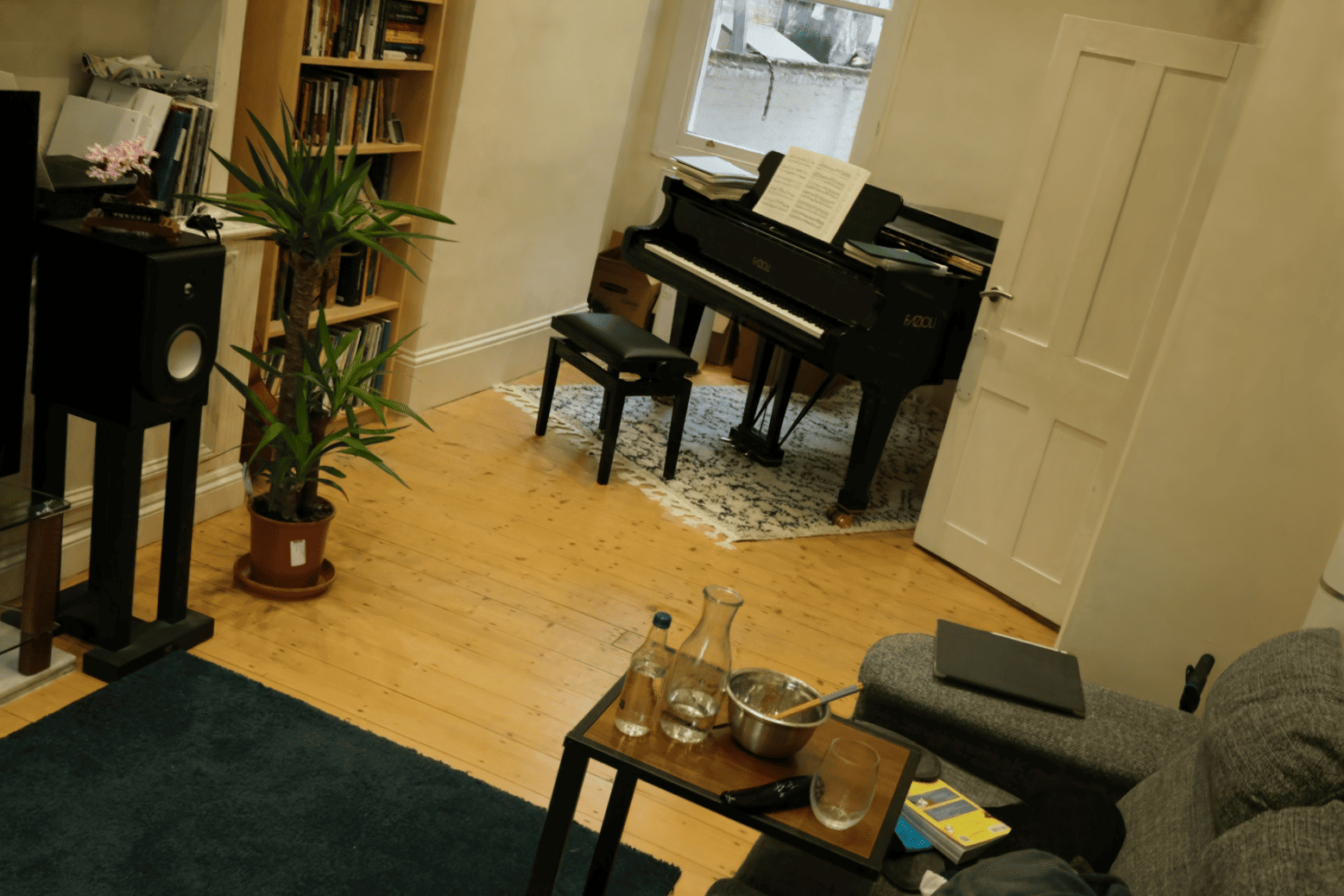}{0.81\mytmplen}{0.594\mytmplen}{0.20\mytmplen}{0.20\mytmplen}{1.2cm}{\mytmplen}{3.5}{red} \\

\end{tabular}
}

\caption{\small \myTitle{Qualitative results}
We visually compare our method to 2DGS~\cite{Huang20242DGaussian} and 3DCS~\cite{Held20253DConvex}.
Triangle Splatting captures finer details and produces more accurate renderings of real-world scenes, with less blurry results than 2DGS, and a higher visual quality than 3DCS~\cite{Held20253DConvex}.
}
\label{fig:qualityresults_supp}
\end{figure*}

\Cref{tab:tt_metrics,tab:4,tab:5,tab:6} present a detailed analysis of the results on the Mip-NeRF360 and Tanks and Temples datasets, while \Cref{fig:qualityresults_supp} presents additional qualitative results.

\subsection{Geometry analysis of \methodname}
\begin{wraptable}{r}{0.3\textwidth}
\vspace*{-1em}
\centering
\resizebox{0.28\textwidth}{!}{
\begin{tabular}{l|c}
Method & \textbf{CD $\downarrow$} \\
\midrule
3DGS \cite{Kerbl20233DGaussian} & 1.96 \\
SuGaR \cite{Guedon2024SuGaR} & 1.33 \\
2DGS \cite{Huang20242DGaussian} & 0.80 \\
BBSplat \cite{Svitov2024BillBoard-arxiv} & 0.91  \\
Ours & 1.06 \\
\end{tabular}
}
\caption{\small \myTitle{Chamfer distance on the DTU dataset\cite{Jensen2014Large}} We report the Chamfer distance on 15 scenes from DTU dataset.
}
\label{tab:dtu_sum}
\vspace{-1.3em}
\end{wraptable}

\begin{figure}[t]
\centering
\setlength{\mytmplen}{0.31\linewidth}
\begin{tabular}{ccc}
\includegraphics[width=\mytmplen]{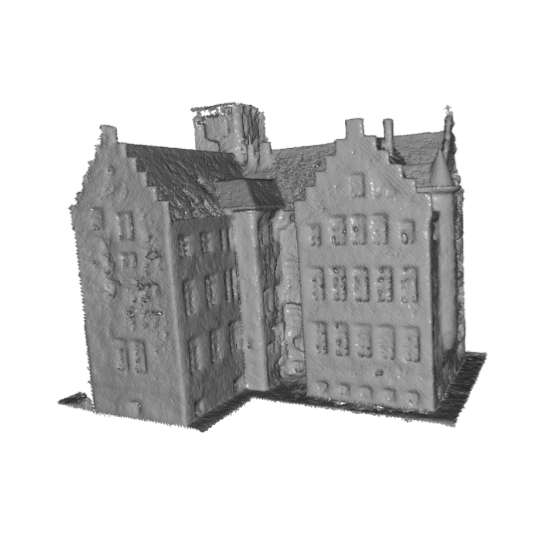} &
\includegraphics[width=\mytmplen]{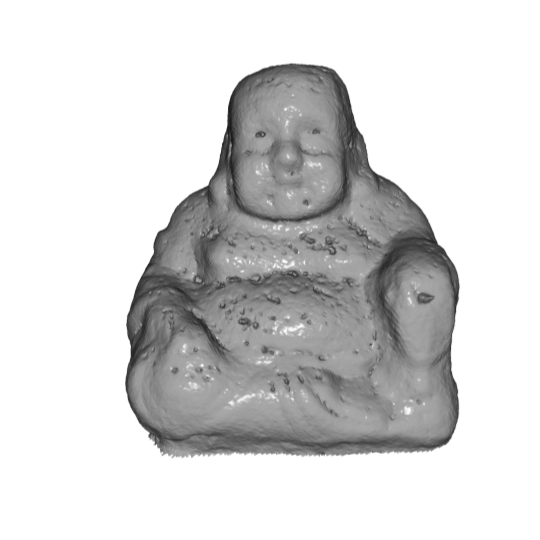} &
\includegraphics[width=\mytmplen]{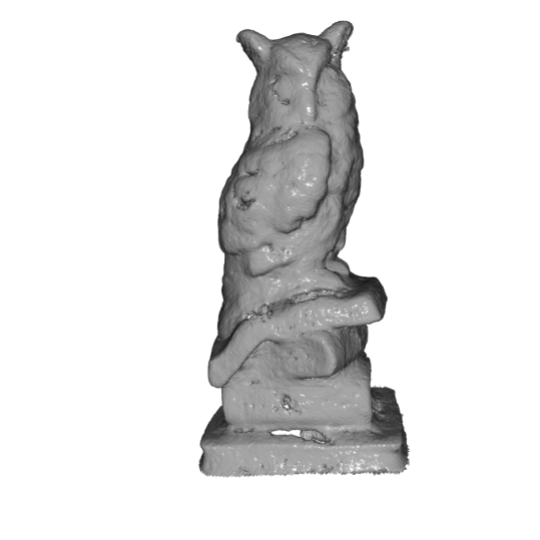} \\
\end{tabular}
\caption{\small
\myTitle{Mesh extraction from depth maps}
We extract meshes by applying TSDF fusion to the predicted depth maps, as followed in 2DGS~\cite{Huang20242DGaussian}.
}
\label{fig:dtu_mesh}
\end{figure}

\begin{figure}[t]
\centering
\setlength{\mytmplen}{0.48\linewidth}
\begin{tabular}{cc}
\includegraphics[width=\mytmplen]{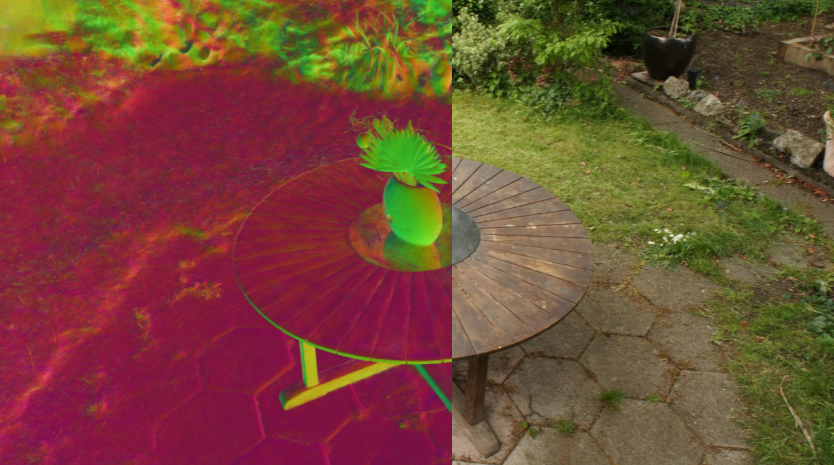} &
\includegraphics[width=\mytmplen]{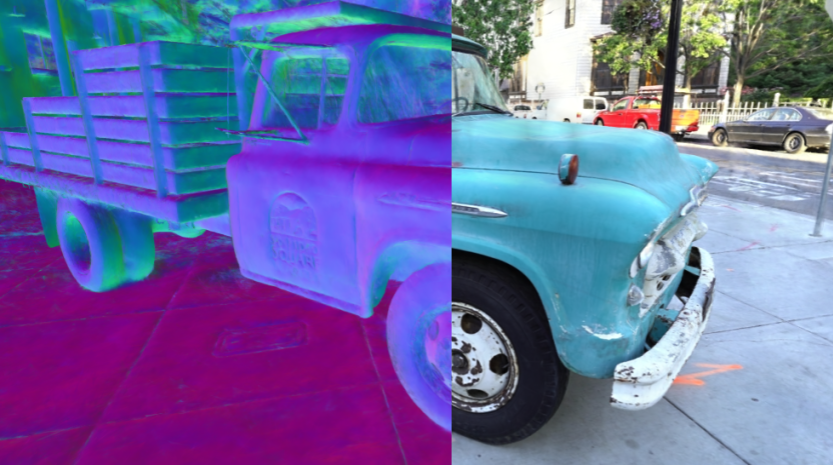} \\
\end{tabular}
\caption{\small
\myTitle{Normal map and rendered image}
The normal map reveals a smooth surface, with the triangle orientations consistently aligned to follow the local geometry.
}
\label{fig:normals}
\end{figure}

Our triangle soup representation is already compatible with standard mesh-based renderers and can be rendered directly without modification.
However, constructing a connected mesh from this representation still requires post-processing.
A promising direction for future research is to design meshing strategies that take full advantage of the explicit triangle structure inherent to our approach.
For completeness, we applied the meshing method used in 2D Gaussian Splatting and present the quantitative results on the DTU dataset~\cite{Jensen2014Large} in \Cref{tab:dtu_sum} and some quantitative results in \Cref{fig:dtu_mesh}.
While this highlights compatibility with existing techniques, we advocate for future work to explore more principled and direct meshing methods tailored specifically to triangle-based representations. 
\Cref{fig:normals} shows the normal map produced by \methodname. The triangles are well aligned with the underlying geometry. For example, on the \textit{Garden} table, all triangles share a consistent orientation and lie flat on the surface.

\newpage

\begin{table}[H]
    \centering
    \renewcommand{\arraystretch}{1.1}
    \tabcolsep=0.15cm
    \begin{tabular}{l|cc|cc|cc}
        \multirow{2}{*}{Method} & \multicolumn{2}{c|}{LPIPS $\downarrow$} & \multicolumn{2}{c|}{PSNR $\uparrow$} & \multicolumn{2}{c}{SSIM $\uparrow$} \\
         & Truck & Train & Truck & Train & Truck & Train \\
        \midrule
        3DGS \cite{Kerbl20233DGaussian} & 0.148 & 0.218 & 25.18 & 21.09 & 0.879 & 0.802 \\
        2DGS \cite{Huang20242DGaussian} & 0.173 & 0.251 & 25.12 & 21.14 & 0.874 & 0.789 \\
        3DCS \cite{Held20253DConvex} & 0.125 & 0.187 & \textbf{25.65} & \textbf{22.23} & 0.882 & 0.820 \\
        \hline
        \methodname & \textbf{0.108} & \textbf{0.179} & 24.94 & 21.33 & \textbf{0.889}  & \textbf{0.823}\\
    \end{tabular}
    \caption{LPIPS, PSNR, and SSIM scores for the Truck and Train scenes of the T\&T dataset.}
    \label{tab:tt_metrics}
\end{table}

\begin{table}[H]
    \centering
          \tabcolsep=0.1cm
    \resizebox{0.98\columnwidth}{!}{      
    \begin{tabular}{l|ccccc|cccc}
     \multirowcell{1}{}  & \multirowcell{1}{Bicycle} & \multirowcell{1}{Flowers} & \multirowcell{1}{Garden} & \multirowcell{1}{Stump} & \multirowcell{1}{Treehill} & \multirowcell{1}{Room} & \multirowcell{1}{Counter} & \multirowcell{1}{Kitchen} & \multirowcell{1}{Bonsai}\\
    \midrule
    3DGS &   0.205 & 0.336 &   \textbf{0.103} &  \textbf{0.210} &   0.317 & 0.220 & 0.204 & 0.129 & 0.205 \\
    2DGS   & 0.218 & 0.346 & 0.115 & 0.222 & 0.329 & 0.223 & 0.208 & 0.133 & 0.214 \\
    3DCS  &  0.216 &   0.322  & 0.113  & 0.227 &   0.317  &   0.193  &    0.182 &   0.117 &   0.182 \\
    \hline
    \methodname & \textbf{0.190} & \textbf{0.284} & 0.106 & 0.214 & \textbf{0.289} & \textbf{0.186} &  \textbf{0.171} & \textbf{0.115} & \textbf{0.169}
    
    \end{tabular}
    }    
    \footnotesize
    \caption{LPIPS score for the MipNerf360 dataset. }
    \label{tab:4}
\end{table}

\begin{table}[H]
    \centering
          \tabcolsep=0.1cm
    \resizebox{0.98\columnwidth}{!}{      
    \begin{tabular}{l|ccccc|cccc}
     \multirowcell{1}{}  & \multirowcell{1}{Bicycle} & \multirowcell{1}{Flowers} & \multirowcell{1}{Garden} & \multirowcell{1}{Stump} & \multirowcell{1}{Treehill} & \multirowcell{1}{Room} & \multirowcell{1}{Counter} & \multirowcell{1}{Kitchen} & \multirowcell{1}{Bonsai}\\
    \midrule
    3DGS &   \textbf{25.24} &   \textbf{21.52} &   \textbf{27.41} &   \textbf{26.55} & \textbf{22.49} & 30.63 & 28.70 & 30.31 & 31.98 \\
    2DGS  & 24.87 & 21.15 & 26.95 & 26.47 & 22.27 & 31.06 & 28.55 & 30.50 & 31.52 \\
    3DCS  & 24.72  & 20.52  & 27.09  & 26.12 &   21.77  &  \textbf{31.70}  &  \textbf{29.02}  &  \textbf{31.96} &   \textbf{32.64} \\
    \hline
    \methodname & 24.9 & 20.85 & 27.20 & 26.29 & 21.94 & 31.05 & 28.90 & 31.32 &31.95
    
    \end{tabular}
    }    
    \footnotesize
    \caption{PSNR score for the MipNerf360 dataset. }
    \label{tab:5}
\end{table}

\begin{table}[H]
    \centering
          \tabcolsep=0.1cm
    \resizebox{0.98\columnwidth}{!}{      
    \begin{tabular}{l|ccccc|cccc}
     \multirowcell{1}{}  & \multirowcell{1}{Bicycle} & \multirowcell{1}{Flowers} & \multirowcell{1}{Garden} & \multirowcell{1}{Stump} & \multirowcell{1}{Treehill} & \multirowcell{1}{Room} & \multirowcell{1}{Counter} & \multirowcell{1}{Kitchen} & \multirowcell{1}{Bonsai}\\
    \midrule
    3DGS &   \textbf{0.771} &   0.605 &  \textbf{0.868} &  \textbf{0.775} & \textbf{0.638} & 0.914 & 0.905 & 0.922 & 0.938 \\
    2DGS  & 0.752 & 0.588 & 0.852 & 0.765 & 0.627 & 0.912 & 0.900 & 0.919 & 0.933 \\
    3DCS  & 0.737  & 0.575  & 0.850 & 0.746 & 0.595  &   0.925  &   0.909 &  \textbf{0.930} &   0.945 \\
    \hline
    \methodname & 0.765 & \textbf{0.614} & 0.863 & 0.759 & 0.611 & \textbf{0.926} & \textbf{0.911} & 0.929 & \textbf{0.947}
    \end{tabular}
    }    
    \footnotesize
    \caption{SSIM score for the MipNerf360 dataset. }
    \label{tab:6}
\end{table}

\newpage

\subsection{Transformation to mesh-based renderer}
\begin{figure}[H]
\centering
\setlength{\mytmplen}{0.48\linewidth}
\begin{tabular}{cc}
\includegraphics[width=\mytmplen]{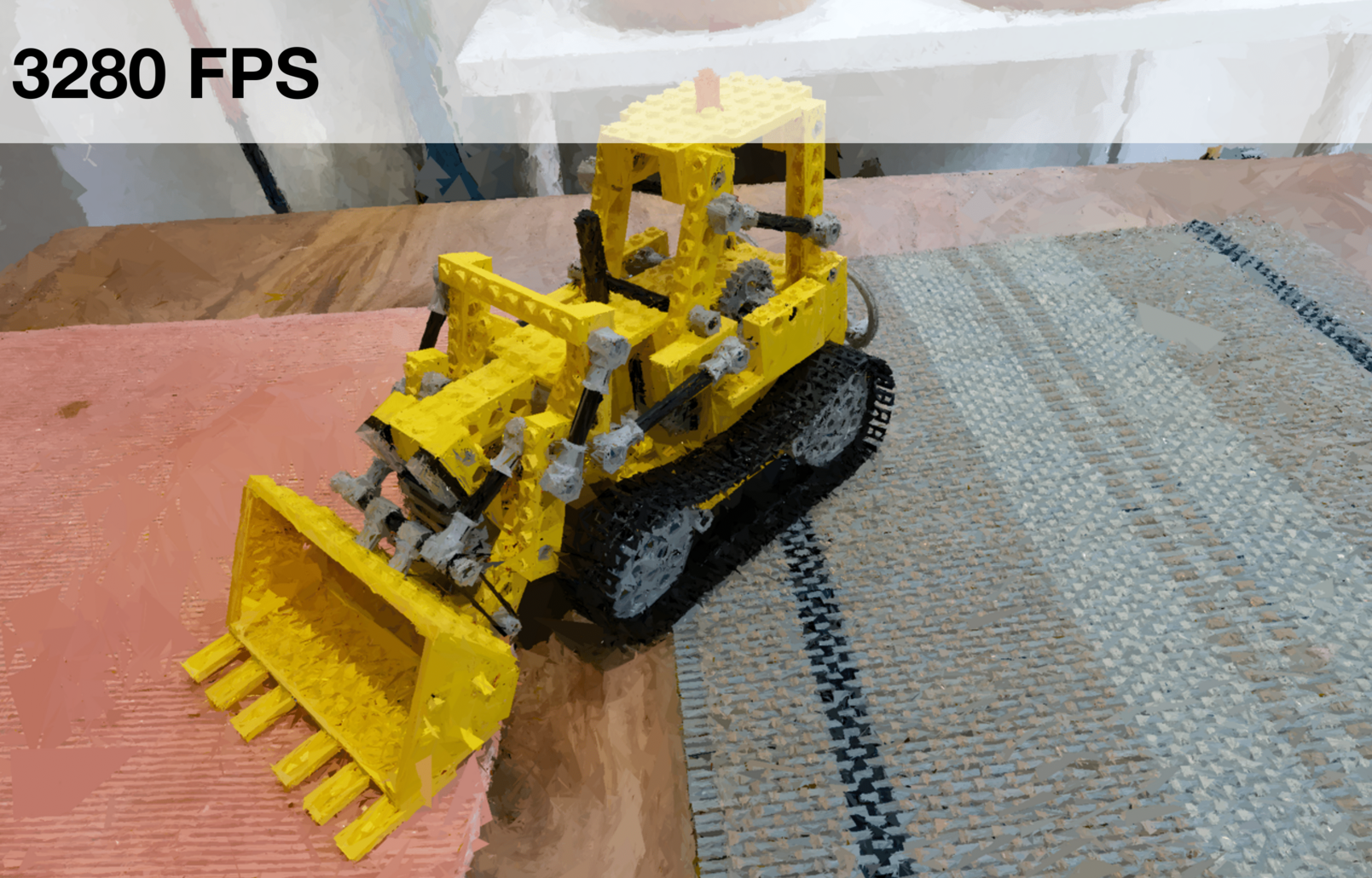} &
\includegraphics[width=\mytmplen]{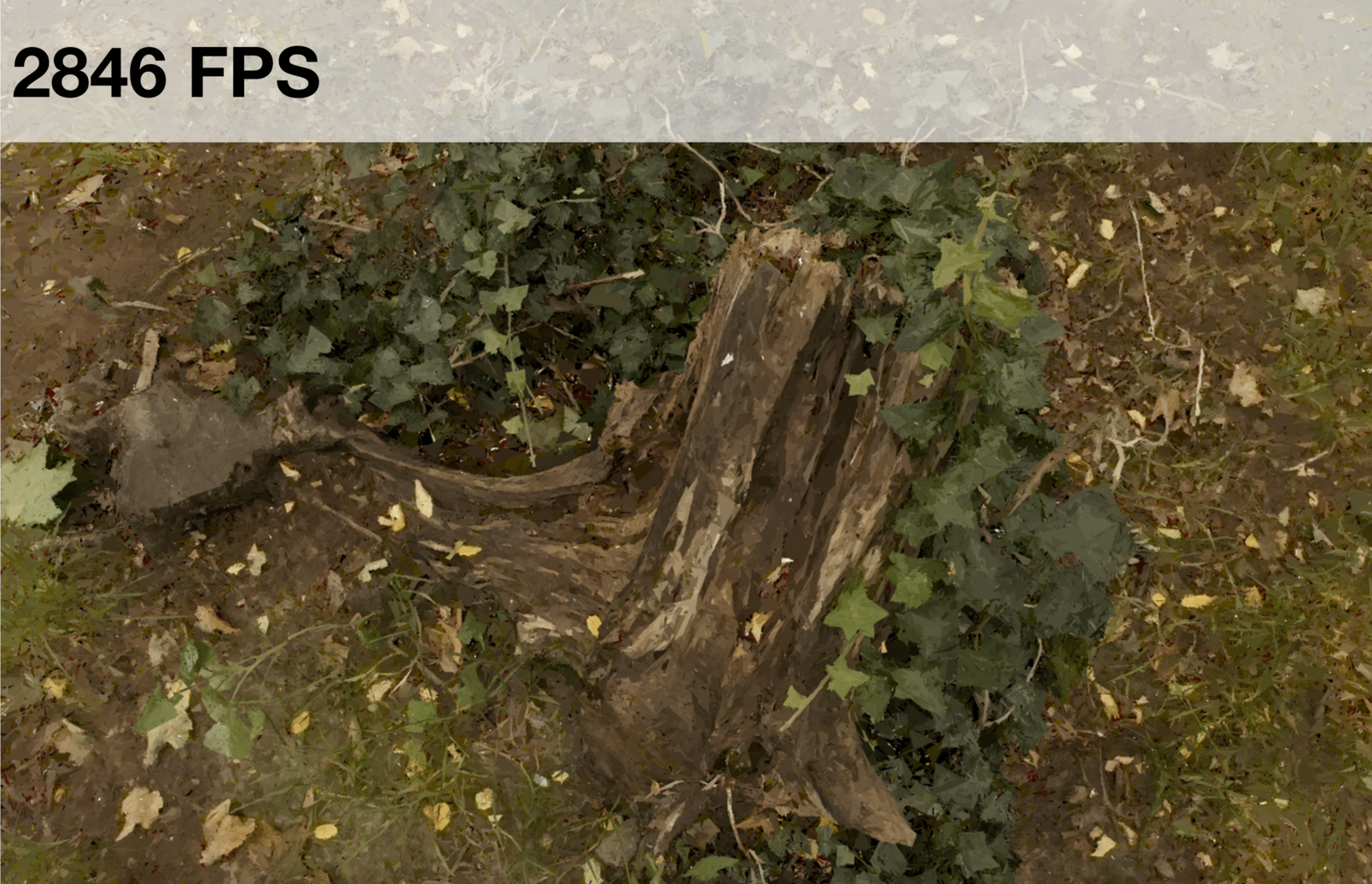} \\
\end{tabular}
\caption{\small
\myTitle{Byproduct of the triangle-based representation}
 In a game engine, we render at 3{,}000 FPS at 1280×720 resolution on a RTX4090.
}
\label{fig:uniy_supp}
\end{figure}

In the final 5{,}000 training iterations, we prune all triangles with opacity below a threshold $\tau_{\text{prune}}$, retaining only solid triangles.
Additionally, we introduce a loss term to encourage higher opacity and lower $\sigma$, ensuring that the final triangles are mostly solid and opaque.
After training, the triangles can be directly converted into any format supported by mesh-based renderers, as our parametrization is fully compatible with standard mesh representations, enabling a seamless transition.
\Cref{fig:uniy_supp} shows some quantitative results with a rendering speed of 3{,}000 FPS.
The visuals are rendered without shaders and were not specifically trained or optimized for game engine fidelity.
Nevertheless, it demonstrates an important first step toward the direct integration of radiance fields into interactive 3D environments. 
Future work could explore training strategies specifically tailored to maximize visual fidelity in mesh-based renderers, paving the way for seamless integration of reconstructed scenes into standard game engines for real-time applications such as AR/VR or interactive simulations.

\end{document}